%% file: main.tex
\documentclass[10pt]{article} 
\usepackage{arxiv}
\usepackage{natbib}
\date{}

\input{math_commands.tex}

\usepackage{xspace}
\usepackage{graphicx}
\usepackage{changepage}
\usepackage[font=small,labelfont=bf]{caption} 
\usepackage{booktabs} 
\usepackage[table,xcdraw]{xcolor}
\usepackage{multirow}
\usepackage{pifont}
\usepackage{makecell}
\usepackage[utf8]{inputenc} 
\usepackage[T1]{fontenc}    
\usepackage{hyperref}       
\usepackage{url}            
\usepackage{booktabs}       
\usepackage{amsfonts}       
\usepackage{nicefrac}       
\usepackage{microtype}      
\usepackage{amsmath}
\usepackage{pifont}
\usepackage{siunitx}
\usepackage{rotating}
\usepackage{graphicx}
\usepackage{subcaption}
\usepackage{multirow}
\usepackage{wrapfig}
\usepackage{listings}
\usepackage{comment}
\usepackage{mathtools}
\usepackage{float}
\usepackage{tabularx}

\usepackage[toc,page,header]{appendix}
\usepackage{minitoc}

\title{A2Perf: Real-World Autonomous Agents Benchmark}
\renewcommand{\headeright}{}
\renewcommand{\undertitle}{}
\author{
Ikechukwu Uchendu$^{1,2*}$ \And
Jason Jabbour$^{2}$ \And
Korneel Van den Berghe$^{3}$ \And
Joel Runevic$^{2}$ \And
Matthew Stewart$^{2}$ \And
Jeffrey Ma$^{2}$ \And
Srivatsan Krishnan$^{2}$ \And
Izzeddin Gur$^{1}$ \And
Austin Huang$^{1}$ \And
Colton Bishop$^{1}$ \And
Paige Bailey$^{1}$ \And
Wenjie Jiang$^{1}$ \And
Ebrahim M. Songhori$^{1}$ \And
Sergio Guadarrama$^{1}$ \And
Jie Tan$^{1}$ \And
Jordan K. Terry$^{4}$ \And
Aleksandra Faust$^{1\S}$ \And
Vijay Janapa Reddi$^{1,2\dagger\S}$
}

\newcommand{\method}{A2Perf\xspace}

\begin{document}

\maketitle
\footnotetext[1]{Google DeepMind}
\footnotetext[2]{Harvard University}
\footnotetext[3]{Delft University of Technology}
\footnotetext[4]{Farama Foundation}
\renewcommand{\thefootnote}{\fnsymbol{footnote}}
\footnotetext[1]{Work done as a student researcher at Google DeepMind.}
\footnotetext[2]{Work done as a visiting researcher at Google DeepMind.}
\footnotetext[4]{Equal advising.}
\renewcommand{\thefootnote}{\arabic{footnote}}
\setcounter{footnote}{0}  
\doparttoc 
\faketableofcontents 

\part{} 

\begin{abstract}

\input{abstract}
\end{abstract}

\section{Introduction}

\label{sec:introduction}

\input{introduction}

\section{Related Work}
\label{sec:related_work}
\input{related_work}

\section{Metrics for Real-World Evaluation}
\label{sec:metrics}
\input{metrics}
fbi

\section{\method~Domains}
\label{sec:domain_background}
\input{domain_background}
\section{Evaluation}
\label{sec:evaluation}
\input{evaluation}
\section{Limitations and Future Work}
\label{sec:limitations_and_future_work}

\input{discussion}
\section{Conclusion}
\label{sec:conclusion}
\input{conclusion}

\bibliography{main}
\bibliographystyle{main}  

\appendix
\input{appendix}

\end{document}

%% file: math_commands.tex
\usepackage{amsmath,amsfonts,bm}

\def\eqref#1{equation~\ref{#1}}

\def\1{\bm{1}}

\DeclareMathAlphabet{\mathsfit}{\encodingdefault}{\sfdefault}{m}{sl}
\SetMathAlphabet{\mathsfit}{bold}{\encodingdefault}{\sfdefault}{bx}{n}



%% file: abstract.tex
Autonomous agents and systems cover a number of application areas, from robotics and digital assistants to combinatorial optimization, all sharing common, unresolved research challenges. It is not sufficient for agents to merely solve a given task; they must generalize to out-of-distribution tasks, perform reliably, and use hardware resources efficiently during training and on-device deployment, among other requirements.
Several classes of methods, such as reinforcement learning and imitation learning, are commonly used to tackle these problems, each with different trade-offs.
However, there is a lack of benchmarking suites that define the environments, datasets, and metrics which can be used to provide a meaningful way for the community to compare progress on applying these methods to real-world problems.
We introduce \method—a benchmarking suite including three environments that closely resemble real-world domains: computer chip floorplanning, web navigation, and quadruped locomotion.
\method provides metrics that track task performance, generalization, system resource efficiency, and reliability, which are all critical to real-world applications.
Using \method, we demonstrate that web navigation agents can achieve latencies comparable to human reaction times on consumer hardware,
reveal important reliability trade-offs between algorithms for quadruped locomotion, and quantify the total energy costs of different learning approaches for computer chip-design.
In addition, we propose a data cost metric to account for the cost incurred acquiring offline data for imitation learning, reinforcement learning, and hybrid algorithms, which allows us to better compare these approaches.
\method also contains baseline implementations of standard algorithms, enabling apples-to-apples comparisons across methods and facilitating progress in real-world autonomy.
As an open-source and extendable benchmark, \method is designed to remain accessible, documented, up-to-date, and useful to the research community over the long term.

%% file: introduction.tex
Autonomous agents observe their environment, make decisions, and perform tasks with minimal human interference \citep{sutton2018reinforcement}.
These agents have been successfully evaluated across a wide range of application domains. However, developing algorithms for autonomous agents that can be deployed in real-world scenarios presents significant challenges \citep{dulac2021challenges}. These challenges include dealing with high-dimensional state and action spaces, partial observability, non-stationarity, sparse rewards, and the need for safety constraints. Furthermore, real-world environments often have multiple objectives, require sample efficiency, and necessitate robust and explainable decision-making. Addressing these challenges is crucial for productionizing reinforcement learning algorithms to real-world problems.

\begin{table}[ht]
\footnotesize
\centering
\setlength{\tabcolsep}{3pt} 
\newcommand{\cmark}{\textcolor{green!80!black}{\ding{51}}}
\newcommand{\xmark}{\textcolor{red}{\ding{55}}}
\begin{tabular}{c>{\centering}p{2.5cm}cccc>{\centering\arraybackslash}p{1.3cm}}
\toprule
\multirow{2}{*}{\textbf{Benchmark}} & \multicolumn{4}{c}{\textbf{Metrics}} & \multirow{2}{*}{\makecell{\textbf{Real-World} \\ \textbf{Tasks}}} & \multirow{2}{*}{\makecell{\textbf{Offline} \\ \textbf{Datasets}}} \\ 
\cmidrule(lr){2-5}
 & \textbf{Generalization} & \textbf{System} & \textbf{Data Cost} & \textbf{Reliability} &  & \\ 
\midrule
\method & \cmark & \cmark & \cmark & \cmark & \cmark & \cmark \\
\midrule
D5RL \citep{rafailov2024d5rl} & \cmark & \xmark & \xmark & \xmark & \cmark & \cmark \\

NeoRL \citep{qin2022neorl} & \xmark & \xmark & \xmark & \xmark & \cmark & \cmark \\
OGBench \citep{park2024ogbench} & \cmark & \xmark & \xmark & \xmark & \cmark & \cmark \\
Meta-World \citep{meta_world} & \cmark & \xmark & \xmark & \xmark & \cmark & \xmark \\
DM Control \citep{tassa2018deepmind} & \xmark & \xmark & \xmark & \xmark & \xmark & \xmark \\
Jumanji \cite{bonnet2023jumanji}  & \cmark & \xmark & \xmark & \xmark & \cmark & \xmark \\
DSRL \cite{liu2023datasets}  & \xmark & \xmark & \xmark & \xmark & \cmark & \cmark \\
Safety Gym \citep{ji2023safety_gym} & \xmark & \xmark & \xmark & \cmark & \cmark & \xmark \\
ALE \citep{bellemare2013arcade} & \xmark & \xmark & \xmark & \xmark & \xmark & \xmark \\
MineRL \citep{guss2019minerl} & \cmark & \xmark & \xmark & \xmark & \xmark & \cmark \\
Loon Benchmark \citep{balloon_learning_env}  & \cmark & \xmark & \xmark & \xmark & \cmark & \cmark \\ 
\bottomrule
\end{tabular}
\vspace{0.5em}
\caption{\method compared to existing benchmarks that evaluate autonomous agents. Checkmarks (\cmark) indicate the presence of a feature or metric, while crosses (\xmark) denote its absence.
\method distinguishes itself by including metrics for generalization, system resource efficiency, data cost, and reliability, in addition to providing real-world tasks and offline datasets.
Here, real-world tasks refer to those that are often performed in industrial or consumer contexts. 
The selected domains in \method are designed to closely mirror real-world challenges, ensuring the relevance and transferability of the benchmark results to practical applications. }
\label{tab:benchmark_comparison}
\vspace{-1.2em}
\end{table}

To enable researchers to develop algorithms with real-world deployment considerations in mind, there is a need for benchmarks that incorporate practical metrics. These include metrics such as the compute required for training and inference, wall-clock time, and effort expended on data collection. While there are existing benchmarks for autonomous agents \citep{guss2019minerl, meta_world, kempka2016vizdoom, bellemare2013arcade, MinigridMiniworld23, tassa2018deepmind}, most only evaluate an agent's raw performance on the same task on which it was trained, without considering numerous other metrics that matter in real-world production training and deployment scenarios.

In this paper, we introduce \method\footnote{\method~code: \url{https://anonymous.4open.science/r/A2Perf-2BFC}}, a benchmarking framework that aims to bridge the gap between algorithms research and real-world applications by providing a comprehensive evaluation platform for autonomous agents, thereby expanding the applicability of reinforcement learning to a wide range of practical domains. In addition, it comes equipped with a critical set of metrics for fair assessment.

\method incorporates three challenging domains based on prior work \citep{coumans2023motion, mirhoseini2021graph, gur2021environment} that closely mirror scenarios that have been demonstrated in the real world: computer chip-floorplanning, website form-filling and navigation, and quadruped locomotion.
In addition, these domains were chosen because they inherently exhibit a small Sim2Real gap.
The computer chip-floorplanning domain \citep{mirhoseini2020chip_placement, mirhoseini2021graph} was used to help create an iteration of Google's tensor processing unit\footnote{History of the Tensor Processing Unit: \url{https://shorturl.at/Bo71S}}, where the agent optimizes the layout of chip components.
In the website form-filling and navigation domain \citep{gur2018learning_to_navigate_the_web, gur2021environment}, agents autonomously navigate and interact with websites in a Google Chrome\footnote{Google Chrome Browser: \url{ https://www.google.com/chrome/}} browser, making it identical to real-world web navigation.
The quadruped locomotion domain \citep{peng2020learning_agile_imitate} has demonstrated successful transfer of learned walking gaits to the Unitree Laikago\footnote{Unitree Laikago: \url{https://shorturl.at/FD6uP}} robot. 

Furthermore, to address the metrics gap, \method provides an open-source benchmarking suite that evaluates agents across four key metric categories: (1) data cost, which quantifies the effort required to gather training data for imitation learning, (2) application performance, relating to the quality of the agent's task-specific execution, and its ability to generalize to tasks that it was not explicitly trained to perform; (3) system resource efficiency, focusing on the hardware resources used during training and inference; and (4) reliability, denoting the consistency of an agent's performance over training and inference. While three domains and for classes of metrics are currently available, A2Perf allows for straightforward expansion to benchmark on custom domains and for custom metrics.

The key contributions of this work include:
\begin{itemize}
\item A \textbf{unified evaluation framework} that combines critical metrics spanning data cost, system resources, reliability, and generalization, applied across three diverse domains with demonstrated real-world applicability: computer chip floorplanning, web navigation, and quadruped locomotion.
\item A novel \textbf{data cost metric} that enables fair comparisons between different learning paradigms (e.g., imitation learning vs reinforcement learning)
\item An \textbf{open-source, extensible implementation} that facilitates reproducible evaluation and community contributions
\end{itemize}

\begin{figure}[!ht]
    \centering
    \includegraphics[width=0.95\linewidth]{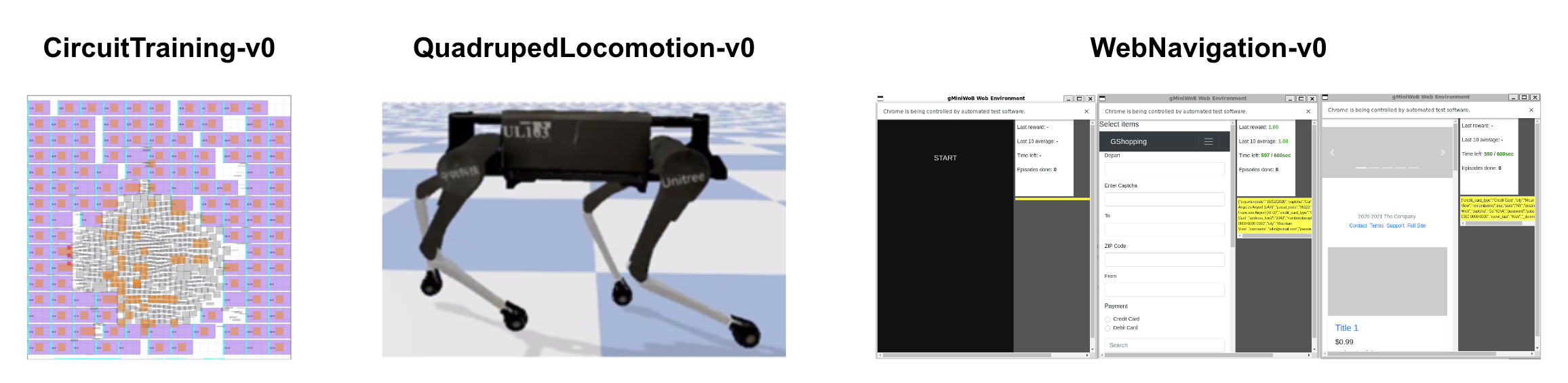}
    \caption{The three domains included in A2Perf: computer chip floorplanning for optimizing integrated circuit layouts, web navigation for automated form filling and website interaction, and quadruped locomotion for robotic control. These specific domains were selected based on their demonstrated transfer from simulation to real-world applications.}
    \label{fig:a2perf-domains}
\end{figure}

Our experimental evaluation yields valuable insights into the real-world applicability of autonomous agents across diverse domains.
In the web navigation domain, we explore the feasibility of deploying agents by analyzing their inference time, power usage, and memory consumption, demonstrating that trained agents can operate with latencies comparable to human reaction times on consumer-grade hardware. Furthermore, the reliability metrics \citep{chan2019measuring} prove crucial in selecting agents for chip floorplanning and quadruped locomotion tasks.
For chip floorplanning, we find that the PPO \citep{schulman2017proximal} algorithm provides more consistent initial placements compared to DDQN \citep{van2015ddqn}, reducing variability for designers.
In quadruped locomotion, PPO exhibits superior stability during training, while SAC \citep{haarnoja2018soft_actor_critic} demonstrates more consistent gaits during deployment, highlighting the importance of considering reliability in real-world scenarios.
These findings underscore A2Perf's ability to provide a comprehensive evaluation of autonomous agents, facilitating their successful deployment in practical applications.

%% file: related_work.tex
\paragraph{Benchmarking Autonomous Agents}

Table \ref{tab:benchmark_comparison} offers a comparison between \method and existing benchmarks, highlighting the unique contributions of our proposed benchmarking suite. Existing benchmarks for autonomous agents, such as those introduced by \citet{brockman2016openai, bellemare2013arcade, tassa2018deepmind}, provide diverse environments for testing various algorithms. However, these benchmarks often focus on specific types of learning algorithms or on evaluating particular desirable qualities in autonomous agents. For example, \citet{fu2020d4rl} and \citet{rl_unplugged} evaluate offline reinforcement learning \citep{levine2020offline}, while \citet{meta_world} focuses on meta-reinforcement learning \citep{wang2016learning_to_reinforcement_learn}. Similarly, \citet{ye2021mastering} tests sample efficiency, \citet{guss2019minerl} challenges agents on long-horizon tasks, and \citet{coin_run} evaluates generalization ability.
While these benchmarks provide insights, they do not fully capture the challenges faced by autonomous agents in real-world applications \citep{dulac2021challenges}.
Environments, benchmarks, and datasets have been made to foster the development of autonomous agents in real-world scenarios, such as aerial balloon navigation~\citep{balloon_learning_env}, autonomous driving~\citep{waymo_open}, website navigation \citep{gur2021environment}, and furniture assembly \citep{lee2021ikea}. Yet, these initiatives are often domain-specific and lack the comprehensive scope needed to evaluate agents across a wide range of real-world challenges  as outlined by prior work \citep{dulac2021challenges}, which forms the basis for our work.

Consequently, there remains a need for more benchmarking suites that encompass a diverse set of tasks and environments, reflecting the complexity and variety of problems encountered in real-world applications. Among recent benchmarks, NeoRL \citep{qin2022neorl} provides realistic environments for stock trading, utility management and industrial control, while OGBench \citep{park2024ogbench} emphasizes realistic tasks in the offline, goal-conditioned setting. Jumanji \citep{bonnet2023jumanji} focuses on providing fast, JAX-accelerated \citep{jax2018github} implementations of combinatorial optimization tasks inspired by industry applications.

\method differentiates itself by incorporating different real-world domains such as web navigation and computer chip floorplanning, while also including system performance, data cost, and reliability metrics in a unified package. This comprehensive approach allows for a more holistic evaluation of autonomous agents across diverse, practically relevant tasks and crucial deployment considerations.

\paragraph{Benchmarking System Performance} In addition to evaluating task-specific performance metrics, analyzing the end-to-end performance cost and examining the hardware resources required to apply learning algorithms on specific environments has gained significant attention~\citep{wu2022sustainable,Howwe’re47:online}. Benchmarks such as MLPerf \citep{reddi2020mlperf} and DAWNBench \citep{Coleman2017DAWNBenchA} have been developed to assess various aspects of commercial deep learning workloads across training and inference, considering a diverse class of systems. Furthermore, recent studies have investigated the environmental impact of deep learning by quantifying the carbon footprint associated with training and inference using large neural network models \citep{patterson2021carbon}. This line of research has also extended to autonomous agents, with works like QuaRL demonstrating reduced energy consumption and emissions through lower-precision distributed training \citep{krishnan2022quarl}. 
Despite these efforts, there remains a need for evaluating the system performance and energy consumption of autonomous agents to provide valuable insights into their practical feasibility and sustainability.

\paragraph{Reliability Metrics for Reinforcement Learning}
Reliability is a concern in reinforcement learning (RL), as current metrics often rely on point estimates of aggregate performance, which fail to capture the true performance of algorithms and make it challenging to draw conclusions about the state-of-the-art \citep{agarwal2021deep_rl_stat_precipice, henderson2018deep_rl_that_matters, colas2018many}.
The increasing complexity of benchmarking tasks has made it infeasible to run hundreds of training runs, necessitating the development of tools to evaluate reliability based on a limited number of runs \citep{agarwal2021deep_rl_stat_precipice}.
For real-world deployments, reliability is essential to ensure that RL algorithms perform consistently and robustly across different conditions and environments.
To assess reliability, it is essential to consider metrics across three axes of variability: time (within a training run), runs (across random seeds), and rollouts of a fixed policy \citep{chan2019measuring}. By incorporating reliability metrics into \method, we will be able to better assess the robustness and consistency of RL algorithms.

%% file: metrics.tex



\subsection{Motivation}
\label{ssec:metrics_motivation}
Deploying autonomous agents in real-world applications generally follows a progression: practitioners first collect demonstration data or other offline data, train the agent using custom or off-the-shelf algorithms, and finally deploy the agent to a target domain.
\method provides metric categories that evaluate each stage: data cost, system performance, application performance, and reliability. 
Data cost quantifies the effort required to collect the initial demonstrations or other offline data used for training.
Application performance metrics are used to evaluate how effectively an agent learns to perform tasks within its domain.
For deployment, system performance determines what computational resources the agent requires, while reliability metrics reveal how consistently it performs beyond simple averages of achieved rewards.
Table~\ref{tab:all_metrics} summarizes the individual metrics corresponding to each category.
The relative importance of these categories varies depending on the specific application domain, so in Section~\ref{sec:domain_background}, we state which metric categories are most critical for each of \method's domains to help guide practitioners in selecting the most suitable agent for their use case.

\begin{table}[!t]
\setlength{\tabcolsep}{3pt} 
\small
\centering
\resizebox{\textwidth}{!}{
\begin{tabular}{>{\centering\arraybackslash}m{2cm}>{\centering\arraybackslash}m{3cm}>{\centering\arraybackslash}m{3cm}>{\centering\arraybackslash}m{4cm}>{\centering\arraybackslash}m{3cm}}
\toprule
\textbf{} & \textbf{Data Cost} & \textbf{System} & \textbf{Reliability} & \textbf{Application} \\
\midrule
\textbf{Training} & 
\begin{tabular}{@{}c@{}}Training Sample Cost\end{tabular} & 
\begin{tabular}{@{}c@{}}Energy\\Power\\RAM Usage\\Wall-Clock Time\end{tabular} & 
\begin{tabular}{@{}c@{}}Dispersion (Runs)\\Dispersion (Time)\\Long-Term Risk (Time)\\Risk (Runs)\\Short-Term Risk (Time)\end{tabular} & 
\begin{tabular}{@{}c@{}}Episodic Returns\\Generalization Returns\end{tabular} \\
\midrule
\textbf{Inference} & 
\begin{tabular}{@{}c@{}}N/A\end{tabular} & 
\begin{tabular}{@{}c@{}}Inference Time\\Power\\RAM Usage\end{tabular} & 
\begin{tabular}{@{}c@{}}Dispersion (Rollouts)\\Risk (Rollouts)\end{tabular} & 
\begin{tabular}{@{}c@{}}N/A \\ \end{tabular} \\
\bottomrule
\end{tabular}
}
\vspace{0.5em}
\caption{\method assesses four categories—data cost, system performance, reliability, and application performance—during training and inference. These metrics provide a comprehensive evaluation of autonomous agents. See Section \ref{sec:metrics} for detailed descriptions of the metric categories. Data Cost is marked as "N/A" at inference time since pre-existing data and demonstrations are only used during training. Application metrics are marked as "N/A" during inference since performance and generalization are evaluated based on the complete training process.}
\label{tab:all_metrics}
\end{table}

\subsection{Data Cost}
\label{sec:training_sample_cost}
Autonomous agents can be trained either with or without expert demonstrations. Methods that leverage expert demonstrations, such as imitation learning (IL) \citep{gail, bc_zero, robotics_transformer, bansal2018chauffeurnet,kelly2019hg,Sermanet2018TimeContrastive}, aim to learn from pre-collected datasets of human or expert agent trajectories. On the other hand, methods like online \citep{mnih2015human} and offline RL \citep{levine2020offline,uchendu2023jump,nair2020awac,ball2023efficient} do not necessarily require expert demonstrations and instead learn through interaction with the environment or sub-optimal demonstration data.

Comparing agent performance trained using different approaches is challenging but important to gain a holistic picture of the costs and trade-offs involved.
IL methods may be more sample efficient than RL methods, as they do not need to interact with the environment online. However, this perspective overlooks the \textit{effort} required to collect demonstration data used for IL.

To facilitate fair comparisons between these approaches, we propose the \textbf{training sample cost} metric, which quantifies the effort required to obtain offline datasets used by the agent.
In this context, we denote the training sample cost of an offline dataset $D$ as $C_D$.
An agent that uses samples from datasets $D_1, D_2, \dots, D_K$ will incur a total training sample cost of $\texttt{Training Sample Cost} = \sum_{i=1}^K C_{D_i}$.
The datasets $D_i$ could be of different \textit{expertise} levels, meaning they contain demonstrations from agents or humans with varying levels of task proficiency.

The training sample cost can be measured with any metric that meaningfully represents the effort required to generate samples for imitation learning. For example, the cost could be expressed in terms of money spent on human labor or computational resources, hours invested in collecting the data, or any other relevant metric.
The choice of metric may depend on the specific application and the type of data being collected since training samples can originate from a variety of sources, such as human operators \citep{mandlekar2020learning}, pre-existing policies \citep{hester2018deep}, or logged experiences from different agents \citep{fujimoto2019off,kostrikov2021offline}.

In \method, we adopt a simplified approach by focusing on datasets generated solely from RL policies, using energy consumption as our training sample cost metric. This design choice enables systematic evaluation while avoiding the complexities of collecting and pricing human demonstrations. Specifically, we define the training sample cost, $C_D$, of a dataset $D$ as the average energy consumed to train the policies that are used to generate the dataset $D$. This can be expressed as:
\begin{align}
C_D = \frac{1}{|\Pi_D|} \sum_{\pi \in \Pi_D} E_{\text{train}}(\pi)
\end{align}where $\Pi_D$ is the set of policies used to generate the dataset $D$, $|\Pi_D|$ denotes the number of policies in this set, and $E_{\text{train}}(\pi)$ represents the energy consumed to train the policy $\pi$.
As we strive for more equitable comparisons between approaches to training autonomous agents, we urge the research community to consider the cost of acquiring training data. To this end, we release datasets for each domain and task in \method, along with their associated training sample costs.
While the specific expertise levels may vary across domains and tasks, we generally consider three categories: \texttt{novice}, \texttt{intermediate}, and \texttt{expert}.
See Appendix \ref{appendix:dataset_collection} for the dataset collection procedure and Appendix \ref{appendix:dataset_information} for details on the dataset format.

\subsection{System Performance}
\label{ssec:system_performance}
System metrics provide insight into the feasibility of deploying autonomous agents, particularly considering the scaling demands on energy and data efficiency \citep{frey2022benchmarking}. \method uses the CodeCarbon library \citep{codecarbon} to track metrics during training, such as energy usage, power draw, RAM consumption, and wall-clock time. Energy and power usage inform the user about the sustainability and costs associated with training the agent, which is particularly important in power-constrained environments or when planning for long-term, continuous training \citep{PARISI201954}. RAM consumption metrics help in understanding the memory efficiency of the training process, as high RAM consumption may limit the settings where the agent can be trained or require costly hardware upgrades \citep{li2023breaking}. During the inference phase, \method records power draw, RAM consumption, and average inference time.


Given that system performance metrics can vary substantially across hardware configurations and software environments, reporting detailed experimental setup information is crucial for reproducibility.
When using \method, researchers should specify their deep learning framework, CPU and GPU models, and Python version to enable meaningful comparisons.
We provide a guideline for reporting results in Section \ref{ssec:community_benchmarking}, and our own experimental configuration is detailed in Appendix \ref{appendix:experimental_setup}. This standardized reporting approach ensures that the research community can accurately interpret and build upon published results.

\subsection{Reliability}
\label{ssec:reliability}

\renewcommand{\arraystretch}{1.2}
\begin{table}[h]
\centering
\small
\resizebox{0.95\linewidth}{!}{
\begin{tabular}{|c|c|>{\centering\arraybackslash}p{6cm}|>{\centering\arraybackslash\hspace{0.5em}}m{4.5cm}<{\hspace{0.5em}}|}
\hline
\textbf{Phase} & \textbf{Metric Name} & \centering \textbf{Description} & \textbf{Equation} \\
\hline
\multirow{5}{*}{\raisebox{-22ex}{\rotatebox[origin=c]{90}{\textbf{Training}}}} 
& Dispersion Within Runs 
& Measures higher-frequency variability using IQR within a sliding window along the detrended training curve. Lower values indicate more stable performance.
& $\displaystyle \frac{1}{T-4} \sum_{t=3}^{T-2} \text{IQR}\left(\{\Delta P_{t'}\}_{t'=t-2}^{t+2}\right)$ \\ \cline{2-4}
& Short-term Risk (CVaR) 
& Estimates extreme short-term performance drops. Lower values indicate less risk of sudden drops.
& $\displaystyle\text{CVaR}_\alpha \left(\Delta P_t\right)_{t=1}^T$ \\ \cline{2-4}
& Long-term Risk (CVaR) 
& Captures potential for long-term performance decrease. Lower values indicate less risk of degradation.
& $\displaystyle\text{CVaR}_\alpha \left(\max_{t' \leq t} P_{t'} - P_t\right)$ \\ \cline{2-4}
& Dispersion Across Runs 
& Measures variance across training runs. Lower values indicate more consistent performance across runs.
& $\displaystyle \frac{1}{T} \sum_{t=1}^{T} \text{IQR}\left(\{P_{t,j}\}_{j=1}^{n}\right)$ \\ \cline{2-4}
& Risk Across Runs (CVaR) 
& Measures expected performance of worst-performing agents. Higher values indicate better worst-case performance.
& $\displaystyle\text{CVaR}_\alpha \left(P_{T,j}\right)_{j=1}^n$ \\ \cline{1-4}
\multirow{2}{*}{\raisebox{-7ex}{\rotatebox[origin=c]{90}{\textbf{Inference}}}} 
& Dispersion Across Rollouts 
& Measures variability in performance across multiple rollouts. Lower values indicate more consistent performance.
& $\displaystyle\text{IQR}\left(R_i\right)_{i=1}^m$ \\ \cline{2-4}
& Risk Across Rollouts (CVaR) 
& Measures worst-case performance during inference. Higher values indicate better worst-case performance.
& $\displaystyle\text{CVaR}_\alpha \left(R_i\right)_{i=1}^m$ \\ \hline
\end{tabular}
}
\vspace{0.5em}
\caption{Reliability Metrics from \citet{chan2019measuring} with Mathematical Formulations. 
$P_t$: performance at time $t$. 
$P_{t,j}$: performance at time $t$ for run $j$. 
$R_i$: performance during rollout $i$. 
$\Delta P_t = P_t - P_{t-1}$: performance change between consecutive time steps (detrended value). 
$\text{CVaR}_\alpha$\protect\footnotemark: Conditional Value at Risk at level $\alpha$. 
IQR: Inter-Quartile Range. 
Sliding window length is 5 time steps centered on $t$, calculated over all $t$ from 3 to $T-2$ to ensure the window is valid. 
$T$: total number of time steps. 
$n$: number of runs (10 for our experiments). 
$m$: number of rollouts (100 for our experiments).}
\label{tab:reliability}
\end{table}
\footnotetext{\url{https://en.wikipedia.org/wiki/Expected_shortfall}}


Reliability signifies safety, accountability, reproducibility, stability, and trustworthiness \citep{chan2019measuring, roszel2021know}.
\method uses the statistical methods proposed by \citet{chan2019measuring} to measure the reliability of autonomous agents during training and inference.
During training, \method examines dispersion across multiple training runs, dispersion over time within a single run, risk across runs, and risk over time.
These metrics provide insights into the variability and worst-case performance of the agent. For example, low dispersion across training runs indicates that the algorithm consistently achieves similar performance regardless of random initialization, while low risk metrics suggest the agent avoids catastrophic performance drops.
For inference, \method measures dispersion and risk across rollouts to assess the consistency and potential suboptimal performance of the final trained agent.
Table \ref{tab:reliability} provides an overview of the reliability metrics tracked by \method, along with how they should be interpreted.
For a detailed description of each metric and their calculation, please refer to the work by \citet{chan2019measuring}.

\subsection{Application Performance}
\label{ssec:application_performance}

Application performance is measured using task performance and generalization.
Task performance is the agent's mean returns when rolled out for $100$ episodes on the task it was trained for.
Since autonomous agents deployed in real-world settings must often handle scenarios that differ from their exact training distribution, measuring generalization to tasks outside this distribution is crucial.
Generalization is computed as the sum of mean returns for all tasks, including the task the agent was trained to perform.

\subsection{Using \method Metrics in Practice}
\label{ssec:metric_selection}

The metrics provided by \method across data cost, application performance, system performance, and reliability offer a holistic view of an agent's performance.
However, the relative importance of these metrics can vary significantly depending on the specific application domain.
For instance, in resource-constrained environments, system performance metrics may be critical, while in safety-critical applications, reliability metrics might take precedence.
In Section \ref{sec:evaluation}, we demonstrate how these metrics can be applied and interpreted in the context of our three benchmark domains: computer chip floorplanning, web navigation, and quadruped locomotion.
\subsection{Community Benchmarking with \method}
\label{ssec:community_benchmarking}
While system performance metrics like energy usage, inference time, and memory consumption can vary significantly across different hardware platforms and software implementations, these measurements become meaningful when properly contextualized. To facilitate fair and useful comparisons, \method will include a community leaderboard where researchers must report:

\begin{itemize}
\item \textbf{Hardware Configuration:}
\begin{itemize}
\item CPU model
\item GPU model
\end{itemize}

\item \textbf{Software Environment:}
\begin{itemize}
\item Deep learning framework (e.g., PyTorch \citep{paszke2019pytorch}, Tensorflow \citep{abadi2016tensorflow}, Jax \citep{jax2018github}, etc.)
\item Python Version
\item Operating System
\end{itemize}

\item \textbf{Metric Results:}
\begin{itemize}
\item Data Cost
\item System Performance
\item Reliability
\item Application Performance
\end{itemize}

\item \textbf{Experimental Details:}
\begin{itemize}
\item Number of random seeds used
\item All Hyperparameter Settings
\end{itemize}

\end{itemize}

To facilitate this standardized reporting and obtain the metric results above, researchers can leverage \method's easy-to-use, open-source codebase.\footnote{The A2Perf codebase is available at \url{https://a2perf.farama.org/}}
The codebase includes detailed tutorials, examples, and docker containers that simplify the evaluation process.
Its modular implementation also allows users to integrate their own custom algorithms without needing to modify the benchmarking code.

By standardizing the reporting of system configurations, researchers can meaningfully compare results across similar hardware and software setups, providing insights into how different agents perform under comparable conditions.
The community leaderboard also enables understanding of performance scaling across different platforms, from resource-constrained environments to high-performance systems.
Furthermore, practitioners can use this information to make informed decisions about deployment requirements and track optimization progress for specific hardware targets.

For example, researchers deploying a quadruped with a specific compute stack could filter the leaderboard entries to find results from comparable system configurations.
As the community contributes results to \method, this repository of performance data will expand across many computing environments, providing comprehensive coverage of different configurations.

%% file: domain_background.tex
\newcommand{\cmark}{\textcolor{green!80!black}{\ding{51}}}
\newcommand{\xmark}{\textcolor{red}{\ding{55}}}

\begin{table}[t!]
    \small
    \centering
    \begin{tabularx}{\textwidth}{>{\raggedright\arraybackslash}Xccc}
        \toprule
        \textbf{Real-World Challenges} & \makecell{Chip \\Floorplanning} & \makecell{Web \\Navigation} & \makecell{Quadruped \\Locomotion}\\ 
        \midrule
        \textbf{(RW1)\textsuperscript{*}} Training offline from fixed logs. & \cmark & \cmark & \cmark \\
        \addlinespace
        \textbf{(RW2)} Learning on the real system from limited samples. & \xmark & \xmark & \cmark \\ 
        \addlinespace
        \textbf{(RW3)} High-dimensional and continuous state and action spaces. & \cmark & \xmark & \cmark \\ 
        \addlinespace
        \textbf{(RW4)} Safety constraints. & \xmark & \cmark & \cmark \\ 
        \addlinespace
        \textbf{(RW5)} Tasks are partially observable, non-stationary or stochastic. & \xmark & \xmark & \cmark \\ 
        \addlinespace
        \textbf{(RW6)} Unspecified, multi-objective or risk sensitive reward functions. & \cmark & \cmark & \cmark \\
        \addlinespace
        \textbf{(RW7)} Need for explainable policies. & \xmark & \cmark & \xmark \\ 
        \addlinespace
        \textbf{(RW8)} Real-time inference at the control frequency of the system. & \xmark & \cmark & \cmark \\ 
        \addlinespace
        \textbf{(RW9)} Delays in actuators, sensors or rewards. & \xmark & \cmark & \cmark  \\ 
        \bottomrule
    \end{tabularx}
    \vspace{0.5em}
    \caption{Real-World Challenges proposed by \citeauthor{dulac2021challenges} \citep{dulac2021challenges}. Checkmarks (\cmark) indicate challenges commonly encountered in the general domain area, while (\xmark) denotes challenges less frequently encountered.
    The challenge marked with an asterisk (*), RW1, applies to all \method domains, as learning from offline data is possible for all environments.
    Each broad challenge is encountered in at least one of the \method domain areas, highlighting the relevance of the selected domains to current real-world reinforcement learning problems.}
    \label{tab:dulac_real_world_challenges}
\end{table}


Our guiding question when selecting domains for \method was ``how can we choose domains that reflect real-world applications of autonomous agents?''
To identify suitable domains, we conducted interviews with industry practitioners to understand where autonomous agents are currently deployed and where they show future promise.
This process led us to three application areas with significant industrial relevance: computer chip floorplanning, quadruped locomotion, and website navigation.

From these industrially relevant application areas, we specifically selected domains with demonstrated simulation-to-reality transfer. This selection criterion enables researchers without access to specialized hardware (like robots or chip fabrication facilities) to make meaningful contributions using simulated environments. The circuit training domain was used in creating an iteration of Google's Tensor Processing Unit (TPU) \citep{mirhoseini2021graph}. The quadruped locomotion domain has been shown to transfer successfully to real Unitree Laikago robots \citep{peng2020learning_agile_imitate}. 
The web navigation domain is derived from MiniWob \citep{shi2017world}, MiniWob++ \citep{liu2018reinforcement}, and gMiniWob \citep{gur2021environment}, and operates in an actual Google Chrome browser, mirroring real-life web interactions. Additionally, \cite{gur2018learning_to_navigate_the_web} showed that policies trained in MiniWob++ transfer to real-life web pages for task completion.

By focusing on domains with demonstrated real-world applicability, progress made within the \method benchmark can directly contribute to improving the performance of downstream real-world (RW) tasks.
We specify how each domain aligns with the real-world challenges presented by \citet{dulac2021challenges} (Table \ref{tab:dulac_real_world_challenges}), and denote which of \method's metric categories are important for each domain.

\subsection{Circuit Training (RW1, RW3, RW6)}

Chip floorplanning involves creating a physical layout for a microprocessor, a task that has resisted automation for decades and requires months of human engineering effort. To address this challenge, Google has made Circuit Training available as an open-source framework that uses RL to generate chip floorplans~\citep{CircuitTraining2021}.
In this domain, an agent places macros (reusable blocks of circuitry) onto the chip canvas, with the objective of optimizing wirelength, congestion, and density.
Even though the state and action spaces are discrete, the number of states and actions increases combinatorially with the number of nodes and cells on the chip (RW3).
As an illustration, \citet{mirhoseini2021graph} calculate that placing 1,000 clusters of nodes on a grid with 1,000 cells results in a state space on the order of $10^{2,500}$, which is vastly larger than the state space of Go at $10^{360}$.
Chip design also involves optimizing for multiple objectives, such as maximizing clock frequency, reducing power consumption, and minimizing chip area (RW6). During training, these objectives are approximated using proxy metrics.
However, evaluating the true objectives requires time-consuming simulations with industry-grade placement tools \footnote{For example, \href{https://www.cadence.com/en_US/home/tools/digital-design-and-signoff/soc-implementation-and-floorplanning/innovus-implementation-system.html}{Cadence Innovus} and \href{https://www.synopsys.com/implementation-and-signoff/physical-implementation/ic-compiler.html}{Synopsys IC Compiler}}.
If the results are unsatisfactory, the proxy metrics must be adjusted, and the agents must be retrained, leading to a costly iterative and resource-intensive process.


\paragraph{Important Metric Categories}
For Circuit Training agents, the following metric categories are most critical for real-world use:

\begin{itemize}
    \item \textbf{Task Performance}: Circuit Training agents must generate high-quality macro placements by minimizing wirelength, congestion, and density of the chip.
    
    \item \textbf{Inference Reliability}: Chip designers use these agents to generate initial macro placements, then manually refine them. Agents must produce consistent macro placements across multiple rollouts. Inconsistent placements would force designers to repeatedly roll out the same policy to try achieving favored initial placements.
    
    \item \textbf{Inference System Performance}: Fast inference time is crucial to enable interactive use by human designers. Designers need to quickly evaluate and refine different placement options.
    
    \item \textbf{Generalization}: The ability to handle new circuit architectures without retraining is vital, as new circuits are frequently created. Strong generalization performance reduces the need to train separate agents for each new netlist.
    
    \item \textbf{Data Cost}: Many circuit netlists are proprietary, and generating high-quality macro placements requires significant human effort. Understanding data collection costs helps evaluate the practicality of different learning approaches.
\end{itemize}


\subsection{Web Navigation (RW1, RW4, RW6, RW7, RW8, RW9)}
\vspace{-0.5em}
Software tools exist to automate browser tasks\footnote{\href{https://www.selenium.dev/documentation/webdriver/}{Selenium}, used in \method, is a popular browser automation tool.}, but due to the varied formatting of websites, hand-crafted algorithms are not a viable solution for general web navigation.
Researchers have begun applying learning algorithms to design agents that can understand web pages \citep{gur2022understanding_html_llm} and automatically navigate through them to fill out forms \citep{gur2021environment, gur2018learning_to_navigate_the_web}.
In \method, we use gMiniWob \cite{gur2021environment} to create mock websites that act as environments for the agent.
See Appendix \ref{appendix:website_generation} for details about the website generation process and agent interaction.
To achieve maximum rewards, the agent must avoid malicious links and advertisement banners (RW4) while correctly filling out all fields in web forms.
The combination of these constraints create a multi-objective reward function (RW6).
The explainability of an agent's decision-making is also important, particularly when agents handle sensitive tasks such as online shopping or investing (RW7).
Finally, agents must be robust to the system challenges of real-time inference, such as inference speed and network delays (RW8, RW9).


\paragraph{Important Metric Categories}
For web navigation agents, the following metric categories are most critical for real-world use:
\begin{itemize}
    \item \textbf{Task Performance}: Agents must accurately complete web forms and navigate sites correctly. 
    
    \item \textbf{Inference System Performance}: Agents need to operate at speeds comparable to human web browsing to provide a seamless user experience. This includes both inference time and resource usage on consumer devices.
    
    \item \textbf{Inference Reliability}: Reliability is crucial for safety, as unreliable agents might occasionally click on malicious links or advertisements. Even rare mistakes in web navigation can have serious consequences.
    
    \item \textbf{Generalization}: Websites vary greatly in design and structure. Agents must adapt to different layouts, styles, and interaction patterns without requiring retraining for each new site.
    
    \item \textbf{Training System Performance}: Web navigation training involves processing HTML pages and running multiple browser instances, creating significant computational demands. 
\end{itemize}

\renewcommand{\arraystretch}{1.0}
\begin{table}[!t]
\resizebox{\textwidth}{!}{%
\begin{tabular}{|l|l|c|c|c|}
\toprule
\multicolumn{5}{|c|}{\textbf{Ariane (Training)}} \\
 &  & \textbf{BC} & \textbf{DDQN} & \textbf{PPO} \\
\textbf{Category} & \textbf{Metric Name} &  &  &  \\
\midrule
\multirow[t]{1}{*}{Data Cost} & Training Sample Cost & 48.28 & 0 & 0 \\
\midrule
\multirow[t]{2}{*}{Application} & Generalization (100 eps. [all tasks]) & -2.18 & -2.19 & -2.05 \\
 & Returns (100 eps.) & -1.10 $\pm$ 0.04 & -1.13 $\pm$ 0.04 & -0.99 $\pm$ 7.25e-03 \\
\cline{1-5}
\multirow[t]{5}{*}{Reliability} & Dispersion Across Runs (IQR) & N/A & 0.03 $\pm$ 0.03 & 0.04 $\pm$ 0.02 \\
 & Dispersion Within Runs (IQR) & N/A & 0.02 $\pm$ 0.03 & 4.77e-03 $\pm$ 4.92e-03 \\
 & Long Term Risk (CVaR) & N/A & 1.20 & 0.03 \\
 & Risk Across Runs (CVaR) & N/A & -1.17 & -1.03 \\
 & Short Term Risk (CVaR) & N/A & 0.07 & 0.01 \\
\cline{1-5}
\multirow[t]{5}{*}{System} & Energy Consumed (kWh) & 0.11 $\pm$ 6.45e-04 & 108.20 $\pm$ 4.29 & 120.53 $\pm$ 2.78 \\
 & GPU Power Usage (W) & 211.35 $\pm$ 16.76 & 585.98 $\pm$ 172.50 & 692.94 $\pm$ 120.08 \\
 & Mean RAM Usage (GB) & 4.72 $\pm$ 0.53 & 849.37 $\pm$ 64.85 & 834.05 $\pm$ 55.90 \\
 & Peak RAM Usage (GB) & 5.25 $\pm$ 0.07 & 889.56 $\pm$ 23.44 & 906.45 $\pm$ 68.01 \\
 & Wall Clock Time (Hours) & 0.48 $\pm$ 2.61e-03 & 21.94 $\pm$ 0.90 & 23.95 $\pm$ 0.54 \\
\midrule
\multicolumn{5}{|c|}{\textbf{Ariane (Inference)}} \\
\multirow[t]{2}{*}{Reliability} & Dispersion Across Rollouts (IQR) & 0.01 & 0.05 & 0.01 \\
 & Risk Across Rollouts (CVaR) & -1.23 & -1.25 & -1.01 \\
\cline{1-5}
\multirow[t]{4}{*}{System} & GPU Power Usage (W) & 136.91 $\pm$ 21.48 & 69.50 $\pm$ 4.60 & 49.43 $\pm$ 30.29 \\
 & Inference Time (ms) & 10.0 $\pm$ 0.46 & 20.0 $\pm$ 2.69 & 20.0 $\pm$ 2.68 \\
 & Mean RAM Usage (GB) & 2.19 $\pm$ 0.21 & 2.15 $\pm$ 0.30 & 2.51 $\pm$ 0.49 \\
 & Peak RAM Usage (GB) & 2.29 $\pm$ 0.01 & 2.28 $\pm$ 0.13 & 2.71 $\pm$ 0.62 \\
\bottomrule
\end{tabular}
}
\vspace{0.5em}
\caption{Metrics for the Ariane Netlist task of CircuitTraining-v0. All metrics are averaged over ten random seeds. We report mean and standard deviation for metrics where it is applicable. BC results are obtained by training on the entire \texttt{intermediate} dataset. Note that the training reliability metrics for BC are marked as ``N/A'' since BC does not perform online rollouts in the environment.}
\label{tab:metrics_circuit_training_ariane}
\vspace{-1em}
\end{table}

\subsection{Quadruped Locomotion (RW1, RW2, RW3, RW4, RW5, RW6, RW8, RW9)}
\vspace{-0.5em}
In recent years, the robotics community has gradually shifted towards training autonomous agents for robotic control.
A prominent example of this trend is seen in quadruped locomotion, where RL has become the dominant technique.
We followed the work of \citet{peng2020learning_agile_imitate}, in which a quadruped robot learns complex locomotion skills such as pacing, trotting, spinning, hop-turning, and side-stepping by imitating motion capture data from a real dog.

Given the physical dynamics involved in quadruped locomotion, research often necessitates learning directly from limited samples on the actual robot (RW2).
Learning walking gaits also involves high-dimensional, continuous state and action spaces (RW3), as the robot needs to precisely control multiple joints and limbs to navigate complex environments. The agent must reason about complex dynamics, avoid unsafe falls (RW4), adapt gaits to various speeds and terrains (RW5), and operate in partially observable environments (RW5) where states like contact forces are not directly measurable.
Optimizing robotic controllers is usually multi-objective (RW6), balancing competing objectives like locomotion speed, stability, satisfying safety constraints, and minimizing energy expenditure. Furthermore, real-time inference (RW8) and dealing with system delays (RW9) are critical for controlling robots, as slow computations or delays can negatively impact stability and performance.


\paragraph{Important Metric Categories}
For quadruped locomotion agents, the following metric categories are most critical for real-world use:

\begin{itemize}
    \item \textbf{Task Performance}: Agents must accurately reproduce desired walking gaits, as poor imitation of natural movements can lead to inefficient or unstable locomotion.
    
    \item \textbf{Inference Reliability}: The agent must maintain smooth, stable motions without sudden movements or changes in behavior. Inconsistent movements could damage the robot or cause falls in real-world environments.
    
    \item \textbf{Inference System Performance}: Quadrupeds require real-time responsiveness from their onboard computers to maintain stability. Both inference speed and energy efficiency are crucial, as robots often operate with limited computing resources and battery power.
    
    \item \textbf{Generalization}: Robots must adapt to different terrains, slopes, and surface conditions without retraining. Strong generalization also helps robots handle variations in their own morphology due to wear or manufacturing differences.
\end{itemize}

%% file: evaluation.tex
Choosing an agent for real-world applications requires understanding the costs and resources needed for training and deployment, as well as the tradeoffs between different algorithms. To this end, our evaluation aims to answer three key questions. (\textbf{Q1}) How can data cost metrics be used in practice to compare methods that use offline data to those that do not? (\textbf{Q2}) How do system performance metrics inform training and deployment feasibility? (\textbf{Q3}) Can reliability metrics reveal tradeoffs between different agents that are not captured by raw task performance?

Our choice of baselines is guided by the action space of each domain. Circuit training and web navigation have discrete action spaces, so we evaluate them using DDQN and PPO, while quadruped locomotion has a continuous action space, so we use PPO and SAC. BC is included across all domains as an imitation learning baseline. For all domains and tasks, results are averaged over ten random seeds to ensure robustness and reproducibility. See Appendix \ref{appendix:additional_experiments} for more experimental results.

\subsection{\textbf{Q1}: Comparing Across Algorithm Types with Data Cost}\vspace{-0.2cm}
\textbf{\textit{How can data cost metrics be used in practice to compare methods that use offline data to those that do not?}}

\begin{figure}[!t]
    \centering
    \includegraphics[width=1.0\linewidth]{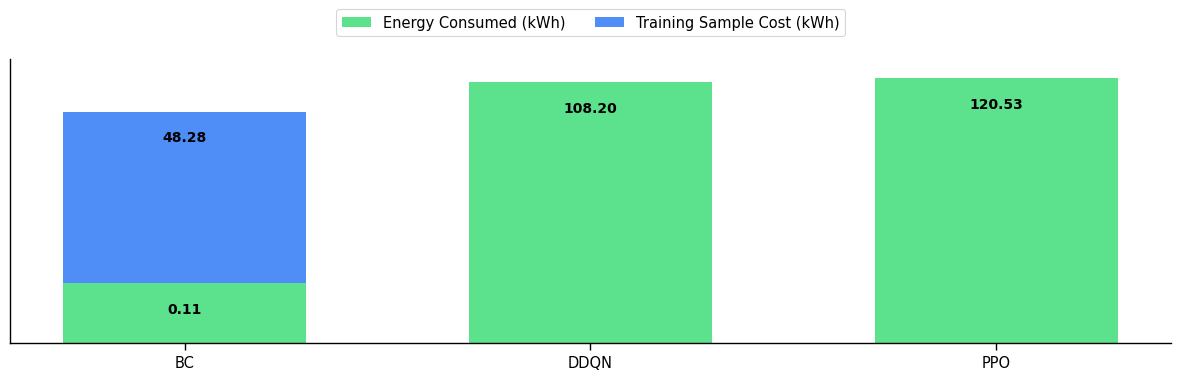}
    \caption{Comparison of energy consumption and training sample cost for BC, DDQN, and PPO on the Ariane Netlist task, enabled by \method. \textbf{Note:} The plot is not to scale for visibility of smaller values. Online methods (DDQN and PPO) have no training sample cost as they are initialized without pre-collected data. BC's energy consumption (0.11 kWh) is significantly smaller than its training sample cost (48.28 kWh), which represents the energy used to generate the training data.}
    \label{fig:ariane_energy_comparison}
\end{figure}

\method provides datasets generated with agents of varying expertise (Section \ref{sec:training_sample_cost}), along with their associated training sample costs.
This enables the comparison of agents by considering both task performance and the cost of acquiring training data, which can vary significantly across different approaches like IL and RL.

Our experiments in the chip floorplanning domain reveal important insights about the true costs of different approaches. While BC's performance is competitive with DDQN and PPO (Table \ref{tab:metrics_circuit_training_ariane}), the training sample cost -- measured as the average energy consumed to train an agent that generates the data -- was $48.28$ kWh. In contrast, online methods like DDQN and PPO learn purely through environment interaction without requiring any pre-collected datasets, resulting in a training sample cost of zero.

The data cost metric allows researchers to combine the training sample cost with the energy consumed during training for a more comprehensive comparison. This approach provides a total energy cost that can be directly compared across offline, online, or hybrid methods. For example, offline training of a BC agent for the Ariane netlist consumed only 0.11 kWh. Therefore, the total energy cost for a BC agent would be $48.39$ kWh ($48.28$ kWh for generating the offline data + $0.11$ kWh for offline training).

When comparing total energy costs, we find that despite requiring pre-collected data, BC's total energy cost ($48.39$ kWh) is still lower than the energy consumed by online methods like DDQN and PPO, which amounted to $108.20$ kWh and $120.53$ kWh, respectively (Figure \ref{fig:ariane_energy_comparison}). For hybrid methods that use both offline data and online environment interactions, the total energy cost would similarly be calculated by adding the training sample cost for the offline data to the energy consumed during the online training phase.

These findings illustrate how data cost metrics can fundamentally change our understanding of algorithm efficiency. Without accounting for training sample costs, offline methods like BC would appear dramatically more efficient than online methods, potentially leading to misguided algorithm selection decisions. For domains with expensive data collection, like chip floorplanning where expert demonstrations may require considerable effort, the training sample cost metric becomes essential for fair comparisons. In contrast, for domains where high-quality data is readily available or inexpensive to collect, traditional online methods might remain preferable despite their higher training energy costs. 

\subsection{\textbf{Q2}: System Performance for Training and Deployment Feasibility}\vspace{-0.2cm}
\textbf{\textit{How do system performance metrics inform training and deployment feasibility?}}

\begin{figure}[!t]
    \centering
    \begin{subfigure}[b]{0.48\textwidth}
        \centering
        \includegraphics[width=\textwidth]{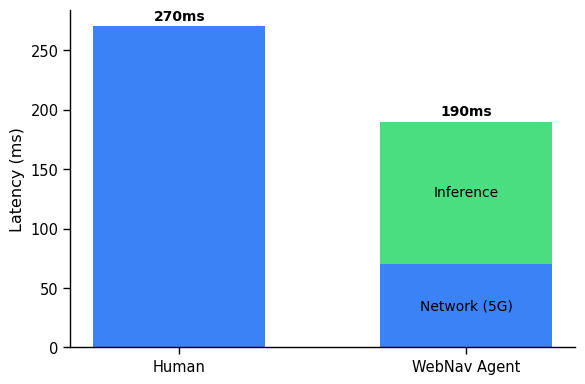}
        \caption{Comparison of web navigation agent latency with human reaction time. Agents are fast enough for real-time form-filling tasks, even when served from the cloud.}
        \label{fig:webnav-agent-latency}
    \end{subfigure}
    \hfill
    \begin{subfigure}[b]{0.38\textwidth}
        \centering
        \includegraphics[width=\textwidth]{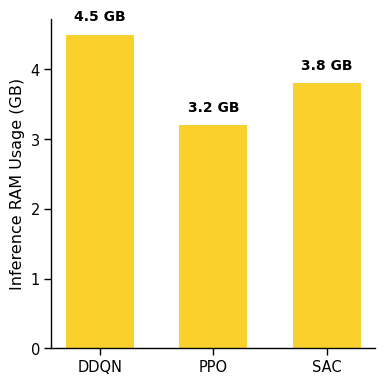}
        \caption{Peak RAM usage for quadruped locomotion agents at inference time.}
        \label{fig:locomotion-ram-usage}
    \end{subfigure}
    \caption{System performance metrics for (a) Web Navigation agents and (b) Quadruped Locomotion agents. The latency comparison demonstrates the feasibility of real-time web interaction, while RAM usage highlights the resource requirements for deploying quadruped locomotion agents.}
    \label{fig:system-performance}
\end{figure}

Our experiments in the web navigation domain highlight the importance of considering hardware constraints and performance requirements of autonomous agents.
During training, PPO agents had a peak RAM usage of $2.3 \pm 0.14$ TB (Appendix Table \ref{tab:appendix_metrics_webnav_difficulty_1_websites_1}).
This high memory footprint can be attributed to the need for distributed experiments running hundreds of Google Chrome processes and storing batches of data, which involves tokenizing the entire DOM\footnote{\url{https://en.wikipedia.org/wiki/Document_Object_Model}} tree of HTML elements on each web page.
Such memory demands can limit the accessibility of training agents, as not all researchers may have access to the necessary hardware resources.
To put this into perspective, training a variant of the GPT-3 language model with approximately 72 billion parameters would require a similar amount of memory, assuming each parameter is stored as a 32-bit floating-point number \citep{brown2020language}.

However, the resource usage of these agents becomes more manageable for deployment.
The 120~ms inference time, when combined with the median round-trip latency of $\sim$68 ms for a 5G network \citep{schafhalter2023leveraging}, results in a total latency of $\sim$200 ms. This combined latency is still faster than the average human reaction time of $\sim$273 ms\footnote{\url{https://humanbenchmark.com/tests/reactiontime/statistics}}, enabling real-time responsiveness during web navigation tasks (Figure \ref{fig:webnav-agent-latency}).
Furthermore, the peak RAM usage of $2.19 \pm 0.09$ GB (Table \ref{tab:appendix_metrics_webnav_difficulty_1_websites_1}) indicates the feasibility of deploying trained agents directly on consumer-grade devices, such as smartphones, though the inference time may be slower on-device.

System performance metrics reveal critical practical constraints that might otherwise be overlooked during algorithm development.
The substantial gap between training and deployment resource requirements across our experiments demonstrates why evaluating both phases is essential. For web navigation agents, the extremely high training RAM requirements suggest that specialized infrastructure or algorithmic optimizations might be needed to make training more accessible to researchers with limited resources. Yet the reasonable inference requirements indicate these agents can be widely deployed once trained. This pattern (resource-intensive training but efficient inference) is common across many domains and highlights the importance of system performance metrics in guiding resource allocation decisions.
By explicitly measuring these metrics, \method helps researchers anticipate deployment constraints and identify potential bottlenecks early in the development process, facilitating more practical and deployable autonomous agents.


\subsection{\textbf{Q3}: Finding Tradeoffs with Reliability Metrics}\vspace{-0.2cm}
\textbf{\textit{Can reliability metrics reveal tradeoffs between different agents that are not captured by raw task performance?}}

\begin{figure}[!t]
    \centering
    \begin{subfigure}[b]{0.48\textwidth}
        \centering
        \includegraphics[width=\textwidth]{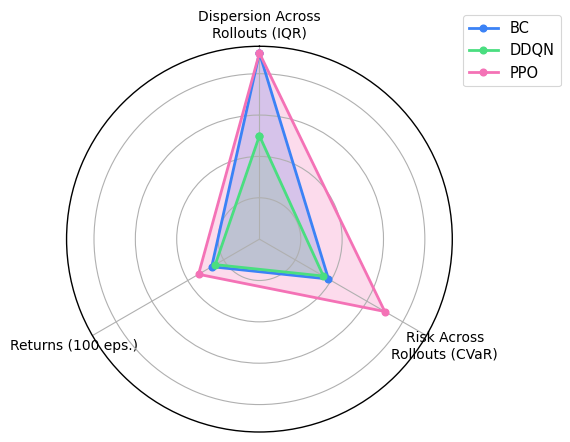}
        \caption{Reliability metrics for chip floorplanning algorithms during inference on the Ariane netlist task. While task performance is similar between PPO and DDQN, PPO demonstrates better reliability metrics, providing a more consistent and predictable experience for human chip designers working with the generated layouts.}
        \label{fig:ariane-reliability}
    \end{subfigure}
    \hfill
    \begin{subfigure}[b]{0.48\textwidth}
        \centering
        \includegraphics[width=\textwidth]{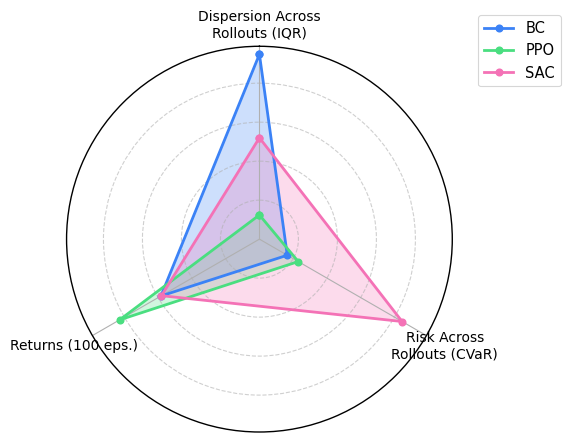}
        \caption{Reliability metrics for quadruped locomotion algorithms during inference on the dog pace task. SAC shows notably better performance, achieving 3.7x better results than PPO in worst-case rollouts and demonstrating more consistent performance at inference time with 1.8x improvement in dispersion across rollouts.}
        \label{fig:dogpace-reliability}
    \end{subfigure}
    \caption{Comparison of reliability metrics across different domains and algorithms. The radar plots illustrate how different algorithms trade off between task performance and reliability.}
    \label{fig:reliability-comparison}
\end{figure}

Computer chip designers using autonomous agents rely on the agent to generate initial placements that they can build upon, so minimizing variability in the agent's performance is crucial.
As shown in Table \ref{tab:metrics_circuit_training_ariane}, the PPO algorithm exhibited lower dispersion across rollouts (IQR of \num{0.01}) compared to DDQN (IQR of \num{0.05}), indicating that PPO is approximately 5x more stable than DDQN when rolling out fixed, trained policies (Figure \ref{fig:ariane-reliability}).
This suggests that PPO would provide more consistent starting points for designers, enabling them to focus on refining and optimizing the floorplan instead of repeatedly rolling out the same policy to get similar initial placements.
Additionally, PPO demonstrated lower risk across rollouts (CVaR of $-1.01$) compared to DDQN (CVaR of $-1.25$), indicating that in the worst-performing rollouts, PPO performs about 1.2x better than DDQN on average, reducing the likelihood of designers starting with poor floorplans that require extensive manual adjustments.

In analyzing the ``Dog Pace" task of QuadrupedLocomotion-v0 (Table \ref{tab:appendix_locomotion_dog_pace}), we observe overlapping error bars on the returns for PPO and SAC.
To better understand their tradeoffs, we use the reliability metrics.
PPO provides a 2x reduction in both short-term and long-term risks compared to SAC, making PPO more stable.
This stability potentially makes PPO a safer option for training quadrupeds in the real world, where less sporadic behavior is needed.
Conversely, SAC performs 3.7x better than PPO in the worst-case rollouts on average and demonstrates a 1.8x improvement in dispersion across rollouts, indicating more consistent gaits during deployment -- essential from a safety perspective (Figure \ref{fig:dogpace-reliability}).

These reliability metrics provide insights that would be missed if we only considered task performance. For quadruped locomotion, the choice between PPO and SAC presents a clear tradeoff: PPO offers stability during training that could benefit researchers developing new control methods, while SAC provides more consistent performance during deployment that is crucial for real-world robot applications where unpredictable behaviors could damage hardware or create unsafe conditions. Similarly, in chip design, reliability metrics reveal PPO's advantage in producing consistent layouts—an attribute that significantly impacts user experience but isn't captured in typical performance metrics. Across domains, these findings demonstrate that reliability metrics are not merely statistical tools but practical indicators that directly relate to user experience, hardware safety, and development efficiency.
By evaluating algorithms through this lens, researchers can better align algorithm selection with the specific reliability requirements of their application domain.

\subsection{ Applying Metrics to Guide Real-World Algorithm Selection}\vspace{-0.2cm}

Our experimental evaluation across three domains demonstrates how different metrics in \method reveal complementary aspects of autonomous agent performance that are essential for real-world deployment decisions. While task performance provides a necessary baseline for comparison, the additional dimensions of data cost, system performance, and reliability metrics offer crucial insights for practitioners making algorithm selection decisions.

The relative importance of different metrics varies significantly by domain, reflecting different priorities in real-world applications.
For chip floorplanning, reliability metrics revealed PPO's advantage in providing consistent initial layouts—a property not evident from task performance alone but critical for human designers who need predictable starting points.
In web navigation, system performance metrics demonstrated that while training requires substantial compute resources, inference can be performed efficiently enough for real-time web interaction.
For quadruped locomotion, reliability metrics exposed a fundamental tradeoff between PPO's training stability and SAC's deployment stability that would be entirely missed by examining only average returns.

These findings illustrate \method's value as a comprehensive evaluation framework that enables informed algorithm selection based on application-specific priorities. Researchers developing new autonomous agent algorithms should consider which metrics matter most for their target domains: data-intensive applications may prioritize training sample cost, resource-constrained deployments might emphasize inference efficiency, safety-critical systems would focus on reliability metrics, and applications requiring adaptability would value generalization performance. By providing this multidimensional perspective, \method helps bridge the gap between algorithm development and successful real-world deployment of autonomous agents.

%% file: discussion.tex
\method includes three domains that cover a diverse range of real-world applications and challenges, but there is room for expansion to a wider range of tasks. Thanks to \method's integration with Gymnasium \citep{towers_gymnasium_2023} (previously OpenAI Gym) and the implementation of baselines using TF-Agents \citep{TFAgents}, adding new domains and baselines is straightforward, making it easy for researchers to contribute to the platform.

Future work could expand \method to include multi-agent domains and tasks, reflecting real-world scenarios where autonomous agents interact with other agents and humans. Many real-world applications inherently involve multiple agents coordinating or competing: autonomous vehicles navigating in traffic, robot teams in warehouse settings, or trading agents in financial markets. Integrating multi-agent domains would require additional metrics to capture interaction dynamics, such as coordination efficiency, communication overhead, and emergent social behaviors. For example, extending the quadruped locomotion domain to include multiple robots collaboratively navigating complex terrain would test both individual control and collective coordination capabilities, potentially revealing new insights about algorithm robustness in social contexts.

Another area of future work is the addition of support for measuring system performance on custom hardware platforms.
This would provide more precise insights into performance in target deployment environments, as current evaluations are conducted primarily on desktop and server machines.
This extension is particularly important for edge computing applications such as robotics, where power constraints, thermal limitations, and specialized accelerators significantly influence real-world performance.
By developing standardized benchmarking procedures for specific deployment platforms such as NVIDIA Jetson\footnote{\url{https://www.nvidia.com/en-us/autonomous-machines/embedded-systems/}}, Google Coral\footnote{\url{https://coral.ai/}}, or custom FPGA implementations, \method could offer more accurate predictions of deployment performance. This would help bridge the gap between research prototypes and production systems by identifying specific optimization opportunities for the target hardware.

To further standardize evaluations in \method, future work should address potential variations due to different hardware setups, Python versions, and code implementations.
Even with our current efforts to ensure reproducibility, subtle differences in environment configurations can lead to meaningful performance variations.
Creating a centralized evaluation server, similar to the approach taken by MLPerf \cite{reddi2020mlperf}, could further standardize comparisons by running all submissions in identical environments. These enhancements would facilitate more accurate comparisons between different computing environments.

As an open-source platform, \method is designed to evolve through community contributions. Researchers can extend the benchmark in multiple ways: by adding new domains through the standard Gymnasium interface, by implementing additional baseline algorithms, or by introducing domain-specific metrics that capture other aspects of real-world performance. The repository includes detailed contribution guidelines, templates, and documentation to facilitate these extensions. This collaborative approach ensures \method remains relevant to emerging research challenges while expanding its coverage of real-world autonomous agent applications.

%% file: conclusion.tex
\noindent 
We need more holistic metrics and representative benchmarks to measure progress. To this end, we introduced \method, a benchmarking suite that can be used for evaluating autonomous agents on challenging tasks from domains such as computer chip floorplanning, web navigation, and quadruped locomotion. \method provides a standardized set of metrics across data cost, application performance, system resource efficiency, and reliability, enabling a comprehensive comparison of different algorithms.
Our evaluations demonstrate \method's effectiveness in identifying the strengths and weaknesses of various approaches to developing autonomous agents.
We encourage the community to contribute new domains, tasks, and algorithms to \method, making it an even more comprehensive platform for benchmarking autonomous agents in real-world-inspired settings.

%% file: appendix.tex
\addcontentsline{toc}{section}{Appendix} 
\part{Appendix} 
\parttoc 

\section{Additional Experiments}
\label{appendix:additional_experiments}

We present an extensive set of additional experiments that showcase A2Perf's capabilities in evaluating autonomous agents across various domains and tasks. The results encompass a wide range of metrics, including data cost, reliability, system performance, and application performance, providing a holistic view of the strengths and limitations of different algorithmic approaches.

The circuit training domain experiments (Appendix~\ref{appendix:circuit_training}) reveal interesting trade-offs between behavioral cloning, DDQN, and PPO in terms of data efficiency, computational requirements, and performance consistency. Moving to the quadruped locomotion domain (Appendix~\ref{appendix:quadruped_locomotion}), we observe how the reliability metrics shed light on the robustness and worst-case behavior of the agents during both training and inference phases. The web navigation domain (Appendix~\ref{appendix:quadruped_locomotion}) introduces an additional layer of complexity, with websites of varying difficulty levels. Here, the system performance metrics highlight the substantial computational demands, particularly in terms of memory usage, associated with training web navigation agents. To further facilitate a clear and intuitive comparison of the algorithms' performance across all domains and tasks, we have included graphical visualizations (Appendix~\ref{appendix:radar_plots}) that summarize the key metrics along different evaluation dimensions.

These experiments show A2Perf's versatility in providing a comprehensive and nuanced evaluation of autonomous agents operating in diverse and realistic settings. By considering multiple performance aspects and presenting the results in both tabular and graphical formats, A2Perf enables researchers and practitioners to gain valuable insights into the behavior and limitations of different algorithmic choices, ultimately guiding the development of more robust and efficient autonomous agents.

\input{appendix/additional_experiments}

\section{Experimental Setup}
\label{appendix:experimental_setup}

\input{appendix/experimental_setup}

\section{Hyperparameters}
\label{appendix:hyperparameters}
\input{appendix/hyperparameters}

\section{Dataset Collection}
\label{appendix:dataset_collection}
\input{appendix/dataset_collection}

\section{Dataset Information}
\label{appendix:dataset_information}
\input{appendix/dataset_information}

\section{Software Usage}
\label{appendix:software_usage}
\input{appendix/software_usage}

\section{Website Generation \& Agent Interaction}
\label{appendix:website_generation}
\input{appendix/website_generation}

%% file: appendix/additional_experiments.tex
\subsection{Circuit Training}
\label{appendix:circuit_training}

This section shows the full set of metrics for the toy macro standard cell and Ariane netlists in the circuit training domain. The results highlight the differences in data cost, reliability, system performance, and application performance between behavioral cloning (BC), DDQN, and PPO.

\begin{table}[!htbp]
\centering
\resizebox{0.9\textwidth}{!}{%
\begin{tabular}{|l|l|c|c|c|}
\toprule
\multicolumn{5}{|c|}{\textbf{Toy Macro Standard Cell (Training)}} \\

 &  & \textbf{BC} & \textbf{DDQN} & \textbf{PPO} \\
\textbf{Category} & \textbf{Metric Name} &  &  &  \\
\midrule
\multirow[t]{1}{*}{Data Cost} & Training Sample Cost (kWh) & 4.44 & 0 & 0 \\
\midrule
\multirow[t]{2}{*}{Application} & Generalization (100 eps. [all tasks]) & -2.19 & -2.20 & -2.13 \\
 & Returns (100 eps.) & -0.97 $\pm$ \num{2.27e-03} & -1.05 $\pm$ 0.04 & -0.97 $\pm$ \num{8.09e-03} \\
\cline{1-5}
\multirow[t]{5}{*}{Reliability} & Dispersion Across Runs (IQR) & N/A & 0.01 $\pm$ 0.01 & 9.07e-03 $\pm$ \num{6.43e-03} \\
 & Dispersion Within Runs (IQR) & N/A & \num{8.80e-03} $\pm$ 0.01 & \num{2.51e-03} $\pm$ \num{3.61e-03} \\
 & Long Term Risk (CVaR) & N/A & 1.10 & 0.04 \\
 & Risk Across Runs (CVaR) & N/A & -1.08 & -0.99 \\
 & Short Term Risk (CVaR) & N/A & 0.03 & \num{9.89e-03} \\
\cline{1-5}
\multirow[t]{4}{*}{System} & Energy Consumed (kWh) & 0.02 $\pm$ \num{1.97e-04} & 5.55 $\pm$ 2.03 & 15.37 $\pm$ 3.79 \\
 & GPU Power Usage (W) & 188.20 $\pm$ 21.98 & 448.00 $\pm$ 200.41 & 307.05 $\pm$ 69.75 \\
 & Peak RAM Usage (GB) & 4.71 $\pm$ 0.02 & 525.99 $\pm$ 205.64 & 675.26 $\pm$ 45.30 \\
 & Wall Clock Time (Hours) & 0.10 $\pm$ \num{1.36e-03} & 0.29 $\pm$ 0.57 & 1.79 $\pm$ 2.16 \\
\cline{1-5}
\midrule
\multicolumn{5}{|c|}{\textbf{Toy Macro Standard Cell (Inference)}} \\
\multirow[t]{2}{*}{Reliability} & Dispersion Across Rollouts (IQR) & \num{1.68e-03} & 0.09 & \num{2.43e-03} \\
 & Risk Across Rollouts (CVaR) & -0.97 & -1.10 & -0.99 \\
\cline{1-5}
\multirow[t]{4}{*}{System} & GPU Power Usage (W) & 104.97 $\pm$ 22.85 & 59.45 $\pm$ 1.43 & 58.97 $\pm$ 1.14 \\
 & Inference Time (ms) & 8.93 $\pm$ 0.51 & 20 $\pm$ 2.69 & 20 $\pm$ 2.67 \\
 & Mean RAM Usage (GB) & 1.92 $\pm$ 0.42 & 1.45 $\pm$ 0.48 & 1.99 $\pm$ 0.30 \\
 & Peak RAM Usage (GB) & 2.14 $\pm$ 0.03 & 2.10 $\pm$ 0.05 & 2.16 $\pm$ 0.07 \\
\bottomrule
\end{tabular}
}
\vspace{0.5em}
\caption{Metrics for the "Toy Macro" netlist task of CircuitTraining-v0. All metrics are averaged over ten random seeds. Note that the training reliability metrics for BC are marked as ``N/A'' since BC does not perform online rollouts in the environment.}
\label{tab:appendix_metrics_circuit_training_toy_macro_stdcell}
\end{table}

\begin{table}[H]
\resizebox{0.9\textwidth}{!}{%
\begin{tabular}{|l|l|c|c|c|}

\toprule
\multicolumn{5}{|c|}{\textbf{Ariane (Training)}} \\
 &  & \textbf{BC} & \textbf{DDQN} & \textbf{PPO} \\
\textbf{Category} & \textbf{Metric Name} &  &  &  \\
\midrule
\multirow[t]{1}{*}{Data Cost} & Training Sample Cost & 48.28 & 0 & 0 \\
\midrule
\multirow[t]{2}{*}{Application} & Generalization (100 eps. [all tasks]) & -2.18 & -2.19 & -2.05 \\
 & Returns (100 eps.) & -1.10 $\pm$ 0.04 & -1.13 $\pm$ 0.04 & -0.99 $\pm$ \num{7.25e-03} \\
\cline{1-5}
\multirow[t]{5}{*}{Reliability} & Dispersion Across Runs (IQR) & N/A & 0.03 $\pm$ 0.03 & 0.04 $\pm$ 0.02 \\
 & Dispersion Within Runs (IQR) & N/A & 0.02 $\pm$ 0.03 & \num{4.77e-03} $\pm$ \num{4.92e-03} \\
 & Long Term Risk (CVaR) & N/A & 1.20 & 0.03 \\
 & Risk Across Runs (CVaR) & N/A & -1.17 & -1.03 \\
 & Short Term Risk (CVaR) & N/A & 0.07 & 0.01 \\
\cline{1-5}
\multirow[t]{5}{*}{System} & Energy Consumed (kWh) & 0.11 $\pm$ \num{6.45e-04} & 108.20 $\pm$ 4.29 & 120.53 $\pm$ 2.78 \\

 & GPU Power Usage (W) & 211.35 $\pm$ 16.76 & 585.98 $\pm$ 172.50 & 692.94 $\pm$ 120.08 \\
 & Mean RAM Usage (GB) & 4.72 $\pm$ 0.53 & 849.37 $\pm$ 64.85 & 834.05 $\pm$ 55.90 \\
 & Peak RAM Usage (GB) & 5.25 $\pm$ 0.07 & 889.56 $\pm$ 23.44 & 906.45 $\pm$ 68.01 \\
 & Wall Clock Time (Hours) & 0.48 $\pm$ 2.61e-03 & 21.94 $\pm$ 0.90 & 23.95 $\pm$ 0.54 \\
\cline{1-5}
\midrule
\multicolumn{5}{|c|}{\textbf{Ariane (Inference)}} \\
\multirow[t]{2}{*}{Reliability} & Dispersion Across Rollouts (IQR) & 0.01 & 0.05 & 0.01 \\
 & Risk Across Rollouts (CVaR) & -1.23 & -1.25 & -1.01 \\
\cline{1-5}
\multirow[t]{4}{*}{System} & GPU Power Usage (W) & 136.91 $\pm$ 21.48 & 69.50 $\pm$ 4.60 & 49.43 $\pm$ 30.29 \\
 & Inference Time (ms) & 10.0 $\pm$ 0.46 & 20.0 $\pm$ 2.69 & 20.0 $\pm$ 2.68 \\
 & Mean RAM Usage (GB) & 2.19 $\pm$ 0.21 & 2.15 $\pm$ 0.30 & 2.51 $\pm$ 0.49 \\
 & Peak RAM Usage (GB) & 2.29 $\pm$ 0.01 & 2.28 $\pm$ 0.13 & 2.71 $\pm$ 0.62 \\
\cline{1-5}
\bottomrule
\end{tabular}
}
\vspace{0.5em}
\caption{Metrics for the Ariane Netlist task of CircuitTraining-v0. All metrics are averaged over ten random seeds. Note that the training reliability metrics for BC are marked as ``N/A'' since BC does not perform online rollouts in the environment.}
\label{tab:appendix_metrics_circuit_training_ariane}
\end{table}

\subsection{Quadruped Locomotion}
\label{appendix:quadruped_locomotion}

This section reports the metrics for the dog pace, trot, and spin gaits in the quadruped locomotion domain. The reliability metrics provide insights into the stability and worst-case performance of the algorithms during training and inference.

\begin{table}[H]
\resizebox{0.9\textwidth}{!}{%
\begin{tabular}{|l|l|c|c|c|}
\toprule
\multicolumn{5}{|c|}{\textbf{Dog Pace (Training)}} \\
 &  & \textbf{BC} & \textbf{PPO} & \textbf{SAC} \\
\textbf{Category} & \textbf{Metric Name} &  &  &  \\
\midrule
\multirow[t]{1}{*}{Data Cost} & Training Sample Cost (kWh) & 22.53 & 0 & 0 \\
\midrule
\multirow[t]{2}{*}{Application} & Generalization (100 eps. [all tasks]) & 3.99 & 3.36 & 5.03 \\
 & Returns (100 eps.) & 7.00 $\pm$ 4.68 & 9.94 $\pm$ 15.59 & 6.96 $\pm$ 6.72 \\
\cline{1-5}
\multirow[t]{5}{*}{Reliability} & Dispersion Across Runs (IQR) & N/A & 9.63 $\pm$ 7.27 & 3.61 $\pm$ 3.88 \\
 & Dispersion Within Runs (IQR) & N/A & 2.22 $\pm$ 1.97 & 2.98 $\pm$ 3.64 \\
 & Long Term Risk (CVaR) & N/A & 13.00 & 25.82 \\
 & Risk Across Runs (CVaR) & N/A & 13.74 & 8.55 \\
 & Short Term Risk (CVaR) & N/A & 5.81 & 10.19 \\
\cline{1-5}
\multirow[t]{5}{*}{System} & Energy Consumed (kWh) & 0.11 $\pm$ 0.02 & 32.46 $\pm$ 0.26 & 36.22 $\pm$ 2.33 \\
 & GPU Power Usage (W) & 240.64 $\pm$ 5.41 & 280.12 $\pm$ 23.69 & 266.37 $\pm$ 9.54 \\
 & Mean RAM Usage (GB) & 3.21 $\pm$ 0.24 & 532.93 $\pm$ 14.28 & 516.24 $\pm$ 75.03 \\
 & Peak RAM Usage (GB) & 3.25 $\pm$ 0.01 & 534.26 $\pm$ 2.04 & 545.16 $\pm$ 0.50 \\
 & Wall Clock Time (Hours) & 0.46 $\pm$ 0.07 & 18.73 $\pm$ 0.19 & 19.41 $\pm$ 2.74 \\
\cline{1-5}
\midrule
\multicolumn{5}{|c|}{\textbf{Dog Pace (Inference)}} \\
\multirow[t]{2}{*}{Reliability} & Dispersion Across Rollouts (IQR) & 0.52 & 8.76 & 4.80 \\
 & Risk Across Rollouts (CVaR) & 0.33 & 0.46 & 1.69 \\
\cline{1-5}
\multirow[t]{4}{*}{System} & GPU Power Usage (W) & 60.37 $\pm$ 1.78 & 59.11 $\pm$ 1.31 & 61.41 $\pm$ 1.96 \\
 & Inference Time (ms) & 2.33 $\pm$ 0.54 & 2.56 $\pm$ 0.39 & 2.52 $\pm$ 0.74 \\

 & Mean RAM Usage (GB) & 1.69 $\pm$ 0.31 & 1.81 $\pm$ 0.14 & 1.71 $\pm$ 0.30 \\
 & Peak RAM Usage (GB) & 1.82 $\pm$ 0.03 & 1.84 $\pm$ 9.05e-03 & 1.85 $\pm$ 0.04 \\
\bottomrule
\end{tabular}
}
\vspace{0.5em}
\caption{Metrics for the "dog pace" gait of QuadrupedLocomotion-v0. All metrics are averaged over ten random seeds. Note that the training reliability metrics for BC are marked as ``N/A'' since BC does not perform online rollouts in the environment.}
\label{tab:appendix_locomotion_dog_pace}
\end{table}

\begin{table}[!htbp]
\resizebox{0.95\linewidth}{!}{%
\begin{tabular}{|l|l|c|c|c|}
\toprule
\multicolumn{5}{|c|}{\textbf{Dog Trot (Training)}} \\
 &  & \textbf{BC} & \textbf{PPO} & \textbf{SAC} \\
\textbf{Category} & \textbf{Metric Name} &  &  &  \\
\midrule
\multirow[t]{1}{*}{Data Cost} & Training Sample Cost (kWh) & 15.77 & 0 & 0 \\
\midrule
\multirow[t]{2}{*}{Application} & Generalization (100 eps. [all tasks]) & 3.87 & 3.09 & 4.49 \\
 & Returns (100 eps.) & 1.06 $\pm$ 0.26 & 1.49 $\pm$ 1.02 & 3.51 $\pm$ 2.88 \\
\cline{1-5}
\multirow[t]{5}{*}{Reliability} & Dispersion Across Runs (IQR) & N/A & 9.07 $\pm$ 4.93 & 0.85 $\pm$ 1.29 \\
 & Dispersion Within Runs (IQR) & N/A & 0.82 $\pm$ 0.84 & 0.93 $\pm$ 1.11 \\
 & Long Term Risk (CVaR) & N/A & 6.79 & 8.46 \\
 & Risk Across Runs (CVaR) & N/A & 6.00 & 2.58 \\
 & Short Term Risk (CVaR) & N/A & 2.41 & 3.20 \\
\cline{1-5}
\multirow[t]{5}{*}{System} & Energy Consumed (kWh) & 0.12 $\pm$ 0.02 & 16.82 $\pm$ 0.29 & 19.17 $\pm$ 0.64 \\
 & GPU Power Usage (W) & 242.12 $\pm$ 7.53 & 277.71 $\pm$ 23.47 & 269.18 $\pm$ 10.12 \\
 & Mean RAM Usage (GB) & 3.21 $\pm$ 0.25 & 535.00 $\pm$ 18.77 & 535.99 $\pm$ 29.49 \\
 & Peak RAM Usage (GB) & 3.26 $\pm$ 0.01 & 536.47 $\pm$ 1.98 & 544.80 $\pm$ 4.39 \\
 & Wall Clock Time (Hours) & 0.46 $\pm$ 0.06 & 18.57 $\pm$ 0.23 & 18.99 $\pm$ 6.78 \\
\cline{1-5}
\midrule
\multicolumn{5}{|c|}{\textbf{Dog Trot (Inference)}} \\
\multirow[t]{2}{*}{Reliability} & Dispersion Across Rollouts (IQR) & 0.32 & 0.89 & 1.25 \\
 & Risk Across Rollouts (CVaR) & 0.63 & 0.36 & 1.33 \\
\cline{1-5}
\multirow[t]{4}{*}{System} & GPU Power Usage (W) & 59.32 $\pm$ 1.08 & 58.91 $\pm$ 1.28 & 59.39 $\pm$ 1.23 \\

 & Inference Time (ms) & 2.32 $\pm$ 0.49 & 2.55 $\pm$ 0.57 & 2.45 $\pm$ 0.35 \\

 & Mean RAM Usage (GB) & 1.66 $\pm$ 0.33 & 1.76 $\pm$ 0.25 & 1.80 $\pm$ 0.17 \\
 & Peak RAM Usage (GB) & 1.82 $\pm$ \num{8.77e-04} & 1.85 $\pm$ 0.02 & 1.85 $\pm$ 0.03 \\
\bottomrule
\end{tabular}
}
\vspace{0.5em}
\caption{Metrics for the "dog trot" gait of QuadrupedLocomotion-v0. All metrics are averaged over ten random seeds. Note that the training reliability metrics for BC are marked as ``N/A'' since BC does not perform online rollouts in the environment.}
\label{tab:appendix_locomotion_dog_trot}
\end{table}

\begin{table}[!htbp]
\resizebox{0.95\linewidth}{!}{%
\begin{tabular}{|l|l|c|c|c|}
\toprule
\multicolumn{5}{|c|}{\textbf{Dog Spin (Training)}} \\
 &  & \textbf{BC} & \textbf{PPO} & \textbf{SAC} \\
\textbf{Category} & \textbf{Metric Name} &  &  &  \\
\midrule
\multirow[t]{1}{*}{Data Cost} & Training Sample Cost (kWh) &30.17 & 0 & 0 \\
\midrule
\multirow[t]{2}{*}{Application} & Generalization (100 eps. [all tasks]) & 3.97 & 2.69 & 4.61 \\
 & Returns (100 eps.) & 1.54 $\pm$ 0.42 & 3.82 $\pm$ 6.22 & 3.84 $\pm$ 1.46 \\
\cline{1-5}
\multirow[t]{5}{*}{Reliability} & Dispersion Across Runs (IQR) & N/A & 7.92 $\pm$ 4.60 & 0.74 $\pm$ 0.76 \\
 & Dispersion Within Runs (IQR) & N/A & 1.00 $\pm$ 1.08 & 0.84 $\pm$ 1.26 \\
 & Long Term Risk (CVaR) & N/A & 8.88 & 14.37 \\
 & Risk Across Runs (CVaR) & N/A & 8.29 & 3.82 \\
 & Short Term Risk (CVaR) & N/A & 3.09 & 2.99 \\
\cline{1-5}
\multirow[t]{5}{*}{System} & Energy Consumed (kWh) & 0.10 $\pm$ 0.04 & 17.42 $\pm$ 0.35 & 18.88 $\pm$ 0.59 \\
 & GPU Power Usage (W) & 216.72 $\pm$ 68.63 & 278.38 $\pm$ 22.60 & 264.46 $\pm$ 9.49 \\
 & Mean RAM Usage (GB) & 3.18 $\pm$ 0.26 & 534.56 $\pm$ 21.28 & 531.27 $\pm$ 55.64 \\
 & Peak RAM Usage (GB) & 3.23 $\pm$ 0.08 & 536.10 $\pm$ 3.03 & 477.22 $\pm$ 172.63 \\
 & Wall Clock Time (Hours) & 0.45 $\pm$ 0.08 & 17.13 $\pm$ 6.07 & 17.02 $\pm$ 9.05 \\
\cline{1-5}
\midrule
\multicolumn{5}{|c|}{\textbf{Dog Spin (Inference)}} \\
\multirow[t]{2}{*}{Reliability} & Dispersion Across Rollouts (IQR) & 0.37 & 2.41 & 1.78 \\
 & Risk Across Rollouts (CVaR) & 0.28 & 0.12 & 0.55 \\
\cline{1-5}
\multirow[t]{4}{*}{System} & GPU Power Usage (W) & 60.10 $\pm$ 1.14 & 59.70 $\pm$ 1.22 & 59.65 $\pm$ 1.73 \\
 & Inference Time (ms) & 2.33 $\pm$ 0.66 & 2.45 $\pm$ 0.48 & 2.41 $\pm$ 0.22 \\

 & Mean RAM Usage (GB) & 1.68 $\pm$ 0.32 & 1.79 $\pm$ 0.22 & 1.75 $\pm$ 0.26 \\
 & Peak RAM Usage (GB) & 1.82 $\pm$ 0.03 & 1.85 $\pm$ 0.02 & 1.84 $\pm$ 0.02 \\
\bottomrule
\end{tabular}
}
\vspace{0.5em}
\caption{Metrics for the "dog spin" gait of QuadrupedLocomotion-v0. All metrics are averaged over ten random seeds. Note that the training reliability metrics for BC are marked as ``N/A'' since BC does not perform online rollouts in the environment.}
\label{tab:appendix_locomotion_dog_spin}

\end{table}

\subsection{Web Navigation}
\label{appendix:web_navigation}

This section details the evaluation on websites of varying difficulty levels in the web navigation domain. The system performance metrics underscore the significant computational requirements, especially in terms of RAM usage, for training web navigation agents.

\begin{table}[!htbp]
\resizebox{0.95\linewidth}{!}{%
\begin{tabular}{|l|l|c|c|c|}
\toprule
\multicolumn{5}{|c|}{\textbf{Difficulty 1, 1 Website (Training)}} \\
 &  & \textbf{BC} & \textbf{DDQN} & \textbf{PPO} \\
\textbf{Category} & \textbf{Metric Name} &  &  &  \\
\midrule
\multirow[t]{1}{*}{Data Cost} & Training Sample Cost (kWh) & 14.15 & 0 & 0 \\
\midrule
\multirow[t]{2}{*}{Application} & Generalization (100 eps. [all tasks]) & -12.94 & -11.15 & -24.54 \\
 & Returns (100 eps.) & -3.57 $\pm$ 2.80 & -7.55 $\pm$ 5.74 & -13.45 $\pm$ 0.51 \\
\cline{1-5}
\multirow[t]{5}{*}{Reliability} & Dispersion Across Runs (IQR) & N/A & 0.73 $\pm$ 0.63 & 4.20 $\pm$ 1.45 \\
 & Dispersion Within Runs (IQR) & N/A & 0.37 $\pm$ 0.68 & 0.57 $\pm$ 0.53 \\
 & Long Term Risk (CVaR) & N/A & 9.32 & 12.12 \\
 & Risk Across Runs (CVaR) & N/A & -2.75 & -13.11 \\
 & Short Term Risk (CVaR) & N/A & 1.79 & 1.86 \\
\cline{1-5}
\multirow[t]{5}{*}{System} & Energy Consumed (kWh) & 0.04 $\pm$ \num{6.02e-04} & 29.56 $\pm$ 7.23 & 28.82 $\pm$ 1.19 \\
 & GPU Power Usage (W) & 125.89 $\pm$ 2.53 & 265.09 $\pm$ 21.50 & 305.15 $\pm$ 34.41 \\
 & Mean RAM Usage (GB) & 4.10 $\pm$ 0.33 & 1140.98 $\pm$ 580.55 & 1592.45 $\pm$ 388.64 \\
 & Peak RAM Usage (GB) & 4.23 $\pm$ 0.04 & 1931.54 $\pm$ 242.31 & 2305.57 $\pm$ 135.48 \\
 & Wall Clock Time (Hours) & 0.31 $\pm$ \num{4.91e-03} & 8.13 $\pm$ 5.17 & 10.50 $\pm$ 0.44 \\
\cline{1-5}
\midrule
\multicolumn{5}{|c|}{\textbf{Difficulty 1, 1 Website (Inference)}} \\
\multirow[t]{2}{*}{Reliability} & Dispersion Across Rollouts (IQR) & 3.36 & 11.75 & 0.50 \\
 & Risk Across Rollouts (CVaR) & -10.65 & -13.25 & -13.75 \\
\cline{1-5}
\multirow[t]{4}{*}{System} & GPU Power Usage (W) & 108.61 $\pm$ 15.76 & 59.61 $\pm$ 1.41 & 60.26 $\pm$ 1.14 \\
 & Inference Time (ms) & 3.07 $\pm$ 0.47 & 110 $\pm$ 9.93 & 120 $\pm$ 9.71 \\

 & Mean RAM Usage (GB) & 1.97 $\pm$ 0.32 & 2.08 $\pm$ 0.20 & 2.12 $\pm$ 0.17 \\
 & Peak RAM Usage (GB) & 2.11 $\pm$ 0.11 & 2.18 $\pm$ 0.11 & 2.19 $\pm$ 0.09 \\
\cline{1-5}
\bottomrule
\end{tabular}
}
\vspace{0.5em}
\caption{Metrics for "difficulty 1, 1 website" task of WebNavigation-v0. All metrics are averaged over ten random seeds. Note that the training reliability metrics for BC are marked as ``N/A'' since BC does not perform online rollouts in the environment.}
\label{tab:appendix_metrics_webnav_difficulty_1_websites_1}
\end{table}

\begin{table}[!htbp]
\resizebox{0.95\linewidth}{!}{%
\begin{tabular}{|l|l|c|c|c|}
\toprule
\multicolumn{5}{|c|}{\textbf{Difficulty 1, 5 Websites (Training)}} \\
 &  & \textbf{BC} & \textbf{DDQN} & \textbf{PPO} \\
\textbf{Category} & \textbf{Metric Name} &  &  &  \\
\midrule
\multirow[t]{1}{*}{Data Cost} & Training Sample Cost (kWh) & 13.66 & 0 & 0 \\
\midrule
\multirow[t]{2}{*}{Application} & Generalization (100 eps. [all tasks]) & -13.34 & -11.03 & -23.86 \\
 & Returns (100 eps.) & -4.87 $\pm$ 3.33 & -3.43 $\pm$ 4.58 & -12.37 $\pm$ 3.53 \\
\cline{1-5}
\multirow[t]{5}{*}{Reliability} & Dispersion Across Runs (IQR) & N/A & 0.43 $\pm$ 0.55 & 3.42 $\pm$ 1.08 \\
 & Dispersion Within Runs (IQR) & N/A & 0.49 $\pm$ 0.97 & 0.75 $\pm$ 0.55 \\
 & Long Term Risk (CVaR) & N/A & 11.27 & 11.70 \\
 & Risk Across Runs (CVaR) & N/A & -1.26 & -12.60 \\
 & Short Term Risk (CVaR) & N/A & 2.47 & 2.05 \\
\cline{1-5}
\multirow[t]{5}{*}{System} & Energy Consumed (kWh) & 0.04 $\pm$ \num{4.82e-04} & 31.59 $\pm$ 5.19 & 28.48 $\pm$ 1.22 \\
 & GPU Power Usage (W) & 126.04 $\pm$ 4.03 & 265.81 $\pm$ 22.08 & 303.28 $\pm$ 34.99 \\
 & Mean RAM Usage (GB) & 4.03 $\pm$ 0.34 & 1206.86 $\pm$ 466.37 & 1545.56 $\pm$ 427.22 \\
 & Peak RAM Usage (GB) & 4.15 $\pm$ 0.11 & 1928.69 $\pm$ 209.62 & 2227.07 $\pm$ 210.77 \\
 & Wall Clock Time (Hours) & 0.30 $\pm$ \num{3.71e-03} & 9.35 $\pm$ 4.70 & 10.45 $\pm$ 0.31 \\

\cline{1-5}
\midrule
\multicolumn{5}{|c|}{\textbf{Difficulty 1, 5 Websites (Inference)}} \\

\multirow[t]{2}{*}{Reliability} & Dispersion Across Rollouts (IQR) & 5.96 & 0.29 & 0.50 \\
 & Risk Across Rollouts (CVaR) & -11.36 & -13.46 & -13.75 \\
\cline{1-5}
\multirow[t]{4}{*}{System} & GPU Power Usage (W) & 108.13 $\pm$ 16.85 & 60.87 $\pm$ 5.78 & 60.17 $\pm$ 1.67 \\
 & Inference Time (ms) & 3.04 $\pm$ 0.44 & 110 $\pm$ 9.83 & 120 $\pm$ 9.21 \\

 & Mean RAM Usage (GB) & 1.97 $\pm$ 0.33 & 2.07 $\pm$ 0.32 & 2.12 $\pm$ 0.16 \\
 & Peak RAM Usage (GB) & 2.12 $\pm$ 0.03 & 2.57 $\pm$ 0.86 & 2.19 $\pm$ 0.01 \\
\cline{1-5}
\bottomrule
\end{tabular}
}
\vspace{0.5em}
\caption{Metrics for "difficulty 1, 5 websites" task of WebNavigation-v0. All metrics are averaged over ten random seeds. Note that the training reliability metrics for BC are marked as ``N/A'' since BC does not perform online rollouts in the environment.}
\label{tab:appendix_metrics_webnav_difficulty_1_websites_5}
\end{table}

\begin{table}[H]
\centering
\resizebox{\textwidth}{!}{%
\begin{tabular}{|l|l|c|c|c|}
\toprule
\multicolumn{5}{|c|}{\textbf{Difficulty 1, 10 Websites (Training)}} \\
 &  & \textbf{BC} & \textbf{DDQN} & \textbf{PPO} \\
\textbf{Category} & \textbf{Metric Name} &  &  &  \\
\midrule
\multirow[t]{1}{*}{Data Cost} & Training Sample Cost (kWh) & 19.71 & 0 & 0 \\
\midrule
Application & Returns (100 eps.) & -4.68 $\pm$ 3.28 & -3.14 $\pm$ 4.24 & -12.73 $\pm$ 2.86 \\
\cline{1-5}
\multirow[t]{5}{*}{Reliability} & Dispersion Across Runs (IQR) & N/A & 0.32 $\pm$ 0.47 & 3.67 $\pm$ 0.63 \\
 & Dispersion Within Runs (IQR) & N/A & 0.42 $\pm$ 0.86 & 0.79 $\pm$ 0.53 \\
 & Long Term Risk (CVaR) & N/A & 9.47 & 11.85 \\
 & Risk Across Runs (CVaR) & N/A & -1.44 & -12.79 \\
 & Short Term Risk (CVaR) & N/A & 2.27 & 1.82 \\
\cline{1-5}
\multirow[t]{5}{*}{System} & Energy Consumed (kWh) & 0.05 $\pm$ \num{2.41e-04} & 27.19 $\pm$ 11.22 & 20.35 $\pm$ 5.77 \\
 & GPU Power Usage (W) & 125.88 $\pm$ 2.33 & 264.98 $\pm$ 24.45 & 304.66 $\pm$ 33.77 \\
 & Mean RAM Usage (GB) & 3.56 $\pm$ 0.39 & 1214.88 $\pm$ 524.77 & 1034.85 $\pm$ 424.83 \\
 & Peak RAM Usage (GB) & 4.10 $\pm$ 0.05 & 1784.37 $\pm$ 641.82 & 1665.15 $\pm$ 395.14 \\
 & Wall Clock Time (Hours) & 0.32 $\pm$ 1.43e-03 & 7.80 $\pm$ 5.20 & 3.10 $\pm$ 5.06 \\
\cline{1-5}
\midrule
\multicolumn{5}{|c|}{\textbf{Difficulty 1, 10 Websites (Inference)}} \\
\multirow[t]{2}{*}{Reliability} & Dispersion Across Rollouts (IQR) & 5.86 & 0.25 & 0.50 \\
 & Risk Across Rollouts (CVaR) & -11.33 & -13.28 & -13.75 \\
\cline{1-5}
\multirow[t]{4}{*}{System} & GPU Power Usage (W) & 108.26 $\pm$ 16.34 & 59.95 $\pm$ 1.49 & 59.67 $\pm$ 1.57 \\
 & Inference Time (ms) & 3.05 $\pm$ 0.45 & 110 $\pm$ 8.42 & 120 $\pm$ 9.90 \\

 & Mean RAM Usage (GB) & 1.97 $\pm$ 0.35 & 2.06 $\pm$ 0.27 & 2.13 $\pm$ 0.16 \\
 & Peak RAM Usage (GB) & 2.13 $\pm$ 0.04 & 2.17 $\pm$ 0.03 & 2.20 $\pm$ 0.03 \\
\cline{1-5}
\bottomrule
\end{tabular}
} 
\vspace{0.5em}
\caption{Metrics for "difficulty 1, 10 websites" task of WebNavigation-v0. All metrics are averaged over ten random seeds. Note that the training reliability metrics for BC are marked as ``N/A'' since BC does not perform online rollouts in the environment.}
\label{tab:appendix_metrics_webnav_difficulty_1_websites_10}
\end{table}

\pagebreak
\subsection{Radar Plots for Easy Visual Comparison}
\label{appendix:radar_plots}

These figures provide a graphical representation of the key metrics across all domains and tasks, enabling a visual comparison of the algorithms' performance along the different evaluation axes.

\begin{figure}[!htbp]
  \centering
  \begin{minipage}[b]{0.53\linewidth}
    \includegraphics[width=\linewidth]{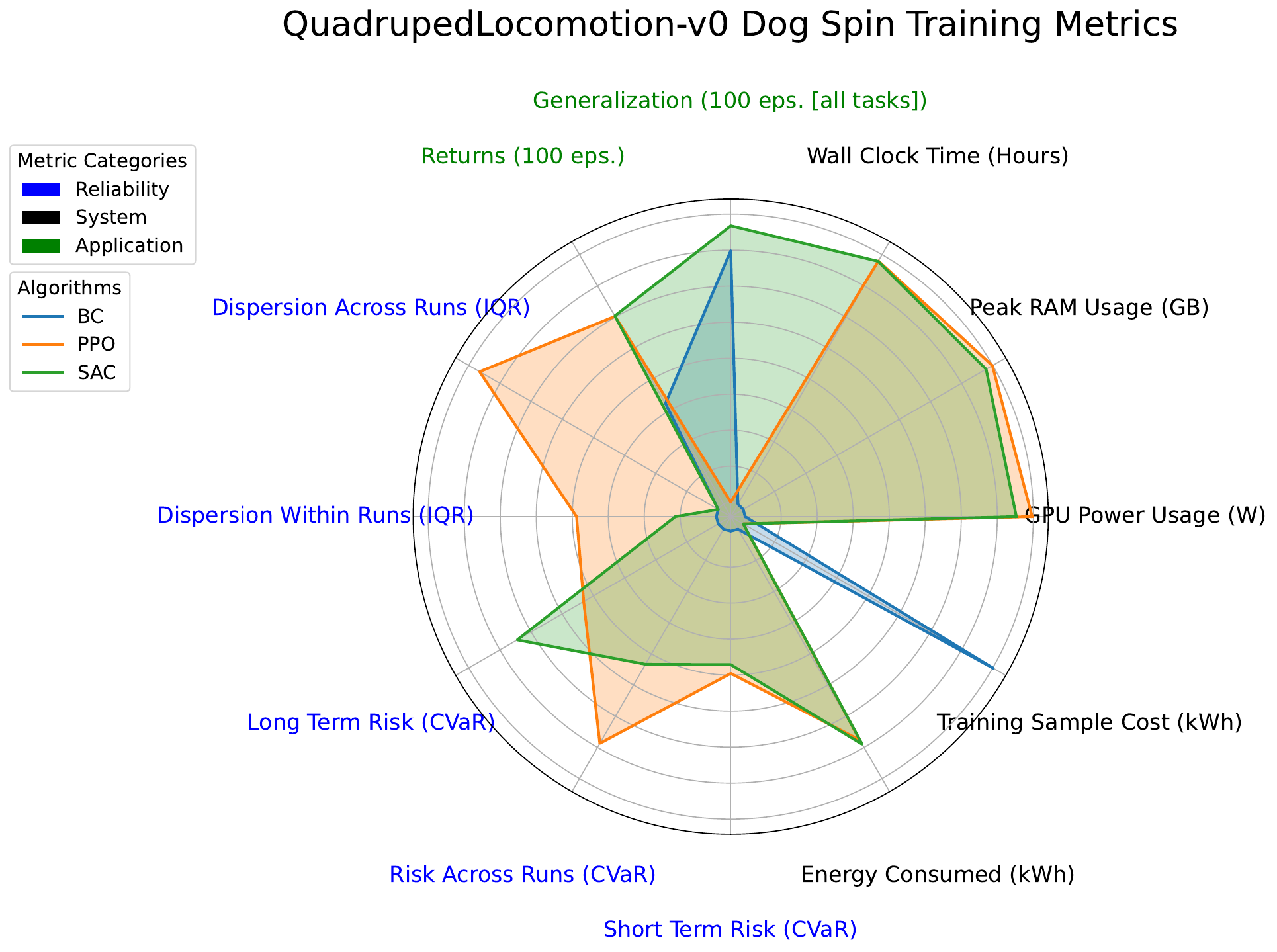}
  \end{minipage}
  \hfill 
  \begin{minipage}[b]{0.4625\linewidth}
    \includegraphics[width=\linewidth]{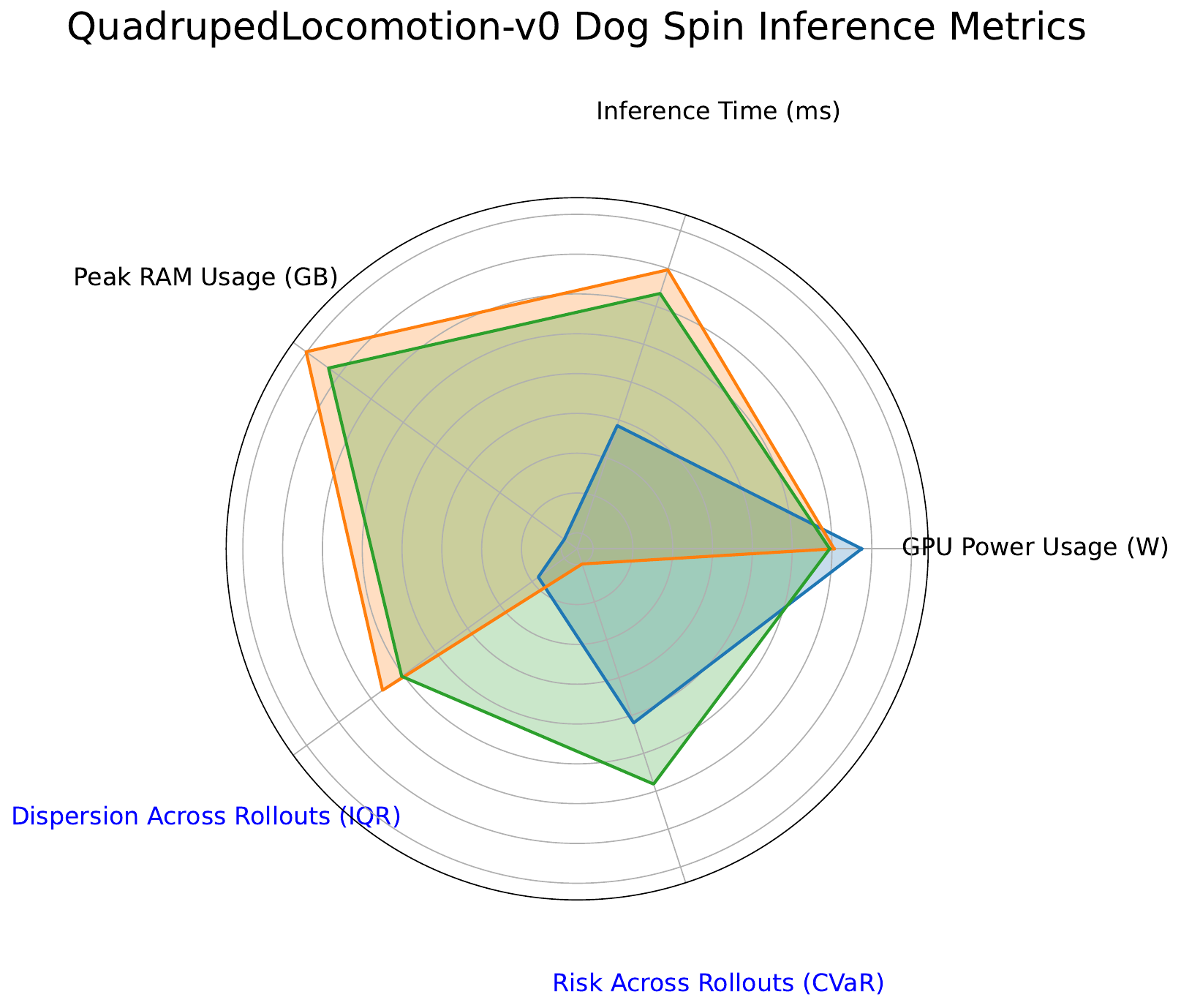}
  \end{minipage}
  \caption{Graphical representation of metrics for the "dog spin" gait of QuadrupedLocomotion-v0}
  \label{fig:appendix_radar_quadruped_locomotion_dog_spin}
\end{figure}

\begin{figure}[!htbp]
  \centering
  \begin{minipage}[b]{0.53\linewidth}
    \includegraphics[width=\linewidth]{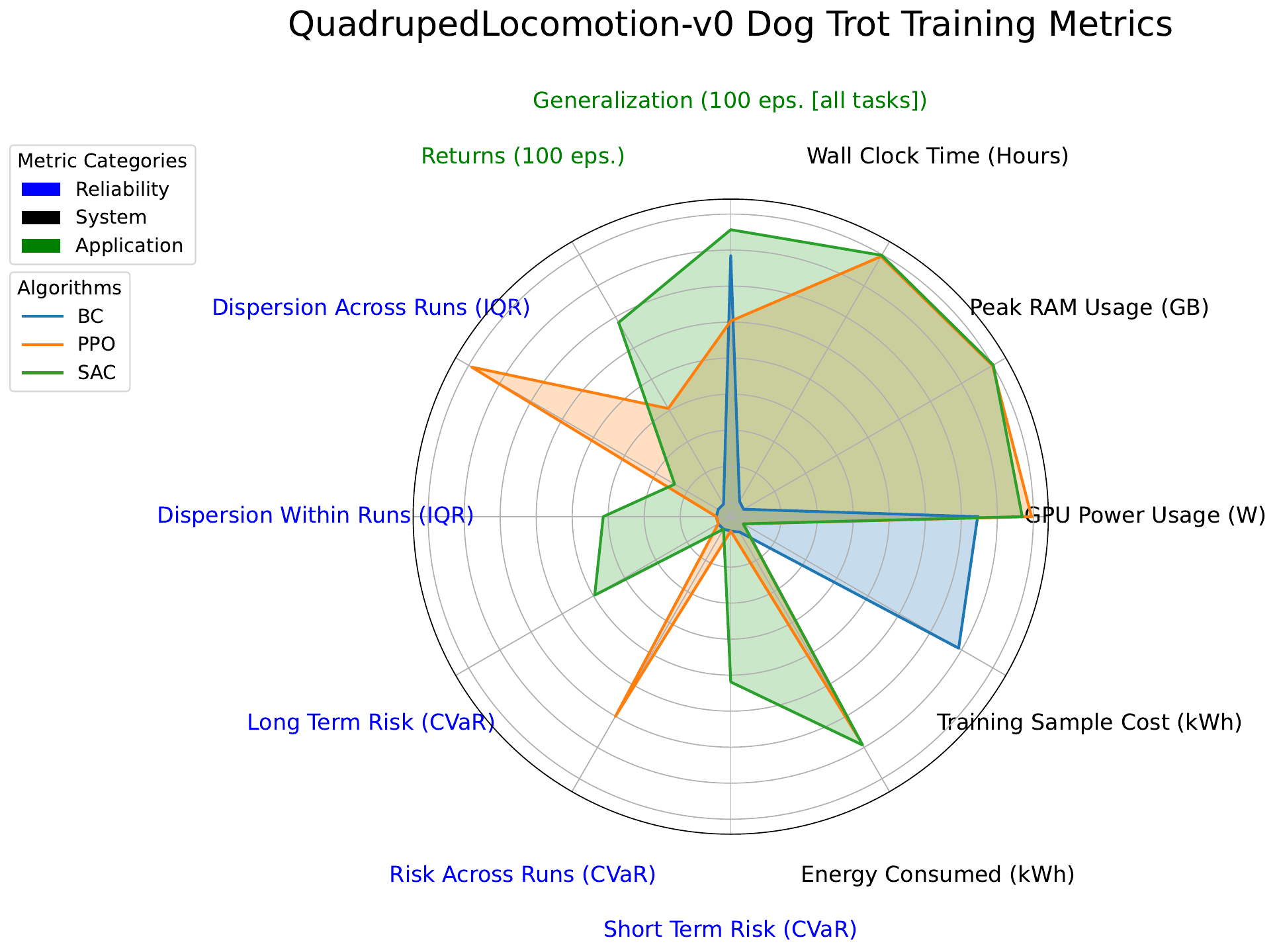}
  \end{minipage}
  \hfill 
  \begin{minipage}[b]{0.4625\linewidth}
    \includegraphics[width=\linewidth]{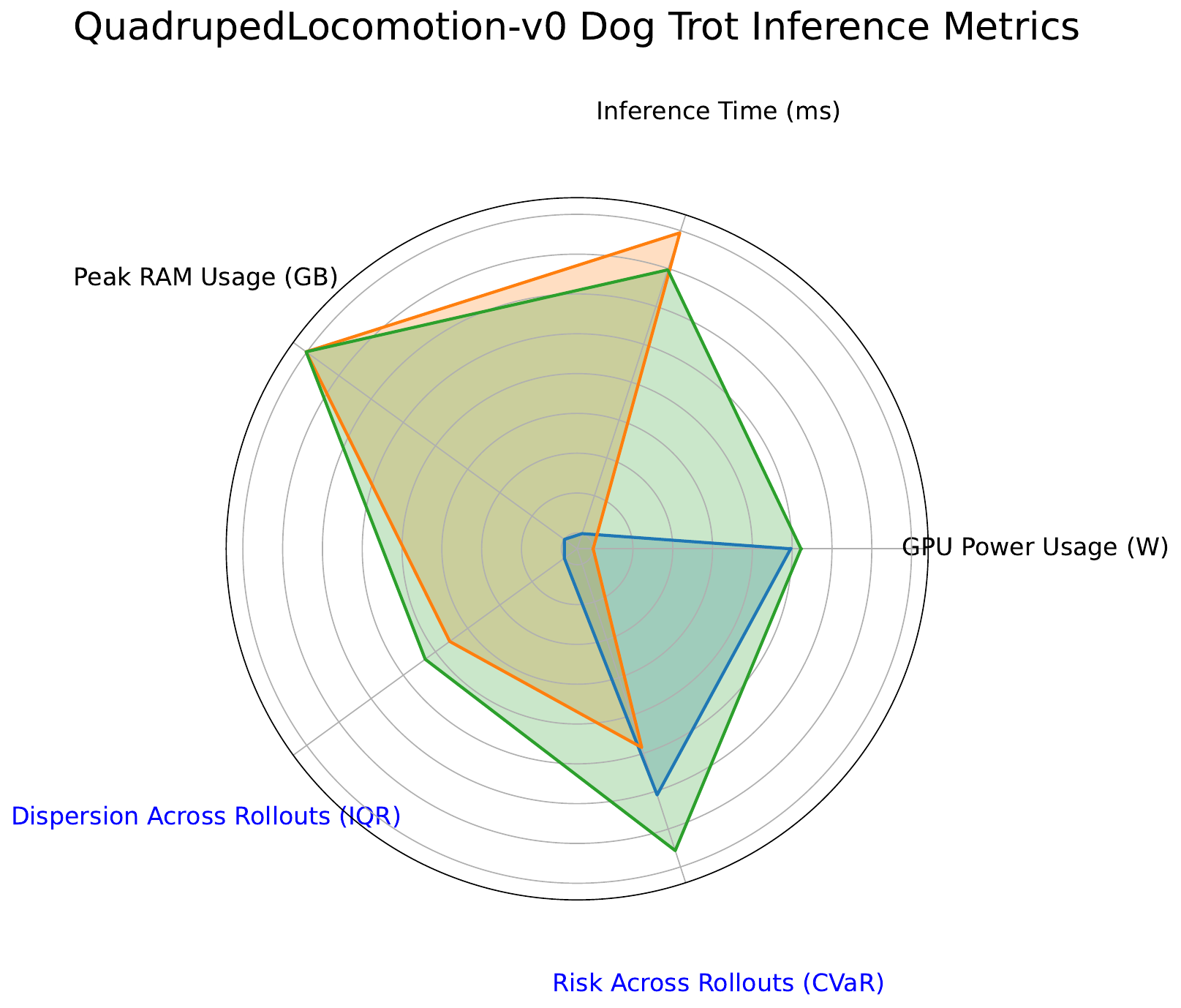}
  \end{minipage}
  \caption{Graphical representation of metrics for the "dog trot" gait of QuadrupedLocomotion-v0}
  \label{fig:appendix_radar_quadruped_locomotion_dog_trot}
\end{figure}
\begin{figure}[H]
  \centering
  \begin{minipage}[b]{0.53\linewidth}
    \includegraphics[width=\linewidth]{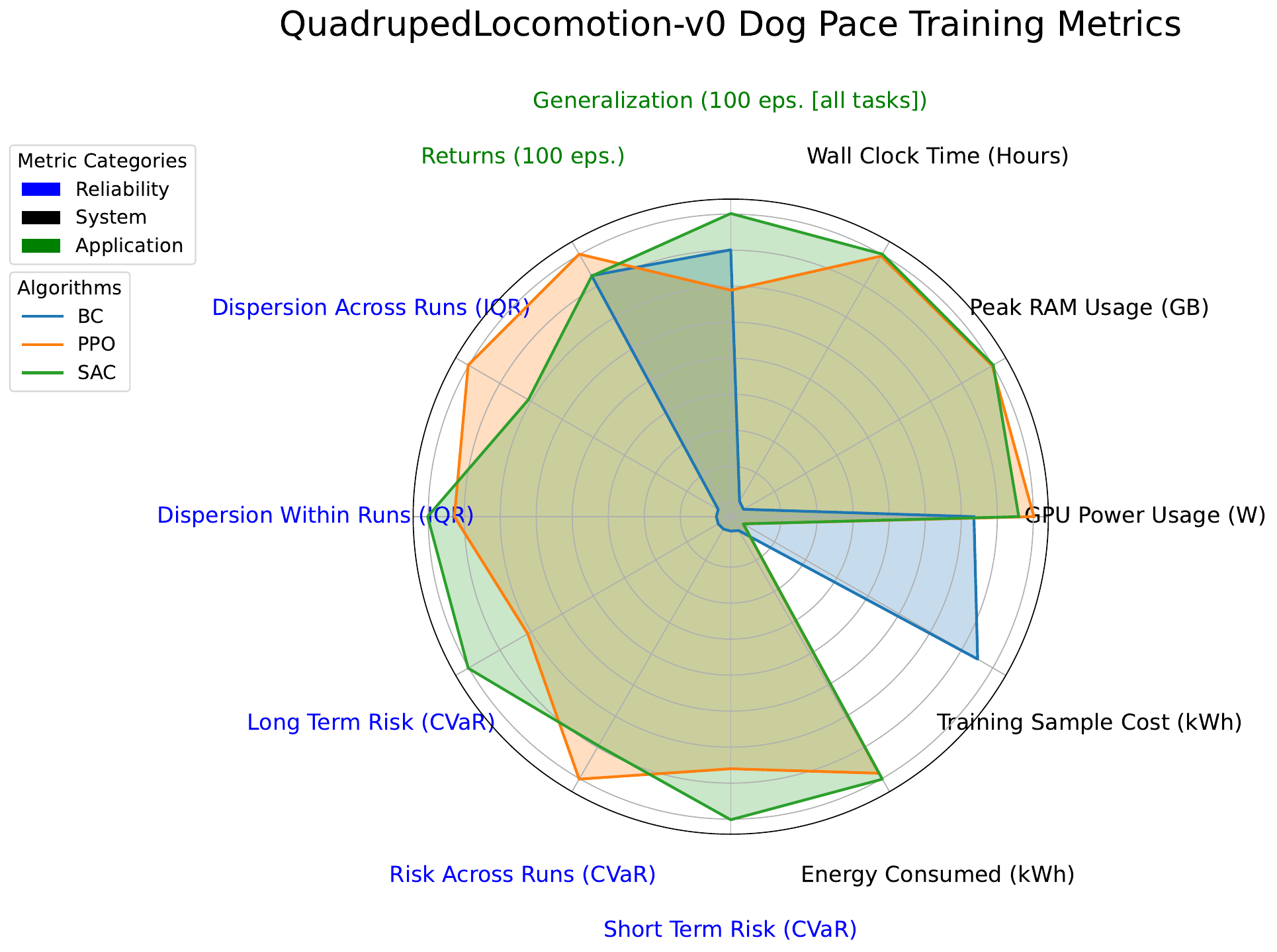}
  \end{minipage}
  \hfill 
  \begin{minipage}[b]{0.4625\linewidth}
    \includegraphics[width=\linewidth]{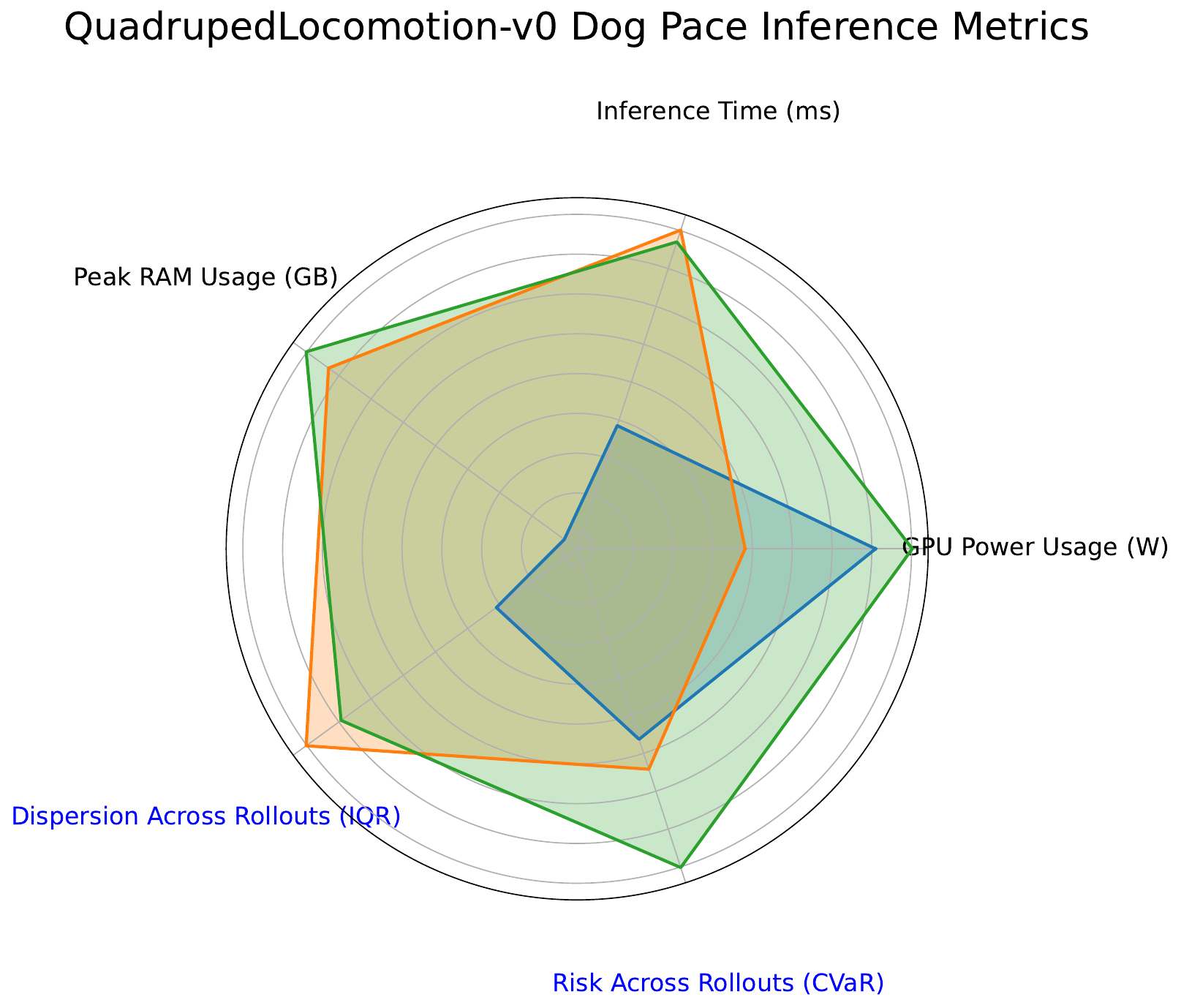}
  \end{minipage}
  \caption{Graphical representation of metrics for the "dog pace" gait of QuadrupedLocomotion-v0}
  \label{fig:appendix_radar_quadruped_locomotion_dog_pace}
\end{figure}

\newpage

\begin{figure}[!htbp]
  \centering
  \begin{minipage}[b]{0.53\linewidth}
    \includegraphics[width=\linewidth]{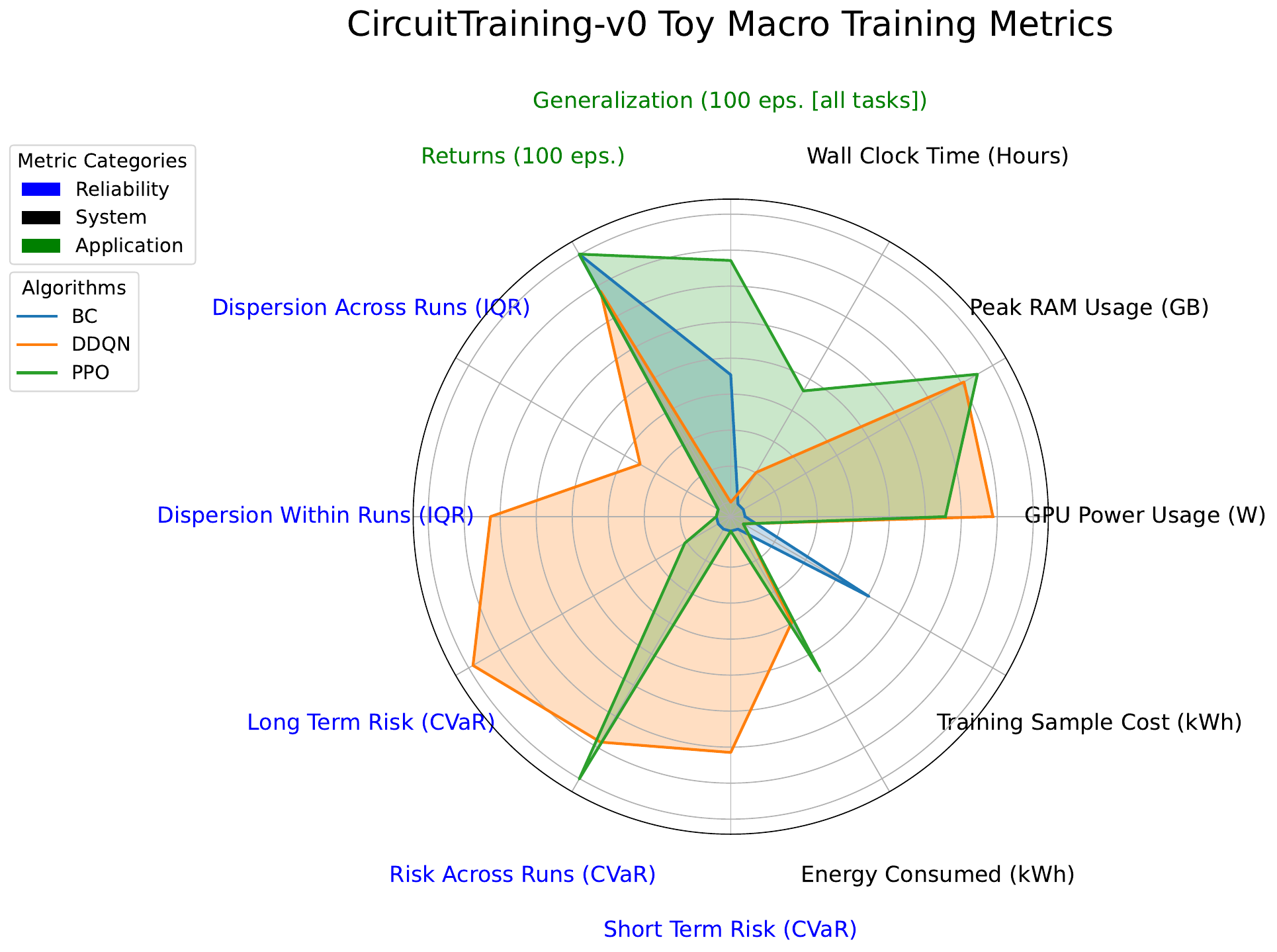}
  \end{minipage}
  \hfill 
  \begin{minipage}[b]{0.44\linewidth}
    \includegraphics[width=\linewidth]{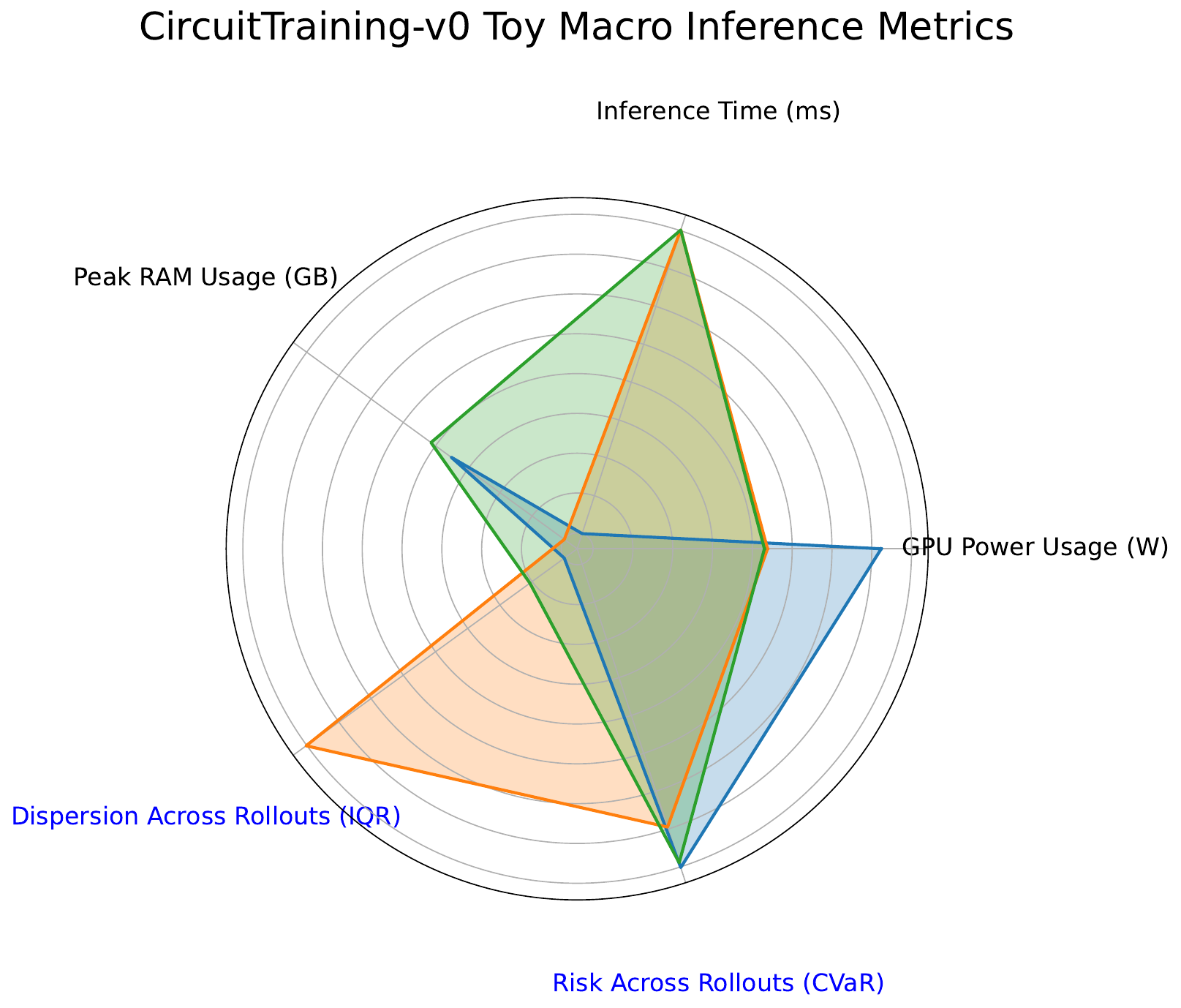}
  \end{minipage}
  \caption{Graphical representation of metrics for the "Toy Macro" netlist task of CircuitTraining-v0}
  \label{fig:appendix_circuit_training_toy_macro}
\end{figure}

\begin{figure}[!htbp]
  \centering
  \begin{minipage}[b]{0.53\linewidth}
    \includegraphics[width=\linewidth]{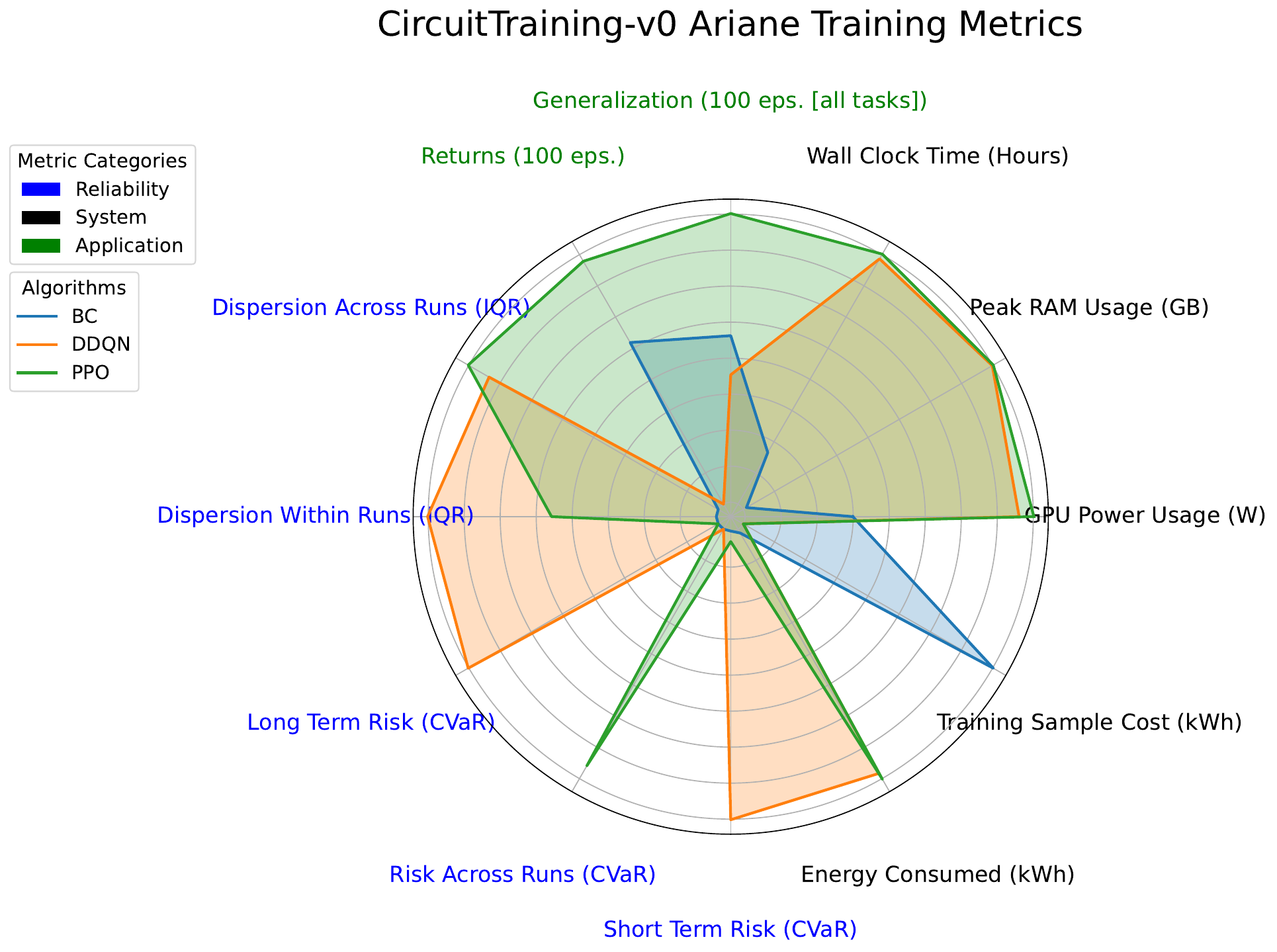}
  \end{minipage}
  \hfill 
  \begin{minipage}[b]{0.44\linewidth}
    \includegraphics[width=\linewidth]{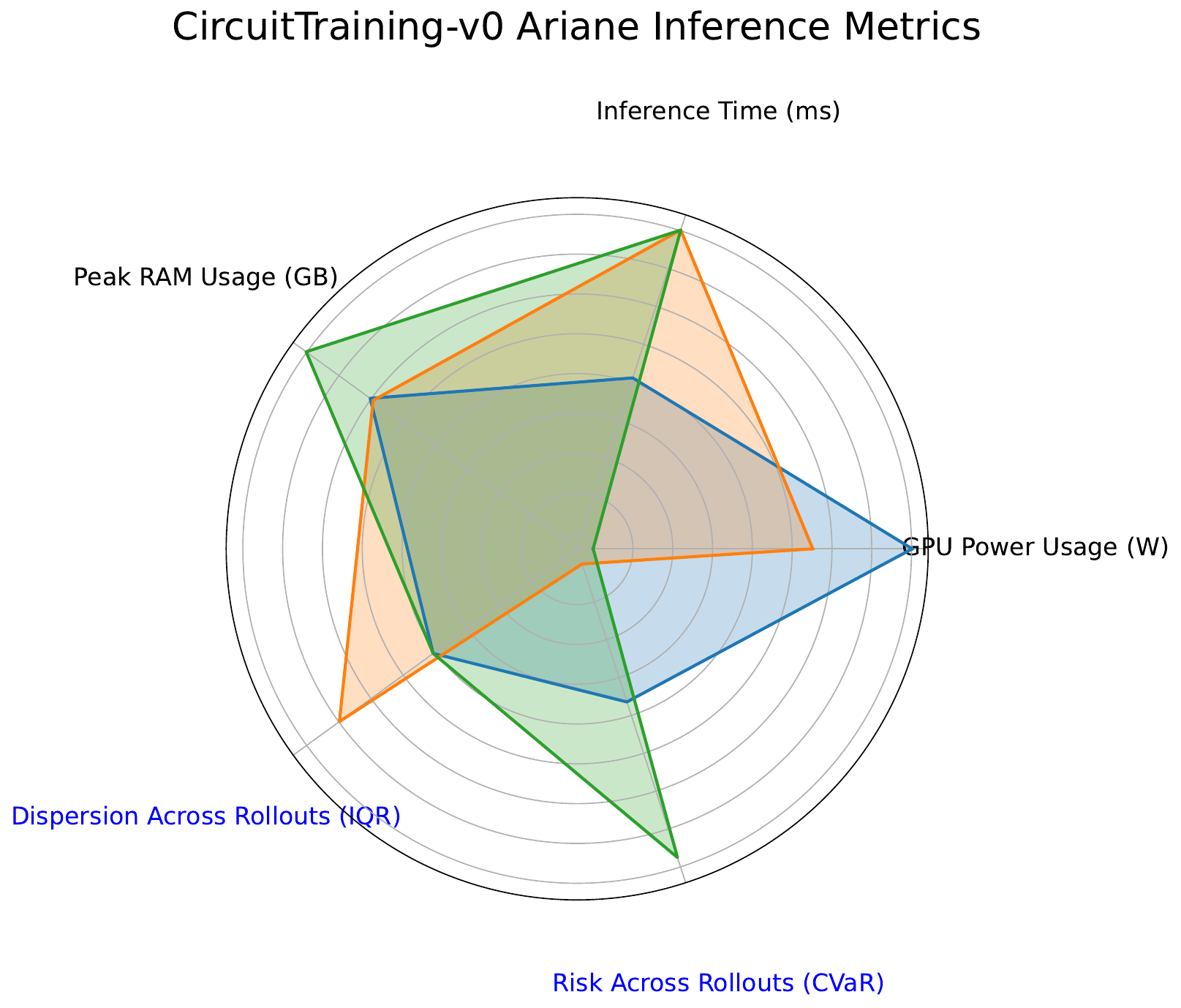}
  \end{minipage}
  \caption{Graphical representation of metrics for the "Ariane" netlist task of CircuitTraining-v0}
  \label{fig:appendix_circuit_training_ariane}
\end{figure}
\newpage

\begin{figure}[!htbp]
  \centering
  \begin{minipage}[b]{0.53\linewidth}
    \includegraphics[width=\linewidth]{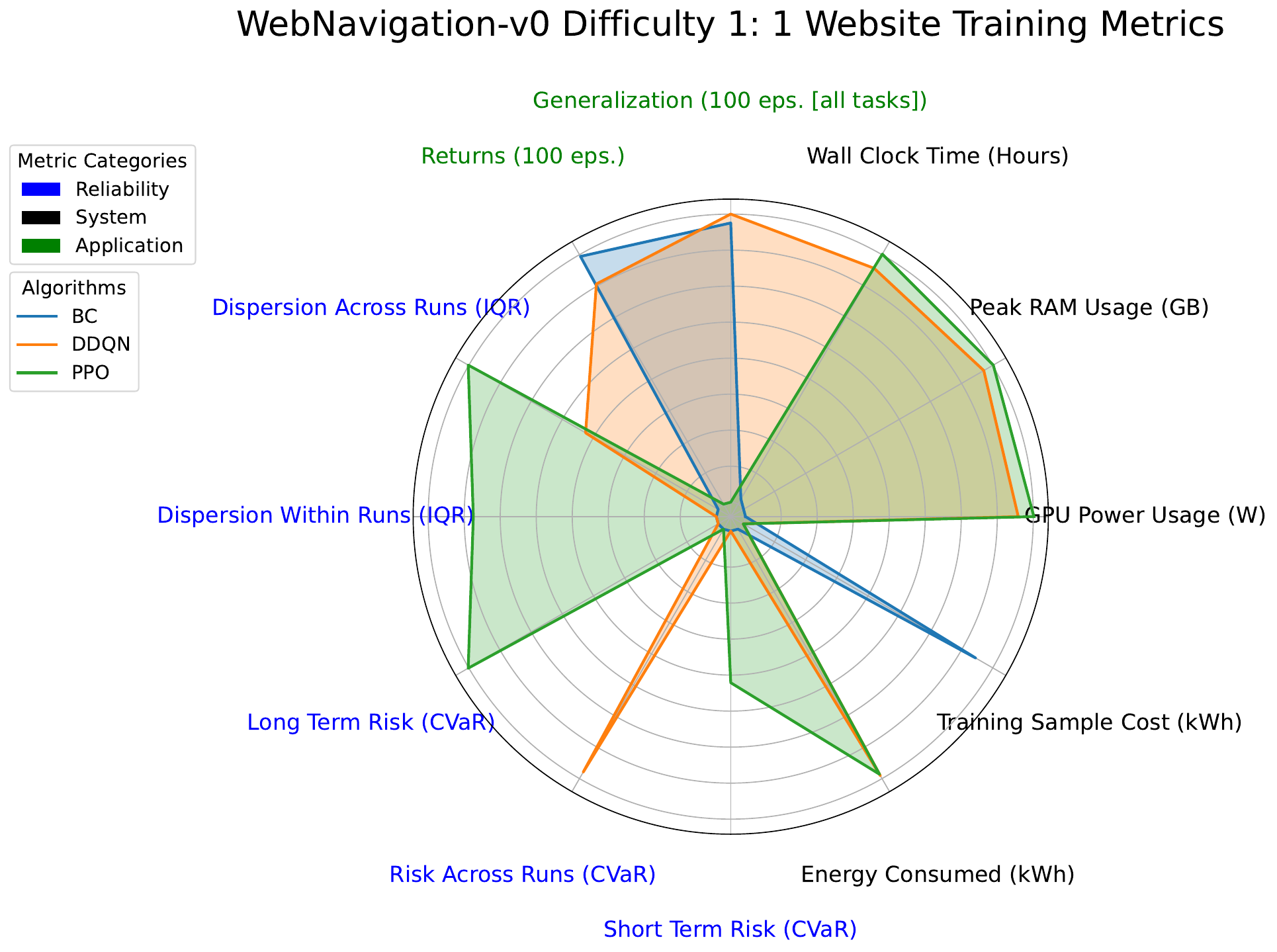}
  \end{minipage}
  \hfill 
  \begin{minipage}[b]{0.4625\linewidth}
    \includegraphics[width=\linewidth]{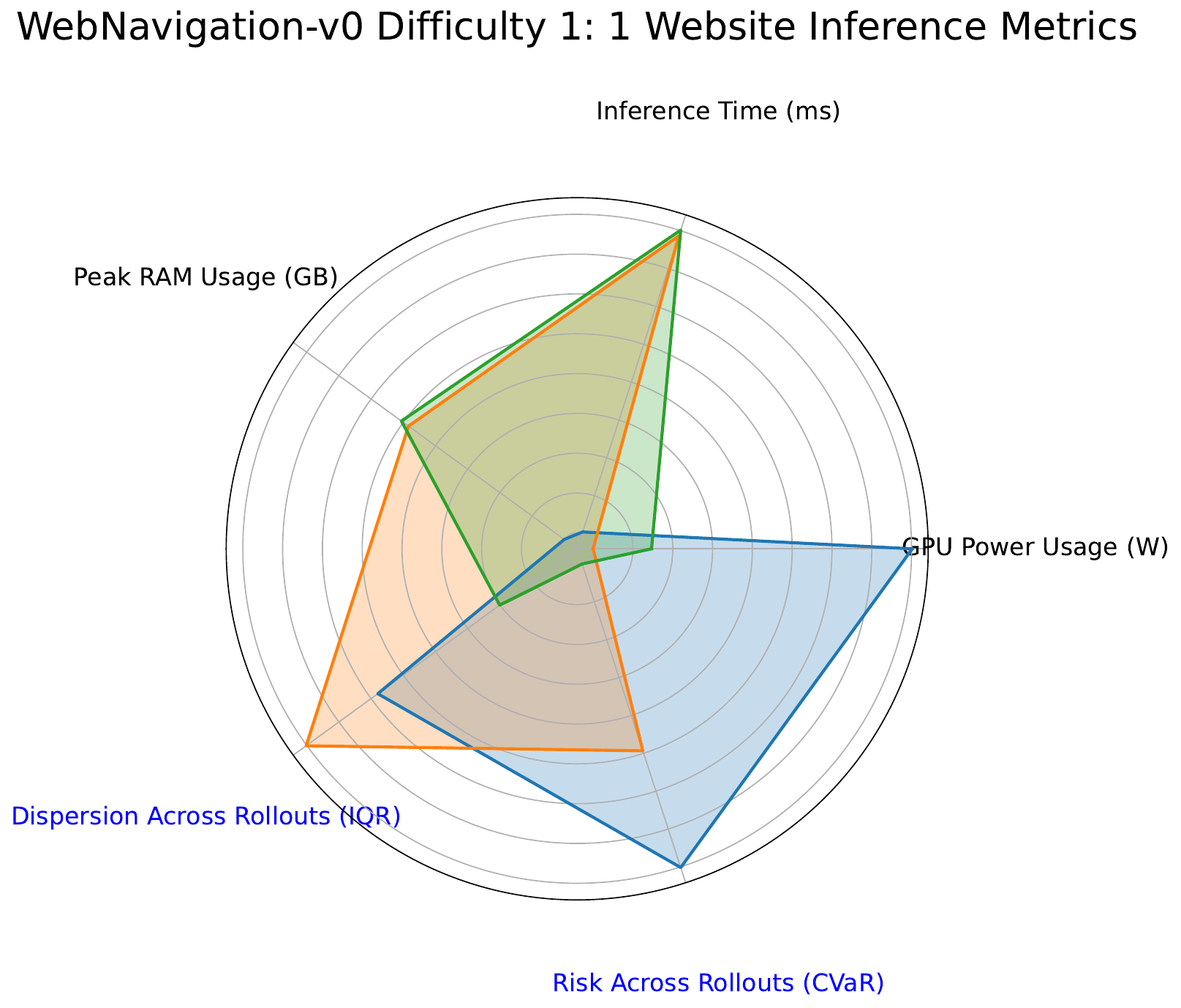}
  \end{minipage}
  \caption{Graphical representation of metrics for the "difficulty 1, 1 website" task of WebNavigation-v0}
  \label{fig:appendix_radar_web_nav_difficulty1_1website}
\end{figure}

\begin{figure}[!htbp]
  \centering
  \begin{minipage}[b]{0.53\linewidth}
    \includegraphics[width=\linewidth]{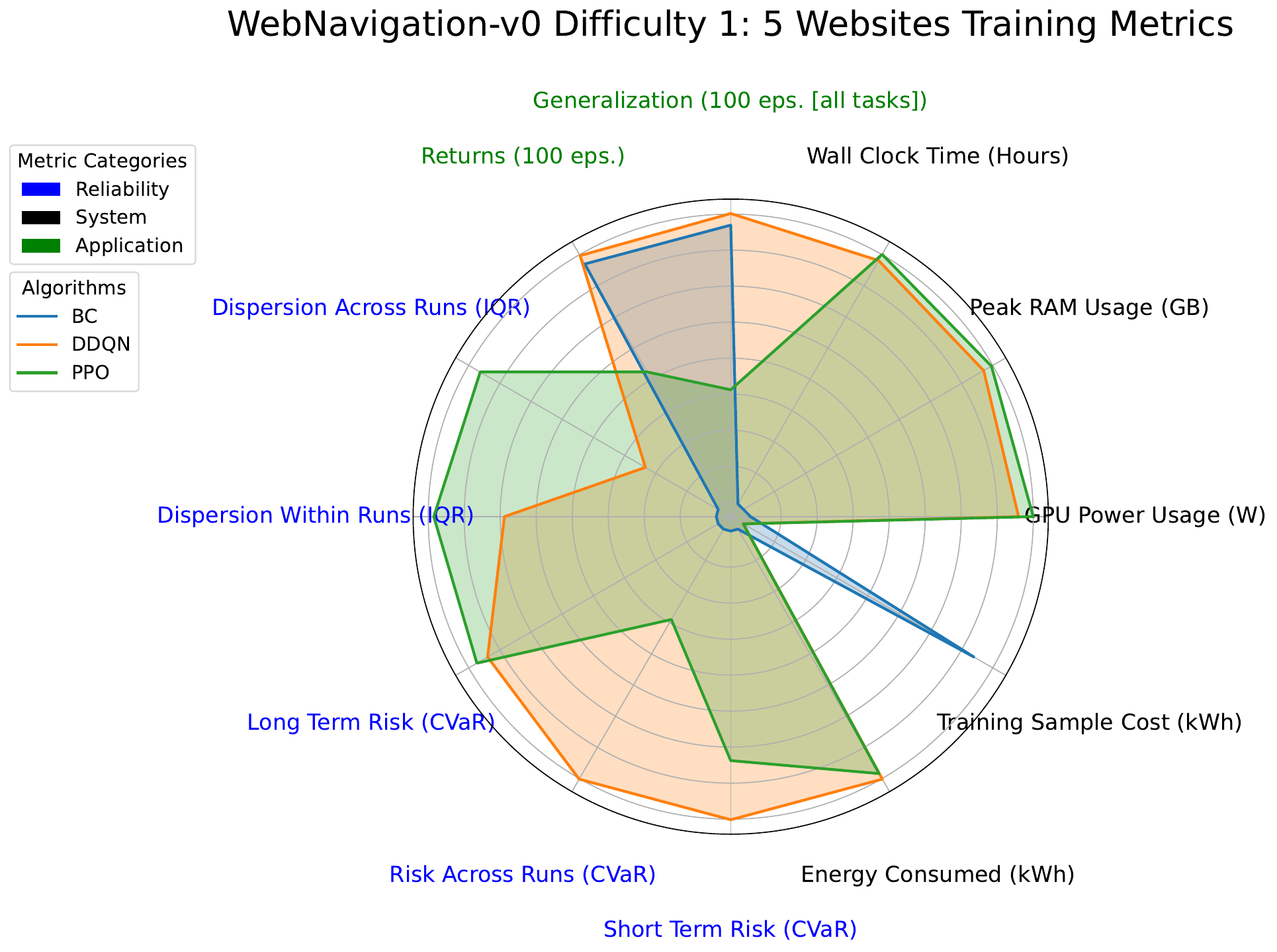}
  \end{minipage}
  \hfill 
  \begin{minipage}[b]{0.4625\linewidth}
    \includegraphics[width=\linewidth]{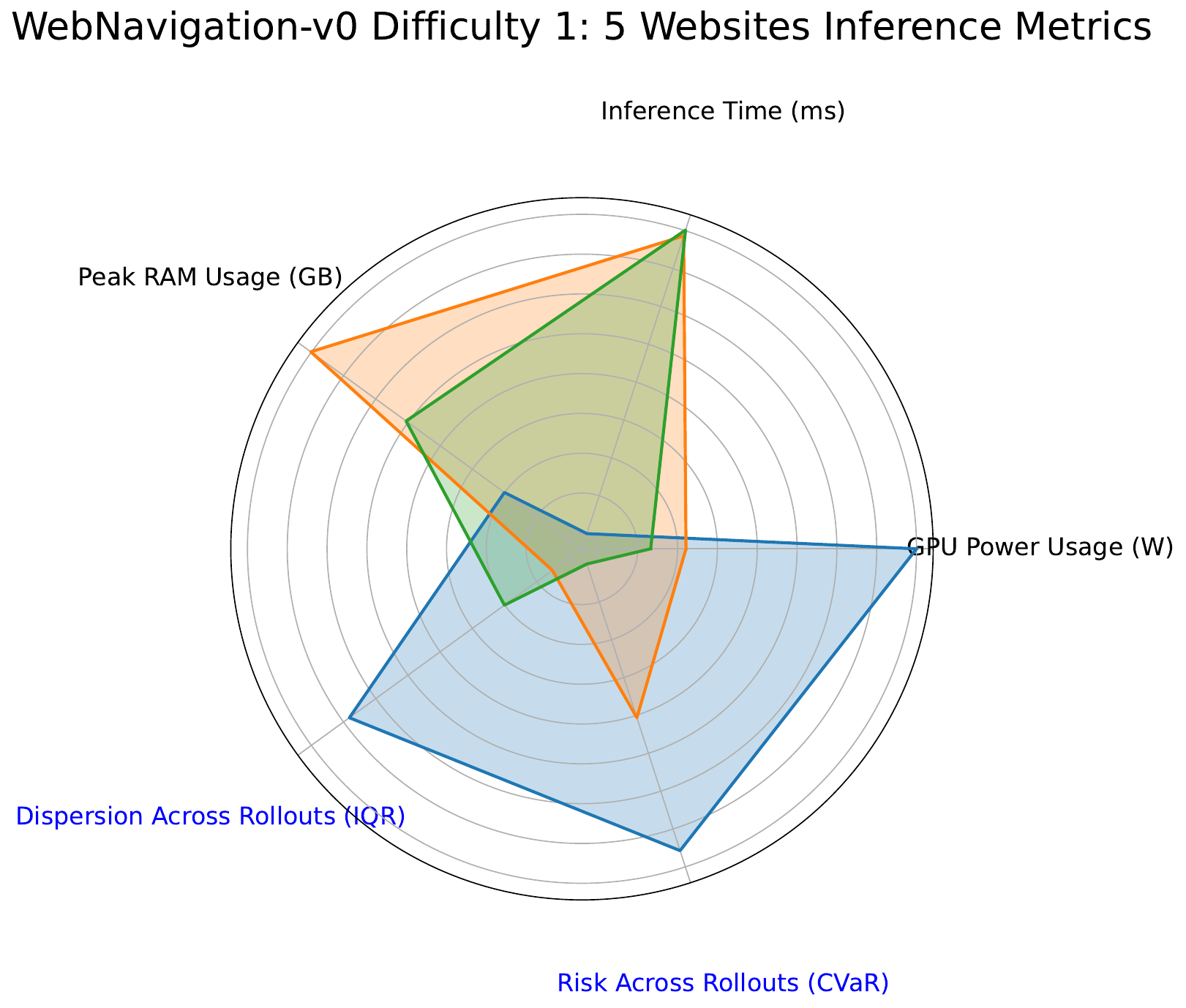}
  \end{minipage}
  \caption{Graphical representation of metrics for the "difficulty 1, 5 websites" task of WebNavigation-v0}
  \label{fig:appendix_radar_web_nav_difficulty1_5websites}
\end{figure}

\begin{figure}[!htbp]
  \centering
  \begin{minipage}[b]{0.53\linewidth}
    \includegraphics[width=\linewidth]{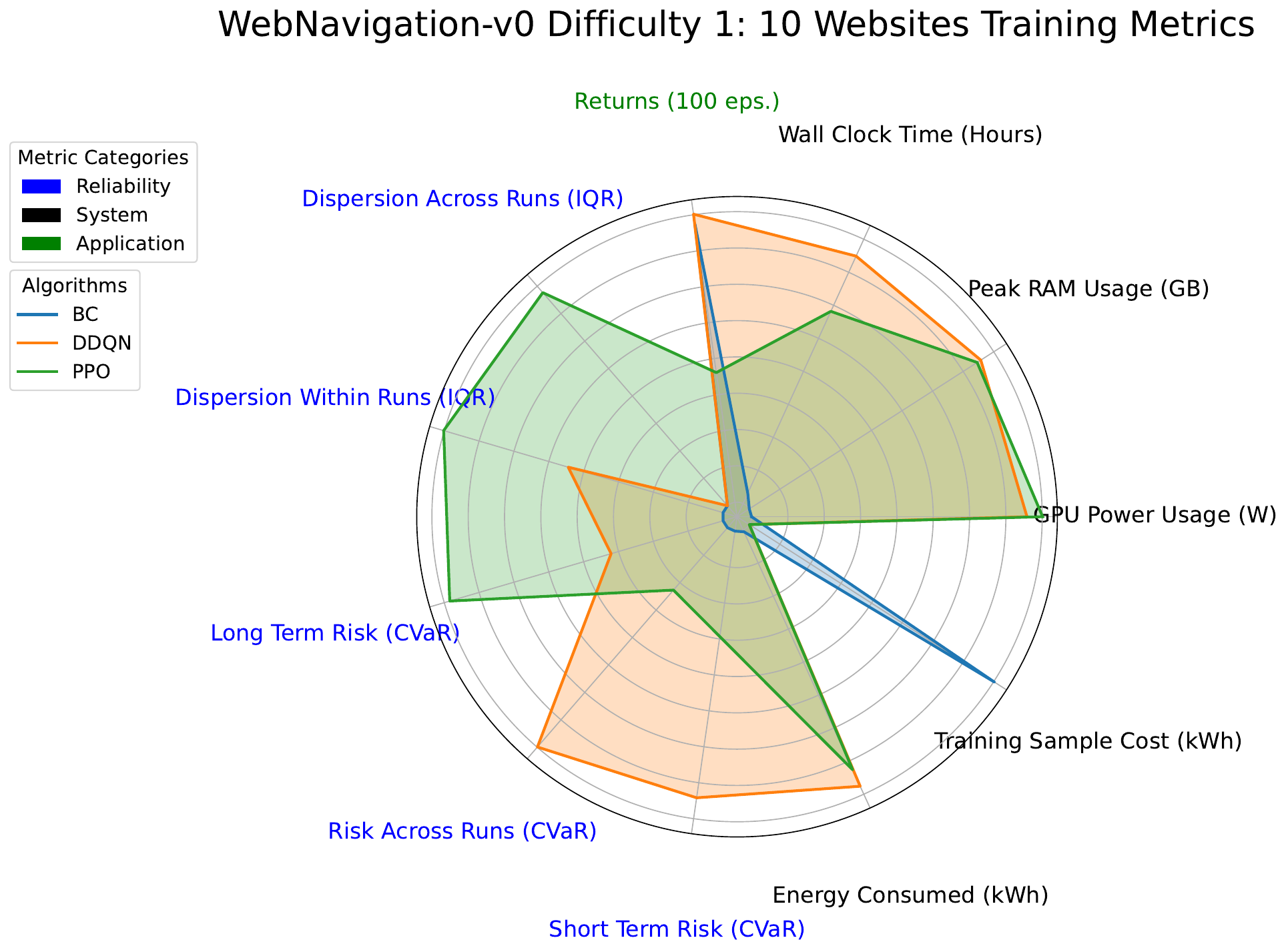}
  \end{minipage}
  \hfill 
  \begin{minipage}[b]{0.4625\linewidth}
    \includegraphics[width=\linewidth]{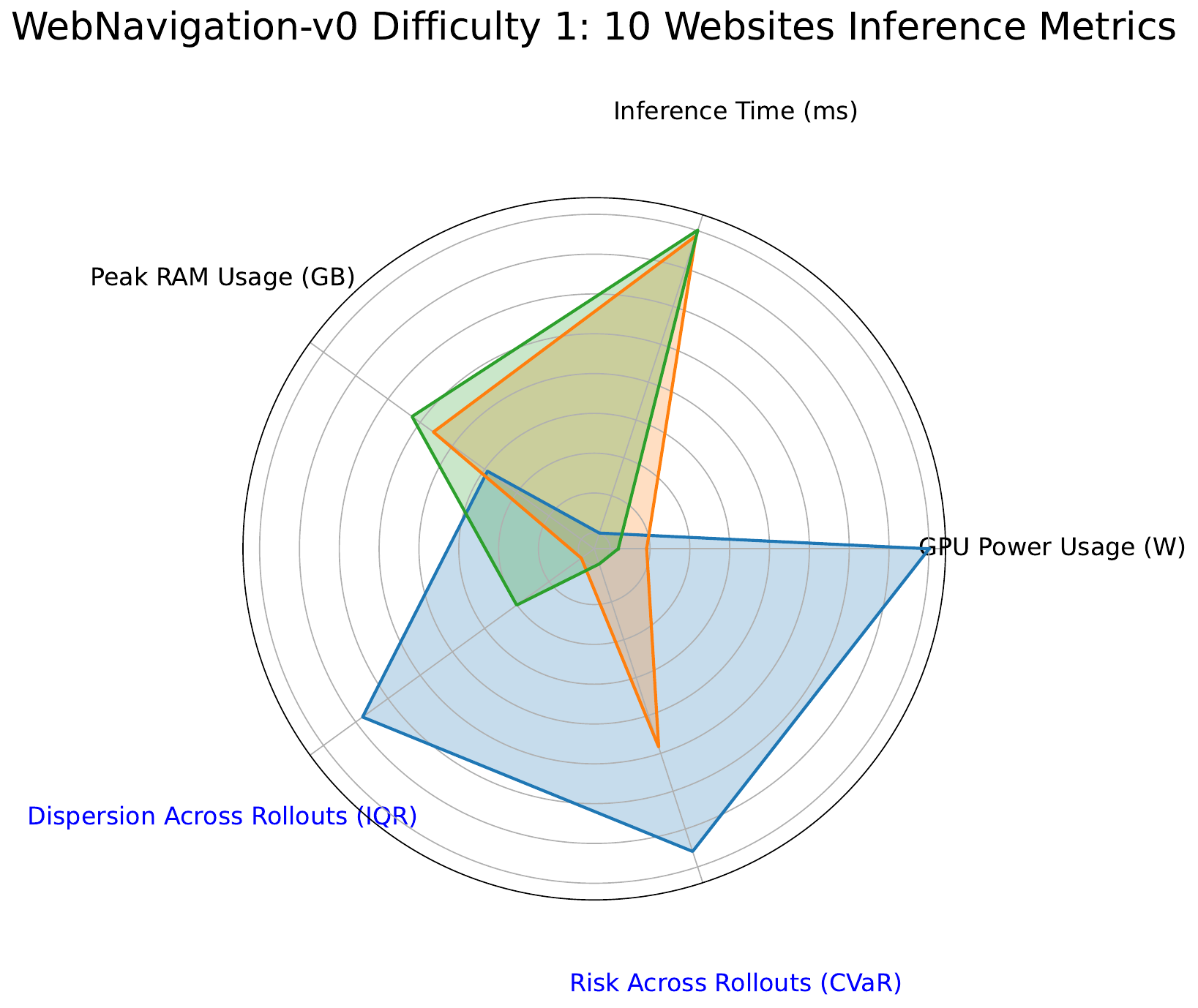}
  \end{minipage}
  \caption{Graphical representation of metrics for the "difficulty 1, 10 websites" task of WebNavigation-v0}
  \label{fig:appendix_radar_web_nav_difficulty1_10websites}
\end{figure}

%% file: appendix/experimental_setup.tex
\subsection{Training}
We used the Tensorflow Agents \cite{TFAgents} library to conduct distributed reinforcement learning experiments across the three domains: computer chip floorplanning, web navigation, and quadruped locomotion.
Our training setup consisted of one training server (a Google Cloud a2-highgpu-8g instance\protect\footnotemark[1]) equipped with four NVIDIA A100 GPUs, and multiple collect servers (Google Cloud n2-standard-96 instances\protect\footnotemark[2]) with 96 vCPUs running in parallel.

\footnotetext[1]{\href{https://cloud.google.com/compute/docs/gpus\#a100-gpus}{cloud.google.com/compute/docs/gpus}}

\footnotetext[2]{\href{https://cloud.google.com/compute/docs/general-purpose-machines\#n2-standard}{cloud.google.com/compute/docs/general-purpose-machines}}

The number of collect jobs running simultaneously varied depending on the specific domain and the available resources (such as CPU and memory) on the collect machines, which are important for running the environments efficiently. When using a collect machine with 96 vCPUs, we adjusted the number of environment instances based on the computational requirements of each domain:

\begin{enumerate}
\item \textbf{Quadruped Locomotion}: With 96 vCPUs on the collect machine, we ran 44 quadruped locomotion environment instances concurrently using Python 3.9.
\item \textbf{Computer Chip Floorplanning}: For the computer chip floorplanning domain, we ran 25 computer chip floorplanning environment instances on a collect machine with 96 vCPUs using Python 3.10.
\item \textbf{Web Navigation}: When running web navigation experiments on a collect machine with 96 vCPUs, we instantiated 40 web navigation environment instances simultaneously using Python 3.10.
\end{enumerate}

The behavioral cloning experiments for all three domains used the same setup as the online training experiments, with one training server equipped with four A100 GPUs.

\subsection{Inference}
For the inference phase, we used a single machine equipped with one NVIDIA V100 GPU to evaluate the trained models across all three domains: computer chip floorplanning, web navigation, and quadruped locomotion. The difference in hardware between the training and inference setups does not affect the application performance metrics, as these metrics are independent of the hardware and reflect the effectiveness of the trained models. However, the system performance metrics, such as inference time and memory usage, may vary depending on the specific hardware used during inference.

%% file: appendix/hyperparameters.tex
\begin{table}[H]
\centering
\begin{tabular}{|l|l|l|l|}
\hline
\textbf{Hyperparameter} & \textbf{BC} & \textbf{PPO} & \textbf{DDQN} \\ \hline
\multicolumn{4}{|c|}{\textbf{Toy Macro Standard Cell}} \\ \hline
Batch Size & 64 & 128 & 256 \\ \hline
Learning Rate & 1e-4 & 4e-4 & 4e-5 \\ \hline
Environment Batch Size & - & 512 & 512 \\ \hline
Number of Epochs & - & 6 & - \\ \hline
Number of Iterations & 200 & 5000 & 10000 \\ \hline
Entropy Regularization & - & 1e-2 & - \\ \hline
Number of Episodes Per Iteration & - & 32 & - \\ \hline
Epsilon Greedy & - & - & 0.3 \\ \hline
Replay Buffer Capacity & - & - & 1000000 \\ \hline
\multicolumn{4}{|c|}{\textbf{Ariane}} \\ \hline
Batch Size & 64 & 128 & 256 \\ \hline
Learning Rate & 1e-4 & 4e-4 & 4e-5 \\ \hline
Environment Batch Size & - & 512 & 512 \\ \hline
Number of Epochs & - & 4 & - \\ \hline
Number of Iterations & 200 & 250 & 100000 \\ \hline
Entropy Regularization & - & 1e-2 & - \\ \hline
Number of Episodes Per Iteration & - & 1024 & - \\ \hline
Epsilon Greedy & - & - & 0.3 \\ \hline
Replay Buffer Capacity & - & - & 10000000 \\ \hline
\end{tabular}
\vspace{0.5em}
\caption{Circuit Training Hyperparameters}

\end{table}

\begin{table}[H]
\centering
\begin{tabular}{|l|l|l|l|}
\hline
\textbf{Hyperparameter} & \textbf{BC} & \textbf{PPO} & \textbf{DDQN} \\ \hline
Batch Size & 128 & 128 & 128 \\ \hline
Learning Rate & 1e-4 & 3e-6 & 3e-6 \\ \hline
Entropy Regularization & - & 1e-2 & - \\ \hline
Number of Episodes Per Iteration & - & 512 & - \\ \hline
Environment Batch Size & - & 512 & 512 \\ \hline
Number of Epochs & - & 4 & - \\ \hline
Number of Iterations & 5000 & 200 & 50000 \\ \hline
Epsilon Greedy & - & - & 0.3 \\ \hline
Replay Buffer Capacity & - & - & 1000000 \\ \hline
Maximum Vocabulary Size & 500 & 500 & 500 \\ \hline
Latent Dimension & 50 & 50 & 50 \\ \hline
Embedding Dimension & 100 & 100 & 100 \\ \hline
Profile Value Dropout & 1.0 & 1.0 & 1.0 \\ \hline
\end{tabular}
\vspace{0.5em}
\caption{Web Navigation Hyperparameters}
\end{table}

\begin{table}[H]
\centering
\begin{tabular}{|l|l|l|l|}
\hline
\textbf{Hyperparameter} & \textbf{BC} & \textbf{PPO} & \textbf{SAC} \\ \hline
Batch Size & 64 & 128 & 256 \\ \hline
Learning Rate & 1e-4 & 1e-5 & 3e-4 \\ \hline
Environment Batch Size & - & 512 & 512 \\ \hline
Number of Epochs & - & 4 & - \\ \hline
Number of Iterations & 1000 & 8000 & 2000000 \\ \hline
Entropy Regularization & - & 1e-2 & - \\ \hline
Number of Episodes Per Iteration & - & 512 & - \\ \hline
Replay Buffer Capacity & - & - & 2000000 \\ \hline
\end{tabular}
\vspace{0.5em}
\caption{Quadruped Locomotion Hyperparameters}

\end{table}

%% file: appendix/dataset_collection.tex
To collect datasets for each domain and task, we periodically saved the policies at fixed intervals throughout the training process. We then evaluated all the saved policies on 100 episodes for each domain and task.
Based on these evaluations, we created a distribution of median returns and assigned an \texttt{expertise} level to each policy as follows:
\begin{enumerate}
\item \texttt{Novice}: The median return lies within one standard deviation below the mean.
\item \texttt{Intermediate}: The median return is within one standard deviation above or below the mean.
\item \texttt{Expert}: The median return is one standard deviation above the mean or higher.
\end{enumerate}
In some cases, certain domains or tasks were too challenging, resulting in no policies of a given skill level. In such instances, we only provide a \texttt{novice} dataset.

%% file: appendix/dataset_information.tex
\begin{enumerate}
\item Dataset documentation and intended uses:
\begin{itemize}
\item The A2Perf datasets consist of data collected from three simulated environments: computer chip floorplanning, web navigation, and quadruped locomotion. The data was generated by running reinforcement learning policies at various stages of training, capturing the experiences of these policies interacting with the respective environments. The datasets are intended for use in offline reinforcement learning, imitation learning, and hybrid approaches, allowing researchers to evaluate and compare different algorithms without the need for online data collection.
\end{itemize}
\item Dataset availability:

\begin{itemize}
\item The datasets can be accessed at:
\begin{itemize}
\item Circuit Training: \url{https://drive.google.com/drive/folders/1UMhLlnYmfbnjBPN_JwVy4YXDUahXrWf6}
\item Quadruped Locomotion: \url{https://drive.google.com/drive/folders/1n1BJFip-reSPif8Bv3jXAnSOgfQAEje7}
\item Web Navigation: \url{https://drive.google.com/drive/folders/13EmCscVatl7Q5EFdWFRpwKlA2yRfonE5}
\end{itemize}
\end{itemize}

\item Data format and usage:
\begin{itemize}
\item The datasets are provided in the widely-used HDF5 format, a data model and file format designed for efficient storage and retrieval of large datasets. Detailed instructions on how to read and use the data with the Minari framework are provided at: \url{https://minari.farama.org/}
\end{itemize}
\item Licensing:
\begin{itemize}
\item The A2Perf datasets are released under the MIT License. The authors confirm that they bear all responsibility in case of violation of rights.
\end{itemize}
\item Maintenance and long-term preservation:
\begin{itemize}
\item The datasets are hosted on a Google Cloud Bucket maintained by the Farama Foundation, a non-profit organization dedicated to supporting open-source machine learning projects. This ensures the long-term availability and accessibility of the datasets for the research community.
\end{itemize}
\end{enumerate}

%% file: appendix/software_usage.tex
A2Perf is a benchmark harness designed to be used flexibly on various machines. The user has the option to either run it in a Docker container or to run the benchmark locally. A Docker container is available in the harness and can be adapted to your needs. If you would like to run the benchmark locally, a guide is available to install the A2Perf benchmark harness on your Linux or MacOS system. While you can benchmark on both operating systems, it is important to note that system performance metrics are tracked using CodeCarbon. This allows to capture energy, power and memory usage at regular time intervals, and uses pyRAPL to compute the Running Average Power Limit (RAPL). However, RAPL uniquely measures power consumption information for Intel CPUs, DRAM for server architectures and GPU for client architectures. When using systems using CPU architectures different then the Intel CPUs, the power consumption metric will return a computed estimate rather than a measured metric.

The benchmark harness allows you to benchmark both the training and inference of your algorithm and agents respectively. In order to benchmark your algorithms, you need to create a submission folder which includes several files which A2Perf calls. First, a training file, \texttt{train.py} contains a function \texttt{train()}, which starts the training process of your algorithm when called. Similarly, \texttt{inference.py} covers the inference of your trained model. This file includes several functions responsible for the loading of your trained model, preprocessing observations and running inference on your model. Using a \texttt{requirements.txt} file, additional Python packages and versioning can be specified. Running the benchmark is done through a command line interface. Using flags, we can pass additional information to the submission to set up the benchmark. A \texttt{gin\_config} flag allows the user to define the settings for your environment and training process. Additionally, we need to pass the path to the submission folder using the \texttt{participant\_module\_path} flag. For a more detailed description, tutorials are available in the repository.

%% file: appendix/website_generation.tex
To create environments for the web navigation tasks, we generate synthetic websites that agents must learn to navigate. These websites serve as training and evaluation environments, where agents need to fill forms and interact with various web elements. Here we describe our procedural website generation process.

To generate the set of websites $W$, we first assume a target number of websites, denoted as $N_{\text{websites}}$.
Following the approach in \citet{gur2021environment} (shown in Table 4 of the paper), we consider 42 possible primitives that can be added to a web page and introduce two additional primitives: a "new page" primitive and a "stop" primitive, resulting in a total of 44 primitives.

The website generation process begins with an empty web page.
We repeatedly sample uniformly from the 44 primitives and add them to the current page.
If the "new page" primitive is selected during the sampling process, we start adding primitives to a new linked page.
If the "stop" primitive is selected, we conclude the generation of the current website and proceed to generate the next website, if necessary.
This process continues until we have generated the desired number of websites, $N_{\text{websites}}$.
Each website in the resulting set $W$ consists of one or more web pages, with each page containing a sampled set of primitives.

We define the difficulty of a web page as the probability of a random agent interacting with the correct primitive(s).
The difficulty of page $p_i$ is given by $-\log \left (\frac{n_{\text{active}}}{n_\text{active}+n_\text{passive}} \right )$, where $n_{\text{active}}$ and $n_{\text{passive}}$ denote the number of active and passive primitives on the page, respectively.
The difficulty of an entire sequence of web pages is determined by summing the difficulty of all individual pages it contains.
Based on these difficulty calculations, we partition the websites into three difficulty levels.
The three levels of difficulty correspond to the probability thresholds of 50\%, 25\%, and 10\% for levels 1, 2, and 3, respectively.
Users can select a specific difficulty level of web navigation by executing Python commands such as \verb|env = gym.make("WebNavigation-Difficulty-01-v0", num_websites=1)|, where the \verb|num_websites| argument defines the pool of websites available for the environment. During training or evaluation, each episode begins by randomly selecting one website from this pool at the specified difficulty level. For example, if \verb|num_websites=10|, the environment will generate 10 websites at the specified difficulty level, and each episode will randomly assign one of these websites for the agent to navigate.
At each timestep, the agent can interact with an HTML element on the page, such as modifying the text field or clicking on the element, with the objective of entering correct information into forms and clicking "next" or "submit" to advance between web pages.

%% file: main.bbl
\begin{thebibliography}{73}
\providecommand{\natexlab}[1]{#1}
\providecommand{\url}[1]{\texttt{#1}}
\expandafter\ifx\csname urlstyle\endcsname\relax
  \providecommand{\doi}[1]{doi: #1}\else
  \providecommand{\doi}{doi: \begingroup \urlstyle{rm}\Url}\fi

\bibitem[Abadi et~al.(2016)Abadi, Barham, Chen, Chen, Davis, Dean, Devin, Ghemawat, Irving, Isard, et~al.]{abadi2016tensorflow}
Mart{\'\i}n Abadi, Paul Barham, Jianmin Chen, Zhifeng Chen, Andy Davis, Jeffrey Dean, Matthieu Devin, Sanjay Ghemawat, Geoffrey Irving, Michael Isard, et~al.
\newblock $\{$TensorFlow$\}$: a system for $\{$Large-Scale$\}$ machine learning.
\newblock In \emph{12th USENIX symposium on operating systems design and implementation (OSDI 16)}, pp.\  265--283, 2016.

\bibitem[Agarwal et~al.(2021)Agarwal, Schwarzer, Castro, Courville, and Bellemare]{agarwal2021deep_rl_stat_precipice}
Rishabh Agarwal, Max Schwarzer, Pablo~Samuel Castro, Aaron~C Courville, and Marc Bellemare.
\newblock Deep reinforcement learning at the edge of the statistical precipice.
\newblock \emph{Advances in neural information processing systems}, 34:\penalty0 29304--29320, 2021.

\bibitem[Ball et~al.(2023)Ball, Smith, Kostrikov, and Levine]{ball2023efficient}
Philip~J Ball, Laura Smith, Ilya Kostrikov, and Sergey Levine.
\newblock Efficient online reinforcement learning with offline data.
\newblock In \emph{International Conference on Machine Learning}, pp.\  1577--1594. PMLR, 2023.

\bibitem[Bansal et~al.(2018)Bansal, Krizhevsky, and Ogale]{bansal2018chauffeurnet}
Mayank Bansal, Alex Krizhevsky, and Abhijit Ogale.
\newblock Chauffeurnet: Learning to drive by imitating the best and synthesizing the worst.
\newblock \emph{arXiv preprint arXiv:1812.03079}, 2018.

\bibitem[Bellemare et~al.(2013)Bellemare, Naddaf, Veness, and Bowling]{bellemare2013arcade}
Marc~G Bellemare, Yavar Naddaf, Joel Veness, and Michael Bowling.
\newblock The arcade learning environment: An evaluation platform for general agents.
\newblock \emph{Journal of Artificial Intelligence Research}, 47:\penalty0 253--279, 2013.

\bibitem[Bonnet et~al.(2023)Bonnet, Luo, Byrne, Surana, Abramowitz, Duckworth, Coyette, Midgley, Tegegn, Kalloniatis, et~al.]{bonnet2023jumanji}
Cl{\'e}ment Bonnet, Daniel Luo, Donal Byrne, Shikha Surana, Sasha Abramowitz, Paul Duckworth, Vincent Coyette, Laurence~I Midgley, Elshadai Tegegn, Tristan Kalloniatis, et~al.
\newblock Jumanji: a diverse suite of scalable reinforcement learning environments in jax.
\newblock \emph{arXiv preprint arXiv:2306.09884}, 2023.

\bibitem[Bradbury et~al.(2018)Bradbury, Frostig, Hawkins, Johnson, Leary, Maclaurin, Necula, Paszke, Vander{P}las, Wanderman-{M}ilne, and Zhang]{jax2018github}
James Bradbury, Roy Frostig, Peter Hawkins, Matthew~James Johnson, Chris Leary, Dougal Maclaurin, George Necula, Adam Paszke, Jake Vander{P}las, Skye Wanderman-{M}ilne, and Qiao Zhang.
\newblock {JAX}: composable transformations of {P}ython+{N}um{P}y programs.
\newblock \url{http://github.com/google/jax}, 2018.
\newblock Version 0.3.13.

\bibitem[Brockman et~al.(2016)Brockman, Cheung, Pettersson, Schneider, Schulman, Tang, and Zaremba]{brockman2016openai}
Greg Brockman, Vicki Cheung, Ludwig Pettersson, Jonas Schneider, John Schulman, Jie Tang, and Wojciech Zaremba.
\newblock Openai gym.
\newblock \emph{arXiv preprint arXiv:1606.01540}, 2016.

\bibitem[Brohan et~al.(2022)Brohan, Brown, Carbajal, Chebotar, Dabis, Finn, Gopalakrishnan, Hausman, Herzog, Hsu, et~al.]{robotics_transformer}
Anthony Brohan, Noah Brown, Justice Carbajal, Yevgen Chebotar, Joseph Dabis, Chelsea Finn, Keerthana Gopalakrishnan, Karol Hausman, Alex Herzog, Jasmine Hsu, et~al.
\newblock Rt-1: Robotics transformer for real-world control at scale.
\newblock \emph{arXiv preprint arXiv:2212.06817}, 2022.

\bibitem[Brown et~al.(2020)Brown, Mann, Ryder, Subbiah, Kaplan, Dhariwal, Neelakantan, Shyam, Sastry, Askell, et~al.]{brown2020language}
Tom~B Brown, Benjamin Mann, Nick Ryder, Melanie Subbiah, Jared Kaplan, Prafulla Dhariwal, Arvind Neelakantan, Pranav Shyam, Girish Sastry, Amanda Askell, et~al.
\newblock Language models are few-shot learners.
\newblock \emph{arXiv preprint arXiv:2005.14165}, 2020.

\bibitem[Chan et~al.(2019)Chan, Fishman, Canny, Korattikara, and Guadarrama]{chan2019measuring}
Stephanie~CY Chan, Samuel Fishman, John Canny, Anoop Korattikara, and Sergio Guadarrama.
\newblock Measuring the reliability of reinforcement learning algorithms.
\newblock \emph{arXiv preprint arXiv:1912.05663}, 2019.

\bibitem[Chevalier-Boisvert et~al.(2023)Chevalier-Boisvert, Dai, Towers, de~Lazcano, Willems, Lahlou, Pal, Castro, and Terry]{MinigridMiniworld23}
Maxime Chevalier-Boisvert, Bolun Dai, Mark Towers, Rodrigo de~Lazcano, Lucas Willems, Salem Lahlou, Suman Pal, Pablo~Samuel Castro, and Jordan Terry.
\newblock Minigrid \& miniworld: Modular \& customizable reinforcement learning environments for goal-oriented tasks.
\newblock \emph{CoRR}, abs/2306.13831, 2023.

\bibitem[Cobbe et~al.(2019)Cobbe, Klimov, Hesse, Kim, and Schulman]{coin_run}
Karl Cobbe, Oleg Klimov, Chris Hesse, Taehoon Kim, and John Schulman.
\newblock Quantifying generalization in reinforcement learning.
\newblock In \emph{International Conference on Machine Learning}, pp.\  1282--1289. PMLR, 2019.

\bibitem[Colas et~al.(2018)Colas, Sigaud, and Oudeyer]{colas2018many}
C{\'e}dric Colas, Olivier Sigaud, and Pierre-Yves Oudeyer.
\newblock How many random seeds? statistical power analysis in deep reinforcement learning experiments.
\newblock \emph{arXiv preprint arXiv:1806.08295}, 2018.

\bibitem[Coleman et~al.(2017)Coleman, Narayanan, Kang, Zhao, Zhang, Nardi, Bailis, Olukotun, R{\'e}, and Zaharia]{Coleman2017DAWNBenchA}
Cody Coleman, Deepak Narayanan, Daniel Kang, Tian Zhao, Jian Zhang, Luigi Nardi, Peter Bailis, Kunle Olukotun, Chris R{\'e}, and Matei Zaharia.
\newblock Dawnbench: An end-to-end deep learning benchmark and competition.
\newblock \emph{Training}, 100\penalty0 (101):\penalty0 102, 2017.

\bibitem[Coumans(2023)]{coumans2023motion}
Erwin Coumans.
\newblock Motion imitation, 2023.
\newblock URL \url{https://github.com/erwincoumans/motion_imitation}.

\bibitem[Dulac-Arnold et~al.(2021)Dulac-Arnold, Levine, Mankowitz, Li, Paduraru, Gowal, and Hester]{dulac2021challenges}
Gabriel Dulac-Arnold, Nir Levine, Daniel~J Mankowitz, Jerry Li, Cosmin Paduraru, Sven Gowal, and Todd Hester.
\newblock Challenges of real-world reinforcement learning: definitions, benchmarks and analysis.
\newblock \emph{Machine Learning}, 110\penalty0 (9):\penalty0 2419--2468, 2021.

\bibitem[Frey et~al.(2022)Frey, Li, McDonald, Zhao, Jones, Bestor, Tiwari, Gadepally, and Samsi]{frey2022benchmarking}
Nathan~C. Frey, Baolin Li, Joseph McDonald, Dan Zhao, Michael Jones, David Bestor, Devesh Tiwari, Vijay Gadepally, and Siddharth Samsi.
\newblock Benchmarking resource usage for efficient distributed deep learning, 2022.

\bibitem[Fu et~al.(2020)Fu, Kumar, Nachum, Tucker, and Levine]{fu2020d4rl}
Justin Fu, Aviral Kumar, Ofir Nachum, George Tucker, and Sergey Levine.
\newblock D4rl: Datasets for deep data-driven reinforcement learning.
\newblock \emph{arXiv preprint arXiv:2004.07219}, 2020.

\bibitem[Fujimoto et~al.(2019)Fujimoto, Meger, and Precup]{fujimoto2019off}
Scott Fujimoto, David Meger, and Doina Precup.
\newblock Off-policy deep reinforcement learning without exploration.
\newblock In \emph{International conference on machine learning}, pp.\  2052--2062. PMLR, 2019.

\bibitem[Greaves et~al.(2021)Greaves, Candido, Dumoulin, Goroshin, Ponda, Bellemare, and Castro]{balloon_learning_env}
Joshua Greaves, Salvatore Candido, Vincent Dumoulin, Ross Goroshin, Sameera~S. Ponda, Marc~G. Bellemare, and Pablo~Samuel Castro.
\newblock {Balloon Learning Environment}, 12 2021.
\newblock URL \url{https://github.com/google/balloon-learning-environment}.

\bibitem[Guadarrama et~al.(2018)Guadarrama, Korattikara, Ramirez, Castro, Holly, Fishman, Wang, Gonina, Wu, Kokiopoulou, Sbaiz, Smith, Bartók, Berent, Harris, Vanhoucke, and Brevdo]{TFAgents}
Sergio Guadarrama, Anoop Korattikara, Oscar Ramirez, Pablo Castro, Ethan Holly, Sam Fishman, Ke~Wang, Ekaterina Gonina, Neal Wu, Efi Kokiopoulou, Luciano Sbaiz, Jamie Smith, Gábor Bartók, Jesse Berent, Chris Harris, Vincent Vanhoucke, and Eugene Brevdo.
\newblock {TF-Agents}: A library for reinforcement learning in tensorflow.
\newblock \url{https://github.com/tensorflow/agents}, 2018.
\newblock URL \url{https://github.com/tensorflow/agents}.
\newblock [Online; accessed 25-June-2019].

\bibitem[Guadarrama et~al.(2021)Guadarrama, Yue, Boyd, Jiang, Songhori, Tam, and Mirhoseini]{CircuitTraining2021}
Sergio Guadarrama, Summer Yue, Toby Boyd, Joe~Wenjie Jiang, Ebrahim Songhori, Terence Tam, and Azalia Mirhoseini.
\newblock {Circuit Training}: An open-source framework for generating chip floor plans with distributed deep reinforcement learning.
\newblock \url{https://github.com/google_research/circuit_training}, 2021.
\newblock URL \url{https://github.com/google_research/circuit_training}.
\newblock [Online; accessed 21-December-2021].

\bibitem[Gulcehre et~al.(2020)Gulcehre, Wang, Novikov, Paine, G{\'o}mez, Zolna, Agarwal, Merel, Mankowitz, Paduraru, et~al.]{rl_unplugged}
Caglar Gulcehre, Ziyu Wang, Alexander Novikov, Thomas Paine, Sergio G{\'o}mez, Konrad Zolna, Rishabh Agarwal, Josh~S Merel, Daniel~J Mankowitz, Cosmin Paduraru, et~al.
\newblock Rl unplugged: A suite of benchmarks for offline reinforcement learning.
\newblock \emph{Advances in Neural Information Processing Systems}, 33:\penalty0 7248--7259, 2020.

\bibitem[Gur et~al.(2018)Gur, Rueckert, Faust, and Hakkani-Tur]{gur2018learning_to_navigate_the_web}
Izzeddin Gur, Ulrich Rueckert, Aleksandra Faust, and Dilek Hakkani-Tur.
\newblock Learning to navigate the web.
\newblock \emph{arXiv preprint arXiv:1812.09195}, 2018.

\bibitem[Gur et~al.(2021)Gur, Jaques, Miao, Choi, Tiwari, Lee, and Faust]{gur2021environment}
Izzeddin Gur, Natasha Jaques, Yingjie Miao, Jongwook Choi, Manoj Tiwari, Honglak Lee, and Aleksandra Faust.
\newblock Environment generation for zero-shot compositional reinforcement learning.
\newblock \emph{Advances in Neural Information Processing Systems}, 34:\penalty0 4157--4169, 2021.

\bibitem[Gur et~al.(2022)Gur, Nachum, Miao, Safdari, Huang, Chowdhery, Narang, Fiedel, and Faust]{gur2022understanding_html_llm}
Izzeddin Gur, Ofir Nachum, Yingjie Miao, Mustafa Safdari, Austin Huang, Aakanksha Chowdhery, Sharan Narang, Noah Fiedel, and Aleksandra Faust.
\newblock Understanding html with large language models.
\newblock \emph{arXiv preprint arXiv:2210.03945}, 2022.

\bibitem[Guss et~al.(2019)Guss, Houghton, Topin, Wang, Codel, Veloso, and Salakhutdinov]{guss2019minerl}
William~H Guss, Brandon Houghton, Nicholay Topin, Phillip Wang, Cayden Codel, Manuela Veloso, and Ruslan Salakhutdinov.
\newblock Minerl: A large-scale dataset of minecraft demonstrations.
\newblock \emph{arXiv preprint arXiv:1907.13440}, 2019.

\bibitem[Haarnoja et~al.(2018)Haarnoja, Zhou, Abbeel, and Levine]{haarnoja2018soft_actor_critic}
Tuomas Haarnoja, Aurick Zhou, Pieter Abbeel, and Sergey Levine.
\newblock Soft actor-critic: Off-policy maximum entropy deep reinforcement learning with a stochastic actor.
\newblock In \emph{International conference on machine learning}, pp.\  1861--1870. PMLR, 2018.

\bibitem[Henderson et~al.(2018)Henderson, Islam, Bachman, Pineau, Precup, and Meger]{henderson2018deep_rl_that_matters}
Peter Henderson, Riashat Islam, Philip Bachman, Joelle Pineau, Doina Precup, and David Meger.
\newblock Deep reinforcement learning that matters.
\newblock In \emph{Proceedings of the AAAI conference on artificial intelligence}, volume~32, 2018.

\bibitem[Hester et~al.(2018)Hester, Vecerik, Pietquin, Lanctot, Schaul, Piot, Horgan, Quan, Sendonaris, Osband, et~al.]{hester2018deep}
Todd Hester, Matej Vecerik, Olivier Pietquin, Marc Lanctot, Tom Schaul, Bilal Piot, Dan Horgan, John Quan, Andrew Sendonaris, Ian Osband, et~al.
\newblock Deep q-learning from demonstrations.
\newblock In \emph{Proceedings of the AAAI conference on artificial intelligence}, volume~32, 2018.

\bibitem[Ho \& Ermon(2016)Ho and Ermon]{gail}
Jonathan Ho and Stefano Ermon.
\newblock Generative adversarial imitation learning.
\newblock \emph{Advances in neural information processing systems}, 29, 2016.

\bibitem[Initiative(2021)]{codecarbon}
MLCO2 Initiative.
\newblock Codecarbon.
\newblock \url{https://github.com/mlco2/codecarbon}, 2021.
\newblock Accessed: June 1, 2023.

\bibitem[Jang et~al.(2022)Jang, Irpan, Khansari, Kappler, Ebert, Lynch, Levine, and Finn]{bc_zero}
Eric Jang, Alex Irpan, Mohi Khansari, Daniel Kappler, Frederik Ebert, Corey Lynch, Sergey Levine, and Chelsea Finn.
\newblock Bc-z: Zero-shot task generalization with robotic imitation learning.
\newblock In \emph{Conference on Robot Learning}, pp.\  991--1002. PMLR, 2022.

\bibitem[Ji et~al.(2023)Ji, Zhang, Zhou, Pan, Huang, Sun, Geng, Zhong, Dai, and Yang]{ji2023safety_gym}
Jiaming Ji, Borong Zhang, Jiayi Zhou, Xuehai Pan, Weidong Huang, Ruiyang Sun, Yiran Geng, Yifan Zhong, Josef Dai, and Yaodong Yang.
\newblock Safety gymnasium: A unified safe reinforcement learning benchmark.
\newblock \emph{Advances in Neural Information Processing Systems}, 36, 2023.

\bibitem[Kelly et~al.(2019)Kelly, Sidrane, Driggs-Campbell, and Kochenderfer]{kelly2019hg}
Michael Kelly, Chelsea Sidrane, Katherine Driggs-Campbell, and Mykel~J Kochenderfer.
\newblock Hg-dagger: Interactive imitation learning with human experts.
\newblock In \emph{2019 International Conference on Robotics and Automation (ICRA)}, pp.\  8077--8083. IEEE, 2019.

\bibitem[Kempka et~al.(2016)Kempka, Wydmuch, Runc, Toczek, and Jaśkowski]{kempka2016vizdoom}
Michał Kempka, Marek Wydmuch, Grzegorz Runc, Jakub Toczek, and Wojciech Jaśkowski.
\newblock Vizdoom: A doom-based ai research platform for visual reinforcement learning, 2016.

\bibitem[Kostrikov et~al.(2021)Kostrikov, Nair, and Levine]{kostrikov2021offline}
Ilya Kostrikov, Ashvin Nair, and Sergey Levine.
\newblock Offline reinforcement learning with implicit q-learning.
\newblock \emph{arXiv preprint arXiv:2110.06169}, 2021.

\bibitem[Krishnan et~al.(2022)Krishnan, Lam, Chitlangia, Wan, Barth-Maron, Faust, and Reddi]{krishnan2022quarl}
Srivatsan Krishnan, Maximilian Lam, Sharad Chitlangia, Zishen Wan, Gabriel Barth-Maron, Aleksandra Faust, and Vijay~Janapa Reddi.
\newblock Quarl: Quantization for fast and environmentally sustainable reinforcement learning, 2022.

\bibitem[Lee et~al.(2021)Lee, Hu, and Lim]{lee2021ikea}
Youngwoon Lee, Edward~S Hu, and Joseph~J Lim.
\newblock Ikea furniture assembly environment for long-horizon complex manipulation tasks.
\newblock In \emph{2021 IEEE International Conference on Robotics and Automation (ICRA)}, pp.\  6343--6349. IEEE, 2021.

\bibitem[Levine et~al.(2020)Levine, Kumar, Tucker, and Fu]{levine2020offline}
Sergey Levine, Aviral Kumar, George Tucker, and Justin Fu.
\newblock Offline reinforcement learning: Tutorial, review, and perspectives on open problems.
\newblock \emph{arXiv preprint arXiv:2005.01643}, 2020.

\bibitem[Li et~al.(2023)Li, Tian, Tam, Ma, and Li]{li2023breaking}
Shitian Li, Chunlin Tian, Kahou Tam, Rui Ma, and Li~Li.
\newblock Breaking on-device training memory wall: A systematic survey, 2023.

\bibitem[Liu et~al.(2018)Liu, Guu, Pasupat, Shi, and Liang]{liu2018reinforcement}
Evan~Zheran Liu, Kelvin Guu, Panupong Pasupat, Tianlin Shi, and Percy Liang.
\newblock Reinforcement learning on web interfaces using workflow-guided exploration.
\newblock In \emph{International Conference on Learning Representations ({ICLR})}, 2018.
\newblock URL \url{https://arxiv.org/abs/1802.08802}.

\bibitem[Liu et~al.(2023)Liu, Guo, Lin, Yao, Zhu, Cen, Hu, Yu, Zhang, Tan, et~al.]{liu2023datasets}
Zuxin Liu, Zijian Guo, Haohong Lin, Yihang Yao, Jiacheng Zhu, Zhepeng Cen, Hanjiang Hu, Wenhao Yu, Tingnan Zhang, Jie Tan, et~al.
\newblock Datasets and benchmarks for offline safe reinforcement learning.
\newblock \emph{arXiv preprint arXiv:2306.09303}, 2023.

\bibitem[Mandlekar et~al.(2020)Mandlekar, Xu, Mart{\'\i}n-Mart{\'\i}n, Savarese, and Fei-Fei]{mandlekar2020learning}
Ajay Mandlekar, Danfei Xu, Roberto Mart{\'\i}n-Mart{\'\i}n, Silvio Savarese, and Li~Fei-Fei.
\newblock Learning to generalize across long-horizon tasks from human demonstrations.
\newblock \emph{arXiv preprint arXiv:2003.06085}, 2020.

\bibitem[Mirhoseini et~al.(2020)Mirhoseini, Goldie, Yazgan, Jiang, Songhori, Wang, Lee, Johnson, Pathak, Bae, et~al.]{mirhoseini2020chip_placement}
Azalia Mirhoseini, Anna Goldie, Mustafa Yazgan, Joe Jiang, Ebrahim Songhori, Shen Wang, Young-Joon Lee, Eric Johnson, Omkar Pathak, Sungmin Bae, et~al.
\newblock Chip placement with deep reinforcement learning.
\newblock \emph{arXiv preprint arXiv:2004.10746}, 2020.

\bibitem[Mirhoseini et~al.(2021)Mirhoseini, Goldie, Yazgan, Jiang, Songhori, Wang, Lee, Johnson, Pathak, Nazi, et~al.]{mirhoseini2021graph}
Azalia Mirhoseini, Anna Goldie, Mustafa Yazgan, Joe~Wenjie Jiang, Ebrahim Songhori, Shen Wang, Young-Joon Lee, Eric Johnson, Omkar Pathak, Azade Nazi, et~al.
\newblock A graph placement methodology for fast chip design.
\newblock \emph{Nature}, 594\penalty0 (7862):\penalty0 207--212, 2021.

\bibitem[Mnih et~al.(2015)Mnih, Kavukcuoglu, Silver, Rusu, Veness, Bellemare, Graves, Riedmiller, Fidjeland, Ostrovski, et~al.]{mnih2015human}
Volodymyr Mnih, Koray Kavukcuoglu, David Silver, Andrei~A Rusu, Joel Veness, Marc~G Bellemare, Alex Graves, Martin Riedmiller, Andreas~K Fidjeland, Georg Ostrovski, et~al.
\newblock Human-level control through deep reinforcement learning.
\newblock \emph{nature}, 518\penalty0 (7540):\penalty0 529--533, 2015.

\bibitem[Nair et~al.(2020)Nair, Gupta, Dalal, and Levine]{nair2020awac}
Ashvin Nair, Abhishek Gupta, Murtaza Dalal, and Sergey Levine.
\newblock Awac: Accelerating online reinforcement learning with offline datasets.
\newblock \emph{arXiv preprint arXiv:2006.09359}, 2020.

\bibitem[Parisi et~al.(2019)Parisi, Kemker, Part, Kanan, and Wermter]{PARISI201954}
German~I. Parisi, Ronald Kemker, Jose~L. Part, Christopher Kanan, and Stefan Wermter.
\newblock Continual lifelong learning with neural networks: A review.
\newblock \emph{Neural Networks}, 113:\penalty0 54--71, 2019.
\newblock ISSN 0893-6080.
\newblock \doi{https://doi.org/10.1016/j.neunet.2019.01.012}.
\newblock URL \url{https://www.sciencedirect.com/science/article/pii/S0893608019300231}.

\bibitem[Park et~al.(2024)Park, Frans, Eysenbach, and Levine]{park2024ogbench}
Seohong Park, Kevin Frans, Benjamin Eysenbach, and Sergey Levine.
\newblock Ogbench: Benchmarking offline goal-conditioned rl.
\newblock \emph{arXiv preprint arXiv:2410.20092}, 2024.

\bibitem[Paszke et~al.(2019)Paszke, Gross, Massa, Lerer, Bradbury, Chanan, Killeen, Lin, Gimelshein, Antiga, et~al.]{paszke2019pytorch}
Adam Paszke, Sam Gross, Francisco Massa, Adam Lerer, James Bradbury, Gregory Chanan, Trevor Killeen, Zeming Lin, Natalia Gimelshein, Luca Antiga, et~al.
\newblock Pytorch: An imperative style, high-performance deep learning library.
\newblock \emph{Advances in neural information processing systems}, 32, 2019.

\bibitem[Patterson(2021)]{Howwe’re47:online}
David Patterson.
\newblock How we’re minimizing ai’s carbon footprint.
\newblock \url{https://blog.google/technology/ai/minimizing-carbon-footprint/}, 2021.
\newblock (Accessed on 06/05/2024).

\bibitem[Patterson et~al.(2021)Patterson, Gonzalez, Le, Liang, Munguia, Rothchild, So, Texier, and Dean]{patterson2021carbon}
David Patterson, Joseph Gonzalez, Quoc Le, Chen Liang, Lluis-Miquel Munguia, Daniel Rothchild, David So, Maud Texier, and Jeff Dean.
\newblock Carbon emissions and large neural network training, 2021.

\bibitem[Peng et~al.(2020)Peng, Coumans, Zhang, Lee, Tan, and Levine]{peng2020learning_agile_imitate}
Xue~Bin Peng, Erwin Coumans, Tingnan Zhang, Tsang-Wei Lee, Jie Tan, and Sergey Levine.
\newblock Learning agile robotic locomotion skills by imitating animals.
\newblock \emph{arXiv preprint arXiv:2004.00784}, 2020.

\bibitem[Qin et~al.(2022)Qin, Zhang, Gao, Chen, Li, Zhang, and Yu]{qin2022neorl}
Rong-Jun Qin, Xingyuan Zhang, Songyi Gao, Xiong-Hui Chen, Zewen Li, Weinan Zhang, and Yang Yu.
\newblock Neorl: A near real-world benchmark for offline reinforcement learning.
\newblock \emph{Advances in Neural Information Processing Systems}, 35:\penalty0 24753--24765, 2022.

\bibitem[Rafailov et~al.(2024)Rafailov, Hatch, Singh, Smith, Kumar, Kostrikov, Hansen-Estruch, Kolev, Ball, Wu, et~al.]{rafailov2024d5rl}
Rafael Rafailov, Kyle Hatch, Anikait Singh, Laura Smith, Aviral Kumar, Ilya Kostrikov, Philippe Hansen-Estruch, Victor Kolev, Philip Ball, Jiajun Wu, et~al.
\newblock D5rl: Diverse datasets for data-driven deep reinforcement learning.
\newblock \emph{arXiv preprint arXiv:2408.08441}, 2024.

\bibitem[Reddi et~al.(2020)Reddi, Cheng, Kanter, Mattson, Schmuelling, Wu, Anderson, Breughe, Charlebois, Chou, et~al.]{reddi2020mlperf}
Vijay~Janapa Reddi, Christine Cheng, David Kanter, Peter Mattson, Guenther Schmuelling, Carole-Jean Wu, Brian Anderson, Maximilien Breughe, Mark Charlebois, William Chou, et~al.
\newblock Mlperf inference benchmark.
\newblock In \emph{2020 ACM/IEEE 47th Annual International Symposium on Computer Architecture (ISCA)}, pp.\  446--459. IEEE, 2020.

\bibitem[Roszel et~al.(2021)Roszel, Norvill, Hilger, and State]{roszel2021know}
Mary Roszel, Robert Norvill, Jean Hilger, and Radu State.
\newblock Know your model (kym): Increasing trust in ai and machine learning.
\newblock \emph{arXiv preprint arXiv:2106.11036}, 2021.

\bibitem[Schafhalter et~al.(2023)Schafhalter, Kalra, Xu, Gonzalez, and Stoica]{schafhalter2023leveraging}
Peter Schafhalter, Sukrit Kalra, Le~Xu, Joseph~E Gonzalez, and Ion Stoica.
\newblock Leveraging cloud computing to make autonomous vehicles safer.
\newblock In \emph{2023 IEEE/RSJ International Conference on Intelligent Robots and Systems (IROS)}, pp.\  5559--5566. IEEE, 2023.

\bibitem[Schulman et~al.(2017)Schulman, Wolski, Dhariwal, Radford, and Klimov]{schulman2017proximal}
John Schulman, Filip Wolski, Prafulla Dhariwal, Alec Radford, and Oleg Klimov.
\newblock Proximal policy optimization algorithms.
\newblock \emph{arXiv preprint arXiv:1707.06347}, 2017.

\bibitem[Sermanet et~al.(2018)Sermanet, Lynch, Chebotar, Hsu, Jang, Schaal, Levine, and Brain]{Sermanet2018TimeContrastive}
Pierre Sermanet, Corey Lynch, Yevgen Chebotar, Jasmine Hsu, Eric Jang, Stefan Schaal, Sergey Levine, and Google Brain.
\newblock Time-contrastive networks: Self-supervised learning from video.
\newblock In \emph{2018 IEEE International Conference on Robotics and Automation (ICRA)}, pp.\  1134--1141, 2018.
\newblock \doi{10.1109/ICRA.2018.8462891}.

\bibitem[Shi et~al.(2017)Shi, Karpathy, Fan, Hernandez, and Liang]{shi2017world}
Tianlin Shi, Andrej Karpathy, Linxi Fan, Jonathan Hernandez, and Percy Liang.
\newblock World of bits: An open-domain platform for web-based agents.
\newblock In \emph{International Conference on Machine Learning}, pp.\  3135--3144. PMLR, 2017.

\bibitem[Sun et~al.(2020)Sun, Kretzschmar, Dotiwalla, Chouard, Patnaik, Tsui, Guo, Zhou, Chai, Caine, et~al.]{waymo_open}
Pei Sun, Henrik Kretzschmar, Xerxes Dotiwalla, Aurelien Chouard, Vijaysai Patnaik, Paul Tsui, James Guo, Yin Zhou, Yuning Chai, Benjamin Caine, et~al.
\newblock Scalability in perception for autonomous driving: Waymo open dataset.
\newblock In \emph{Proceedings of the IEEE/CVF conference on computer vision and pattern recognition}, pp.\  2446--2454, 2020.

\bibitem[Sutton \& Barto(2018)Sutton and Barto]{sutton2018reinforcement}
Richard~S Sutton and Andrew~G Barto.
\newblock \emph{Reinforcement learning: An introduction}.
\newblock MIT press, 2018.

\bibitem[Tassa et~al.(2018)Tassa, Doron, Muldal, Erez, Li, de~Las~Casas, Budden, Abdolmaleki, Merel, Lefrancq, Lillicrap, and Riedmiller]{tassa2018deepmind}
Yuval Tassa, Yotam Doron, Alistair Muldal, Tom Erez, Yazhe Li, Diego de~Las~Casas, David Budden, Abbas Abdolmaleki, Josh Merel, Andrew Lefrancq, Timothy Lillicrap, and Martin Riedmiller.
\newblock Deepmind control suite, 2018.

\bibitem[Towers et~al.(2023)Towers, Terry, Kwiatkowski, Balis, Cola, Deleu, Goulão, Kallinteris, KG, Krimmel, Perez-Vicente, Pierré, Schulhoff, Tai, Shen, and Younis]{towers_gymnasium_2023}
Mark Towers, Jordan~K. Terry, Ariel Kwiatkowski, John~U. Balis, Gianluca~de Cola, Tristan Deleu, Manuel Goulão, Andreas Kallinteris, Arjun KG, Markus Krimmel, Rodrigo Perez-Vicente, Andrea Pierré, Sander Schulhoff, Jun~Jet Tai, Andrew Tan~Jin Shen, and Omar~G. Younis.
\newblock Gymnasium, March 2023.
\newblock URL \url{https://zenodo.org/record/8127025}.

\bibitem[Uchendu et~al.(2023)Uchendu, Xiao, Lu, Zhu, Yan, Simon, Bennice, Fu, Ma, Jiao, et~al.]{uchendu2023jump}
Ikechukwu Uchendu, Ted Xiao, Yao Lu, Banghua Zhu, Mengyuan Yan, Jos{\'e}phine Simon, Matthew Bennice, Chuyuan Fu, Cong Ma, Jiantao Jiao, et~al.
\newblock Jump-start reinforcement learning.
\newblock In \emph{International Conference on Machine Learning}, pp.\  34556--34583. PMLR, 2023.

\bibitem[van Hasselt et~al.(2015)van Hasselt, Guez, and Silver]{van2015ddqn}
HADO van Hasselt, ARTHUR Guez, and DAVID Silver.
\newblock Deep reinforcement learning with double q-learning. arxiv e-prints.
\newblock \emph{arXiv preprint arXiv:1509.06461}, 2015.

\bibitem[Wang et~al.(2016)Wang, Kurth-Nelson, Tirumala, Soyer, Leibo, Munos, Blundell, Kumaran, and Botvinick]{wang2016learning_to_reinforcement_learn}
Jane~X Wang, Zeb Kurth-Nelson, Dhruva Tirumala, Hubert Soyer, Joel~Z Leibo, Remi Munos, Charles Blundell, Dharshan Kumaran, and Matt Botvinick.
\newblock Learning to reinforcement learn.
\newblock \emph{arXiv preprint arXiv:1611.05763}, 2016.

\bibitem[Wu et~al.(2022)Wu, Raghavendra, Gupta, Acun, Ardalani, Maeng, Chang, Aga, Huang, Bai, et~al.]{wu2022sustainable}
Carole-Jean Wu, Ramya Raghavendra, Udit Gupta, Bilge Acun, Newsha Ardalani, Kiwan Maeng, Gloria Chang, Fiona Aga, Jinshi Huang, Charles Bai, et~al.
\newblock Sustainable ai: Environmental implications, challenges and opportunities.
\newblock \emph{Proceedings of Machine Learning and Systems}, 4:\penalty0 795--813, 2022.

\bibitem[Ye et~al.(2021)Ye, Liu, Kurutach, Abbeel, and Gao]{ye2021mastering}
Weirui Ye, Shaohuai Liu, Thanard Kurutach, Pieter Abbeel, and Yang Gao.
\newblock Mastering atari games with limited data.
\newblock \emph{Advances in Neural Information Processing Systems}, 34:\penalty0 25476--25488, 2021.

\bibitem[Yu et~al.(2020)Yu, Quillen, He, Julian, Hausman, Finn, and Levine]{meta_world}
Tianhe Yu, Deirdre Quillen, Zhanpeng He, Ryan Julian, Karol Hausman, Chelsea Finn, and Sergey Levine.
\newblock Meta-world: A benchmark and evaluation for multi-task and meta reinforcement learning.
\newblock In \emph{Conference on robot learning}, pp.\  1094--1100. PMLR, 2020.

\end{thebibliography}
